\documentclass[10pt,journal,compsoc]{IEEEtran}

\ifCLASSOPTIONcompsoc
  \usepackage[nocompress]{cite}
\else
  \usepackage{cite}
\fi
\ifCLASSINFOpdf
\else
\fi
\usepackage{color}
\definecolor{mybar}{rgb}{1.0, 0.4, 0.0}
\newcommand\cbar[3][mybar]{\colorbox{#1}{\color{black}\framebox(#2,#3){}}}

\usepackage{url}
\usepackage{hyperref}
\usepackage{subfigure}
\usepackage{graphicx}
\usepackage{amsmath}
\usepackage{booktabs}
\usepackage{algorithm}
\usepackage{algorithmic}

\usepackage{threeparttable}

\usepackage{amssymb}
\usepackage{multirow}
\usepackage{colortbl}
\usepackage{tabu}
\usepackage{pifont}
\usepackage{colortbl}
\usepackage[dvipsnames]{xcolor}
\usepackage{enumitem}
\usepackage{ragged2e}

\usepackage{tikz}
\usepackage{amssymb,mathrsfs,amsmath}

\usepackage[final]{changes}
\definechangesauthor[name={Per cusse}, color=orange]{per}

\begin{document}
%
\title{Hard-aware Instance Adaptive  \\Self-training for Unsupervised Cross-domain Semantic Segmentation}
%
%
%
%
\author{Chuang~Zhu$^{*}$,~\IEEEmembership{Member,~IEEE,}
        Kebin~Liu$^{*}$,
        Wenqi~Tang,
        Ke~Mei,
        Jiaqi~Zou,
        and~Tiejun~Huang,~\IEEEmembership{Senior~Member,~IEEE}
\IEEEcompsocitemizethanks{
\IEEEcompsocthanksitem Chuang Zhu, Kebin Liu, and Wenqi Tang are with the School of Artificial Intelligence, Beijing University of Posts and Telecommunications, Beijing 100876, China. (E-mail: czhu@bupt.edu.cn, liukebin@bupt.edu.cn, tangwenqi@bupt.edu.cn)
\IEEEcompsocthanksitem Ke Mei is with Tencent Wechat AI, Beijing 100080, China. (E-mail: raykoomei@tencent.com)
\IEEEcompsocthanksitem Jiaqi Zou is with the School of Information and Communication Engineering, Beijing University of Posts and Telecommunications, Beijing 100876, China. (E-mail: jqzou@bupt.edu.cn)
\IEEEcompsocthanksitem Tiejun Huang is with the School of Electronics Engineering and Computer Science, Peking University, No.5 Yiheyuan Road Haidian District, Beijing 100871, China. (E-mail: tjhuang@pku.edu.cn)
\IEEEcompsocthanksitem $^*$These authors contribute equally to this work.}
}

%
%

\markboth{IEEE Transactions on Pattern Analysis and Machine Intelligence}%
{Shell \MakeLowercase{\textit{et al.}}: Bare Advanced Demo of IEEEtran.cls for IEEE Computer Society Journals}
%



\IEEEtitleabstractindextext{%
\begin{abstract}
\justifying
The divergence between labeled training data and unlabeled testing data is a significant challenge for recent deep learning models. Unsupervised domain adaptation (UDA) attempts to solve such problem. Recent works show that self-training is a powerful approach to UDA. However, existing methods have difficulty in balancing the scalability and performance. In this paper, we propose a hard-aware instance adaptive self-training framework for UDA on the task of semantic segmentation. To effectively improve the quality and diversity of pseudo-labels, we develop a novel pseudo-label generation strategy with an instance adaptive selector. We further enrich the hard class pseudo-labels with inter-image information through a skillfully designed hard-aware pseudo-label augmentation. Besides, we propose the region-adaptive regularization to smooth the pseudo-label region and sharpen the non-pseudo-label region. For the non-pseudo-label region, consistency constraint is also constructed to introduce stronger supervision signals during model optimization. Our method is so concise and efficient that it is easy to be generalized to other UDA methods. Experiments on GTA5 $\rightarrow$ Cityscapes, SYNTHIA $\rightarrow$ Cityscapes, and Cityscapes $\rightarrow$ Oxford RobotCar demonstrate the superior performance of our approach compared with the state-of-the-art methods. {Our codes are available at \url{https://github.com/bupt-ai-cz/HIAST}.}
\end{abstract}

}

\maketitle

\IEEEdisplaynontitleabstractindextext

%
\IEEEpeerreviewmaketitle

\IEEEraisesectionheading{\section{Introduction}\label{sec:introduction}}

\IEEEPARstart{D}EEP neural networks have achieved remarkable success in the field of semantic segmentation, yielding notable progress \cite{chen2017deeplab, DeepLabv3_chen2017rethinking,chen2018encoder}. However, domain shifts usually hamper the generalization of the segmentation model to unseen environments.
Domain shifts refer to the divergence between training data (source domain) and the testing data (target domain), induced by factors such as the variance in illumination, object viewpoints, and image backgrounds \cite{chen2018domain, tsai2018learning}.
Such domain shifts often lead to the phenomenon that the trained model suffers from a significant performance drop on the unlabeled target domain.
To tackle this issue, an intuitive solution is to manually build dense pixel-level annotations for each target domain, which is notoriously laborious and expensive. The unsupervised domain adaptation (UDA) methods aim to address the above problem by transferring knowledge from the labeled source domain to the unlabeled target domain.

Recently, the adversarial training (AT) methods have received critical attention for UDA semantic segmentation{\cite{tsai2018learning, luo2019taking, luosignificance, du2019ssf, vu2019advent, Tsai_adaptseg_ICCV19, yang2019adversarial, huang2020contextual, Haoran_2020_ECCV, BCDM_Li2021BiClassifierDM, CDGA_Kim_Joung_Kim_Park_Kim_Sohn_2021}}. These methods aim to minimize a series of adversarial losses to learn invariant representations across domains, thereby aligning source and target feature distributions. 
More recently, an alternative research line to reduce domain shifts focuses on building schemes based on the self-training (ST) framework{\cite{zou2018unsupervised, zou2019confidence, lian2019constructing, MLSL_iqbal2020mlsl, CCM_li2020content, PIT_lv2020cross, PixMatch_melas2021pixmatch, CLST_Marsden2021ContrastiveLA, SPCL_xie2021spcl, MetaCorrection_guo2021metacorrection, DACS_Tranheden_2021_WACV, RCCR_zhou2021domain, SAC_araslanov2021self, DSP_gao2021dsp, CAMix_zhou2021context, xie2022sepico, chen2022deliberated}}. These works iteratively train the model by using both the labeled source domain data and generated pseudo-labels of unlabeled target domain data, thus achieving alignment between source and target domains. 
Besides, several works{\cite{zhang2019category, IntraDA_pan2020unsupervised, zheng2019unsupervised, DTST_wang2020differential, DAST_yu2021dast, Zheng2021RectifyingPL, UPLR_wang2021uncertainty, ProDA_CRA_wang2021cross, li2022class}} have explored combining AT and ST methods, which shows great potential for UDA semantic segmentation. 
The combined methods with image-to-image translation are also proposed{\cite{hoffman2018cycada, LDR_yang2020label, CDAM_yang2021context, Yang_2020_CVPR, MCS_chung2022maximizing, SFG_DA_cardace2022shallow, li2019bidirectional, LTIR_kim2020learning, SAI2I_Luigi2020, ContenTransfer_Lee2021UnsupervisedDA, ITEN_piva2021exploiting, TridentAdapt_shen2021tridentadapt, DPL_cheng2021dual, CoarseToFine_ma2021coarse}}, which minimize the domain gap under the assist of style transfer.
Through carefully designed network structure, these methods achieve the state-of-the-art (SOTA) performance on the benchmark. Although these mixed methods can achieve better performance, the serious coupling between sub-modules causes the degradation of scalability and flexibility.

This paper aims to propose a self-training framework for UDA semantic segmentation, which has good scalability that can be easily applied to other non-self-training methods and achieves new state-of-the-art performance. To achieve this, we locate the main obstacles of existing self-training methods are how to generate high-quality and class-balanced pseudo-labels. This paper designs a hard-aware adaptive pseudo-label generation strategy and model regularization to solve these obstacles.
\begin{figure*}[htb]
\centering
\subfigure[\added{Ground truth}]{
\includegraphics[width=0.235\linewidth]{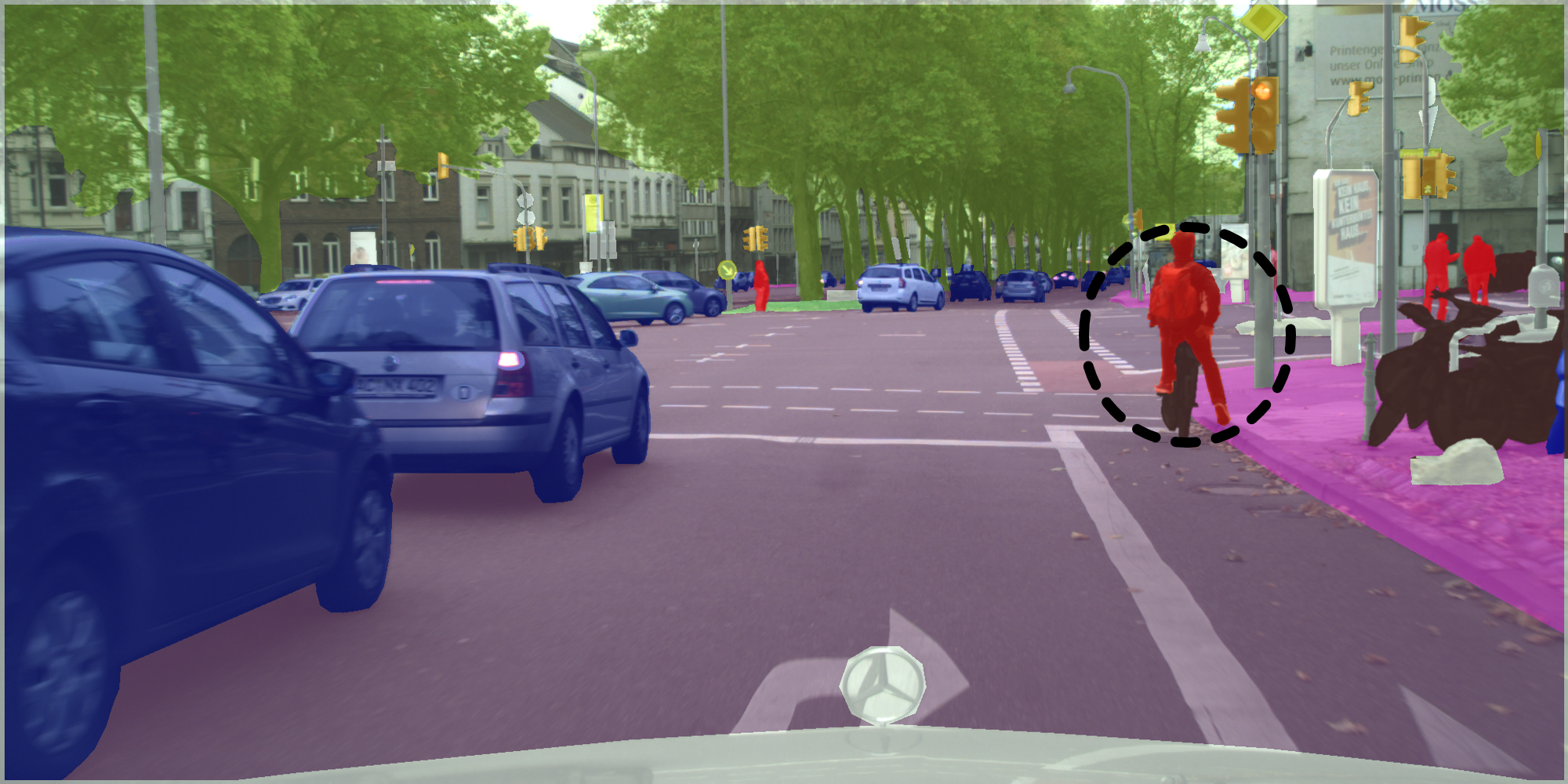}
}
\subfigure[\added{CBST\cite{zou2018unsupervised}}]{
\includegraphics[width=0.235\linewidth]{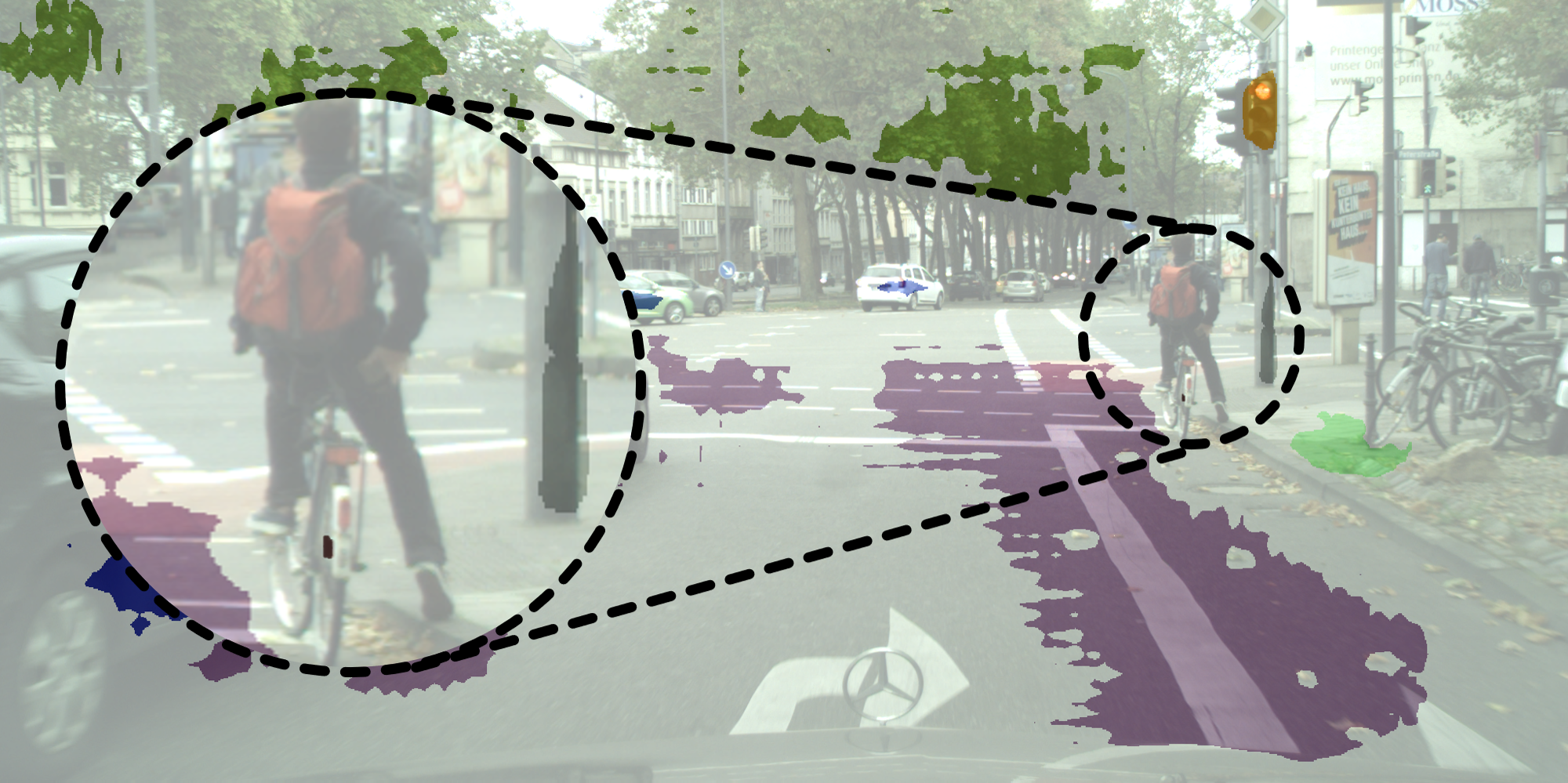}
\label{fig:intro_pl_cbst}
}
\subfigure[\added{IAST (Ours)}]{
\includegraphics[width=0.235\linewidth]{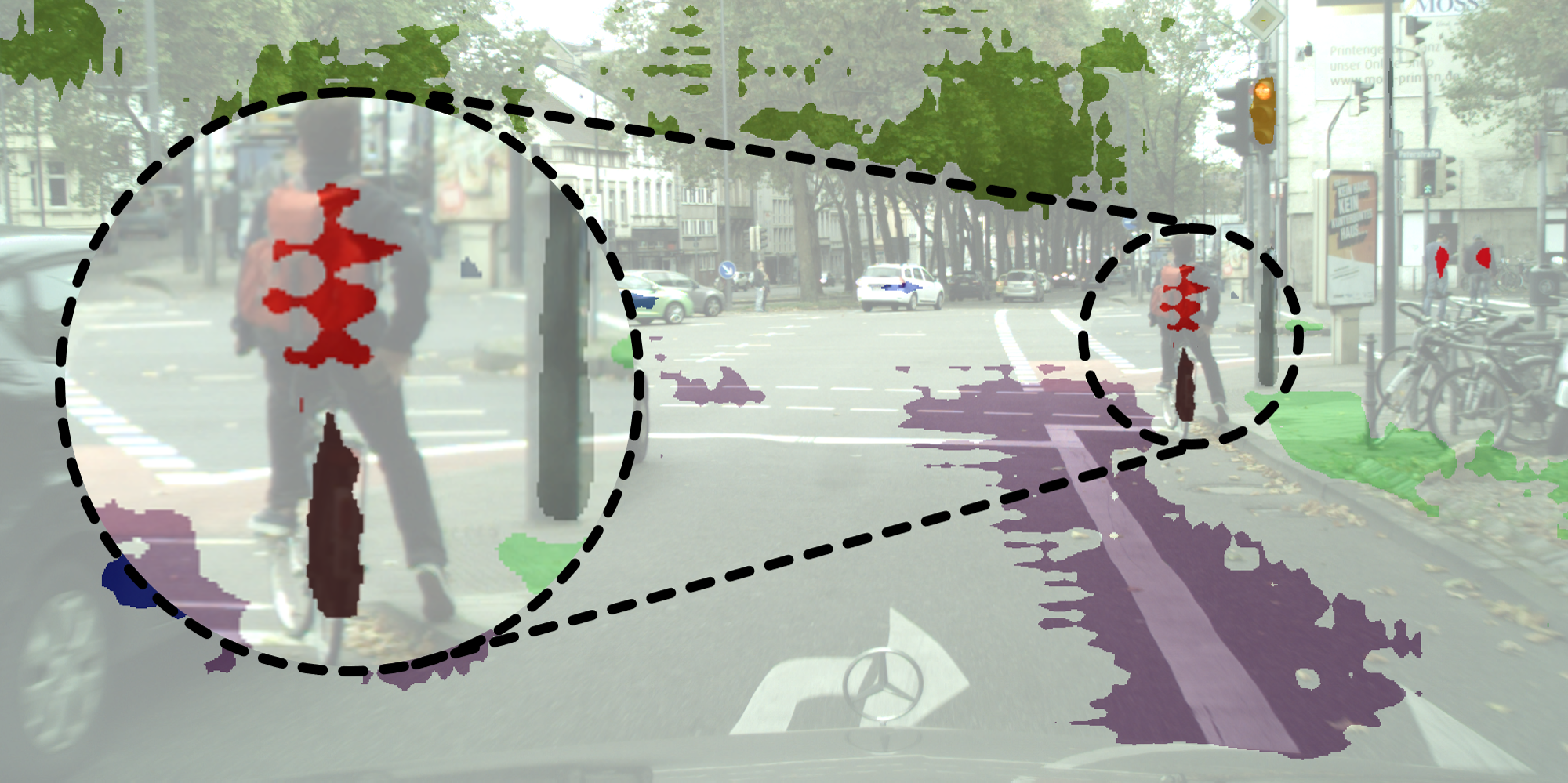}
\label{fig:intro_pl_iast}
}
\subfigure[\added{HIAST (Ours)}]{
\includegraphics[width=0.235\linewidth]{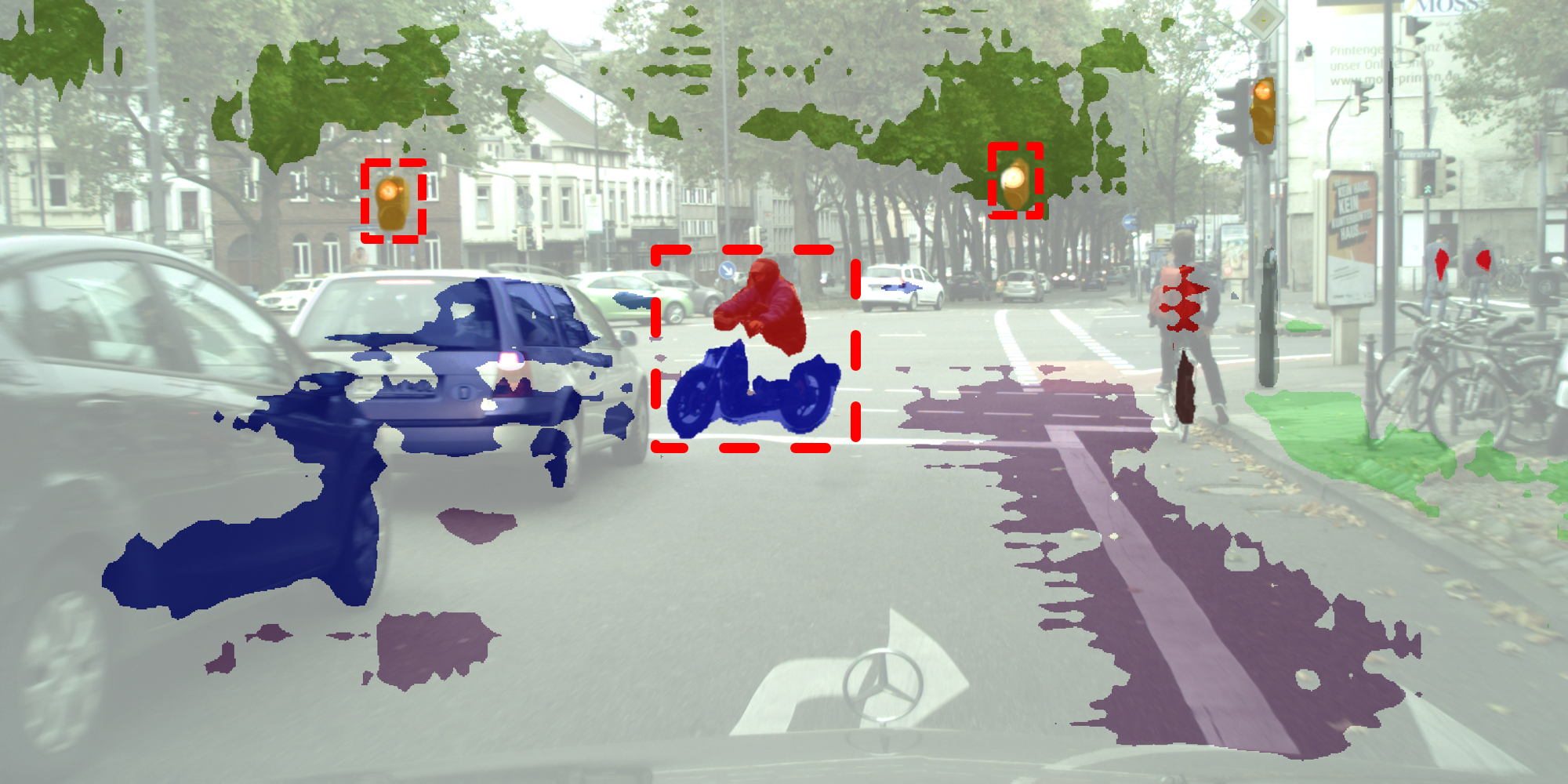}
\label{fig:intro_pl_hiast}
}
\vspace{-12pt}
\caption{Illustration for the results of pseudo-labels. (a): Ground truth. (b): CBST is biased to such predominant classes as road and vegetation, other classes are almost ignored. (c): IAST has improved the diversity of categories and produced more valid regions, especially for these hard classes such as rider and bike. (d): HIAST further improves the proportion of hard classes by augmentation which transfers pixels (regions surrounded by red dashed line) from other target domain images. For a fair comparison, all pseudo-labels are generated by the same model, with the proportion of about 20\% for pseudo-labels in the target domain dataset.}
\vspace{-8pt}
\label{fig:itro_pl}
\end{figure*}

The existing pseudo-label generation suffers from information redundancy and noise. The generator tends to keep pixels with high confidences as pseudo-labels and ignore pixels with low confidences. Because of this conservative threshold selection, it is inefficient when more similar pixels with high confidences are applied to training. The class-balanced self-training (CBST) \cite{zou2018unsupervised} utilizes rank-based reference confidences for each class among all related images. This will result in the ignorance of key information for some images, where almost all the hard-class pixels have low predicted scores. For example, in Fig. \ref{fig:intro_pl_cbst}, the pseudo-labels generated by CBST are concentrated on the road, while pedestrians are ignored. Therefore, it is important to design an instance adaptive pseudo-label generation strategy to reduce data redundancy and increase the diversity of pseudo-labels. Besides, the pseudo-label areas for predominant classes, such as road, car, and vegetation, are usually much larger than the other small-size categories, such as person, traffic sign, and traffic light. Due to the lack of high-quality pseudo-label regions for these small-size hard classes, the trained model is prone to bias to the predominant easy classes.

In this work, we propose a hard-aware instance adaptive self-training framework (HIAST) for UDA semantic segmentation, as shown in Fig. \ref{fig:intro_framework}. Firstly we initialize the segmentation model by adversarial training. Then we employ an instance adaptive selector (IAS) in considering pseudo-label diversity during the training process. We also develop a hard-aware pseudo-label augmentation (HPLA) scheme with inter-image information to further improve the performance for hard classes. Besides, we design region-adaptive regularization, which has different roles in pseudo-label regions and non-pseudo-label regions. 
The main contributions of our work are summarized as follows:

\begin{itemize}[leftmargin=1em]
\item We propose a new hard-aware instance adaptive self-training framework. Our method significantly improve
current state-of-the-art methods on the open UDA semantic segmentation benchmarks.
\item We design the IAS to involve more useful information for training. It effectively improves the quality of pseudo-labels, especially for hard classes. To further improve and enrich pseudo-labels for hard classes, we propose the HPLA scheme to conduct pseudo-label fusion among different images from the target domain. The proposed HPLA can select hard classes and dynamically raise their proportions during training, without introducing extra domain gap and noise.

\begin{figure}[htb]
\centering
\subfigure[Warm-up]{
\begin{minipage}[t]{0.9\linewidth}
\hspace{-10.5pt}
\includegraphics[width=\linewidth]{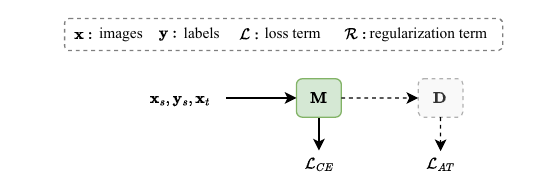}
\end{minipage}
}

\vspace{-5pt}
\subfigure[Pseudo-label generation]{
\begin{minipage}[t]{0.65\linewidth}
\hspace{-12.5pt}
\includegraphics[width=\linewidth]{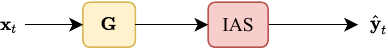}
\end{minipage}
}

\vspace{-5pt}
\subfigure[Self-training]{
\begin{minipage}[t]{0.8\linewidth}
\includegraphics[width=\linewidth]{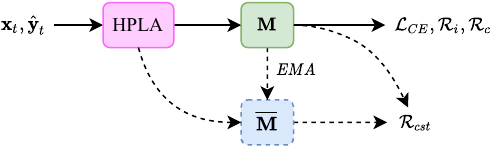}
\end{minipage}
}
\vspace{-10pt}
\caption{The pipeline of proposed HIAST. (a): Warm-up phase, an initial model $\mathbf{M}$ is trained using AT with discriminator $\mathbf{D}$. (b): \added{Pseudo-label generation phase, the selector IAS filters pseudo-labels generated by $\mathbf{G}$, where $\mathbf{G}$ is the frozen $\mathbf{M}$.} (c): Self-training phase, target images and corresponding pseudo-labels are augmented by HPLA during self-training; $\overline{\mathbf{M}}$ is the copy of $\mathbf{M}$ and updated by exponential moving average (EMA) strategy with $\mathbf{M}$.}
\vspace{-10pt}
\label{fig:intro_framework}
\end{figure}

\item Region-based regularization is designed to smooth the prediction of the pseudo-label region and sharpen the prediction of the non-pseudo-label region. A consistency constraint is further developed to provide stronger supervising information for the non-pseudo-label region.
\item We propose a general approach that makes it easy to apply our framework to other methods. Moreover, our framework can also be extended to semi-supervised semantic segmentation tasks. 
In addition, we design a novel parameter selection method under the UDA setting, which can choose parameters without any ground-truth.
\end{itemize}

We have presented the preliminary version of this work in IAST\cite{mei2020instance}. This paper further extends the previous work in several aspects. First, we extend IAST to HIAST by introducing the hard-aware pseudo-label augmentation (HPLA). Second, we build a consistency constraint regularization to provide stronger supervision for the ignored region. Third, we conduct a comprehensive comparison between our method and the recent SOTAs. 
{Furthermore, we introduce a novel parameter selection strategy under the UDA setting.} 
Last, we show more in-depth analysis and discussion regarding the effectiveness and limitations of our work.

The paper is organized as follows: in Section \ref{sec:RelatedWork}, we review techniques that are related to our work; in Section \ref{sec:pre}, we show some preliminaries about UDA for semantic segmentation, self-training, and adversarial training; in Section \ref{sec:proposed-method}, we build the whole HIAST framework based on IAST; after that, we report the experimental results in Section \ref{sec:experiment};finally, the conclusions of this paper are summarized in Section \ref{sec:conclusion}.

\section{Related Works}
\label{sec:RelatedWork}

\subsection{UDA for Semantic Segmentation}

\noindent\textbf{Adversarial Training.} The adversarial training based UDA model for semantic segmentation usually consists of one generator and one discriminator. The generator is leveraged to extract image features and obtain the final predicted segmentation maps, while the discriminator is trained for the domain prediction. With adversarial training, the gap of feature representations between source and target domains is gradually reduced. AdaptSegNet\cite{tsai2018learning} considers the spatial similarities between different domains and performs the output space domain adaptation by a multi-level network. Furthermore, PatchAlign\cite{Tsai_adaptseg_ICCV19} bridges source and target domains by a clustered space of patch-wise features, and aligns feature representations of patches between source and target domains by adversarial training. However, the aforementioned methods just focus on the global alignment and the category-specific adaptation is ignored. CLAN\cite{luo2019taking} takes the category-level joint distribution into account to achieve the adaptive alignment for different classes; specifically, it introduces a category-level adversarial network to enforce local semantic consistency, and it also increases the weights of the adversarial loss for the poorly aligned classes, thus producing better performance. Similarly, FADA\cite{Haoran_2020_ECCV} leverages a novel domain discriminator to capture category-level information; furthermore, it also generalizes the binary domain label to domain encoding for better fine-grained feature alignment. In our scheme, we first apply the adversarial training for warm-up to roughly align the source and target domains on output space before the self-training phase.

\noindent\textbf{Self-training.} Self-training schemes are commonly used in semi-supervised learning (SSL) areas \cite{li2005setred}. For UDA, self-training iteratively trains the model by using both the labeled source domain data and generated pseudo-labels of unlabelled target domain data, which can provide domain-specific information, and thus achieve the alignment between the source and target domains \cite{triguero2015self}. A threshold is usually required for pseudo-label generation. However, the strategy of constant threshold neglects the differences between categories, which will result in the model training biases towards easy classes, thus ruining the adaptation performance for hard classes. To solve this problem, CBST\cite{zou2018unsupervised} proposes a class-balanced self-training scheme for UDA semantic segmentation, which designs threshold for each category and shows competitive domain adaptation performance. {Moreover, IR$^2$F-RMM\cite{gong2023continuous} further corrects the selected pseudo-labels by regarding the pixel-level pseudo-label values as continuous signals.}
CRST\cite{zou2019confidence} integrates a variety of confidence regularization to the selected pseudo-labels, producing better domain adaption results. {More recently, Adaboost\cite{9852150} has focused on hard samples in the target domain to mitigate the issue of significant model performance fluctuations during training.} Meanwhile, the self-training methods combined with adversarial training have also been proposed, such as \cite{zheng2019unsupervised, zhang2019category, IntraDA_pan2020unsupervised, DTST_wang2020differential, DAST_yu2021dast}.

Despite the success of self-training on UDA semantic segmentation, the serious problem of lacking supervision information for hard classes has not been solved, thus yielding poor performance for the long-tail categories in the target domain. The discrepancy between instances has not been considered yet in the current threshold strategy, thus introducing extra noise and reducing available pseudo-labels.

Different from the above UDA methods, we first combine the adversarial training and self-training flexibly to roughly align the source and target domains. Then we take a different pseudo-label generating approach, IAST, to obtain more high-quality pseudo-labels for hard classes, thus improving the diversity of target domain pseudo-labels and realizing fine domain alignment.

\subsection{Copy-and-Paste in UDA}

The augmentation of copy-and-paste has gained significant progress in the area of classification and segmentation, where the key idea is to mix regions from multiple images, thus producing a new image with fused information. Copy-and-paste originates from MixUp\cite{MixUp_zhang2017mixup} and CutMix\cite{CutMix_yun2019cutmix} for classification and is applied to semantic segmentation by \cite{CopyPaste_ghiasi2021simple, dwibedi2017cut, remez2018learning}. ClassMix\cite{ClassMix_olsson2021classmix} has further extended copy-and-paste to semi-supervised learning.

DACS\cite{DACS_Tranheden_2021_WACV} is the first work that integrates copy-and-paste into UDA for semantic segmentation, by directly copying pixels and masks of selected classes from the source domain to the target domain. Recently, ContentTransfer\cite{ContenTransfer_Lee2021UnsupervisedDA} firstly separates the content and style of an image to achieve domain alignment, and then conducts the same copy-and-paste for the long-tail classes to improve the class imbalance.

Despite the alleviation of class imbalance by transferring pixels from the source domain, the inevitably introduced domain gaps could suppress the further improvement of performance. Compared with these methods, our scheme is directly performed on the target domain with pseudo-labels and thus avoids bringing additional domain shifts. The difference in results between the source and target domains is shown in Appendix C. Additionally, our method can adaptively select hard classes and significantly increase their frequencies.

\subsection{Model Optimization for UDA}

\noindent\textbf{Regularization.} Regularization refers to schemes that intend to reduce the testing error and thus make the trained model generalize well to unseen data \cite{goodfellow2016deep, kukavcka2017regularization}. For deep learning, different kinds of regularization schemes such as weight decay \cite{krizhevsky2012imagenet} and label smoothing \cite{szegedy2016rethinking} are proposed. The recent work \cite{zou2019confidence} has designed label and model regularization under self-training architecture for UDA; however, the proposed regularization scheme is just applied to pseudo-label regions. The regularization on ignored regions could be also necessary, such as the spatial priors in CBST\cite{zou2018unsupervised}, which are class frequencies counted in the source domain.

\noindent\textbf{Consistency Training.} As a general model optimization strategy, consistency training has been widely spread in the semi-supervised learning for classification\cite{MixMatch_berthelot2019mixmatch, ReMixMatch_berthelot2019remixmatch,FixMatch_sohn2020fixmatch, UDA_NEURIPS2020_44feb009}, with the assumption that the model should output similar predictions when fed different perturbed versions of the same image.

More recently, consistency training is utilized for self-training based UDA semantic segmentation\cite{CrDoCo_chen2019crdoco, zhou2020uncertainty, SAC_araslanov2021self, PixMatch_melas2021pixmatch} under the Mean-Teacher\cite{MeanTeacher_Tarvainen2017MeanTA} framework. Mean-Teacher consists of a teacher model and a student model, where the teacher model shares the same architecture as the student model and is updated by the student model with an exponential moving average (EMA) strategy. The key idea behind Mean-Teacher is that the model should be invariant to perturbations of unlabelled data.

CrDoCo\cite{CrDoCo_chen2019crdoco} proposes an image-to-image translation method and enforces the model to produce consistent predictions for target images with different styles by a cross-domain consistency loss, thus minimizing the domain shifts on pixel-level. 
Furthermore, Zhou et al.\cite{zhou2020uncertainty} equip the consistency regularization with an uncertainty-aware module, which dynamically adjusts the weight of consistency loss to mine samples with high confidence for domain adaptation. 
Recently, SAC\cite{SAC_araslanov2021self} performs consistency training across multiple scales and flips, and modifies thresholds and loss weights for long-tail categories, producing better performance. 
Besides, PixMatch\cite{PixMatch_melas2021pixmatch} has explored the influences of different consistency schemes for the self-training based UDA method on semantic segmentation. 

Different from the above methods, our optimizing strategy adaptively imposes suitable constraints on the confident and ignored regions at the same time. Note that we also perform consistency training as a constraint on the ignored region to strengthen supervising information during model training.

\section{Preliminary}
\label{sec:pre}
\subsection{UDA for Semantic Segmentation}
It is assumed that there are two domains: source domain $S$ and target domain $T$. The source domain includes images $\mathbb{X}_{S}=\{\mathbf{x}_{s}\}$, semantic masks $\mathbb{Y}_{S}=\{\mathbf{y}_{s}\}$, and the target domain only has images $\mathbb{X}_{T}=\{\mathbf{x}_{t}\}$. In UDA, the semantic segmentation model $\mathbf{M}$ is trained only from the ground truth $\mathbb{Y}_{S}$ as the supervision signal. UDA semantic segmentation model can be defined as follows:
$$\{\mathbb{X}_{S}, \mathbb{Y}_{S}, \mathbb{X}_{T}\}\Rightarrow \mathbf{M}.$$

$\mathbf{M}$ uses some special losses and domain adaptation methods to learn domain-invariant feature representations, thereby aligning the feature distribution of two domains.

\subsection{Self-training for UDA}
Because the ground truth labels of the target domain are not available, we can treat the target domain as an extra unlabeled dataset. In this case, the UDA task can be transformed into the SSL task. Self-training is an effective method for SSL. The problem can be described as the following form:
\begin{equation}
\begin{aligned}
\min \limits_{\mathbf{w}} \mathcal{L}_{CE} = 
& - \frac{1}{\left | \mathbb{X}_{S} \right |} \sum_{\mathbf{x_{s}} \in \mathbb{X}_{S}} \sum_{c=1}^{C}y_{s}^{(c)}\log p(c|\mathbf{x}_{s}, \mathbf{w}) \\
& - \frac{1}{\left | \mathbb{X}_{T} \right |} \sum_{\mathbf{x}_{t} \in \mathbb{X}_{T}} \sum_{c=1}^{C}\hat{y}_{t}^{(c)}\log p(c|\mathbf{x}_{t}, \mathbf{w}),
\label{eq:1}
\end{aligned}
\end{equation}
where $C$ is the number of classes, $y_{s}^{(c)}$ indicates the label of class $c$ in the source domain, and $\hat{y}_{t}^{(c)}$ indicates the pseudo-label of class $c$ in the target domain. $\mathbf{x}_{s}$ and $\mathbf{x}_{t}$ are input images, $\mathbf{w}$ indicates weights of $\mathbf{M}$,  $p(c|\mathbf{x},\mathbf{w})$ is the predicted probability of class $c$ in softmax output, and $\left | \mathbb{X} \right |$ indicates the number of images.

In particular, $\hat{\mathbb{Y}}_{T}=\{ \hat{\mathbf{y}}_{t}\}$ are pseudo-labels generated by the trained model, which is limited to a one-hot vector (only single 1 and all the others 0) or an all-zero vector.
The pseudo-label can be used as approximate target ground truth.

\subsection{Adversarial Training for UDA}
Adversarial training based UDA methods use a discriminator $\mathbf{D}$ to align feature distributions, and discriminator $\mathbf{D}$ attempts to distinguish the feature distribution of source (label 1) and target (label 0) domains in the output space. The segmentation model $\mathbf{M}$ (with image $\textbf{x}$ as input) attempts to fool the discriminator for confusing the feature distribution of source and target domains, thereby obtaining domain-invariant feature representations. The optimization process of $\mathbf{M}$ and $\mathbf{D}$ is shown in Eq.\eqref{eq:2_M} and Eq.\eqref{eq:2_D}:
\begin{equation}
\begin{aligned}
\min \limits_{\mathbf{M}} \mathcal{L}_{M} = &- \frac{1}{\left | \mathbb{X}_{S} \right |} \sum_{\mathbf{x}_{s} \in \mathbb{X}_{S}} \sum_{c=1}^{C}y_{s}^{(c)}\log p(c|\mathbf{x}_{s}, \mathbf{w}) \\
&+ \frac{\lambda_{AT}}{\left | \mathbb{X}_{T} \right |} \sum_{\mathbf{x}_{t} \in \mathbb{X}_{T}} \left [ \mathbf{D}(\mathbf{M}(\mathbf{x}_{t}, \mathbf{w})) - \boldsymbol{1}  \right ]^{2},
\label{eq:2_M}
\end{aligned}
\end{equation}
where the first term is the cross-entropy loss of the source domain, and $p(c|\mathbf{x},\mathbf{w})$ is the predicted probability of class $c$ in softmax output $\mathbf{M}(\mathbf{x}, \mathbf{w})\in\mathbb{R}^{H\times W\times C}$; the second term uses a mean squared error (MSE) as the adversarial loss with corresponding weight $\lambda_{AT}$.
\begin{equation}
\begin{aligned}
\max \limits_{\mathbf{D}} \mathcal{L}_{D} &=\frac{1}{\left | \mathbb{X}_{S} \right |} \sum_{\mathbf{x}_{s} \in \mathbb{X}_{S}} \left [ \mathbf{D}(\mathbf{M}(\mathbf{x}_{s}, \mathbf{w})) \right ]^{2} \\
&+ \frac{1}{\left | \mathbb{X}_{T} \right |} \sum_{\mathbf{x}_{t} \in \mathbb{X}_{T}} \left [ \mathbf{D}(\mathbf{M}(\mathbf{x}_{t}, \mathbf{w})) - \boldsymbol{1}  \right ]^{2}.
\label{eq:2_D}
\end{aligned}
\end{equation}

\section{Proposed Method}
\label{sec:proposed-method}
\subsection{Overview of Our Method}

\begin{figure*}[htb] 
\centering
\subfigure[Pseudo-label generation]{
\label{fig:framework_generate_pseudo_label}
\includegraphics[width=0.9\linewidth]{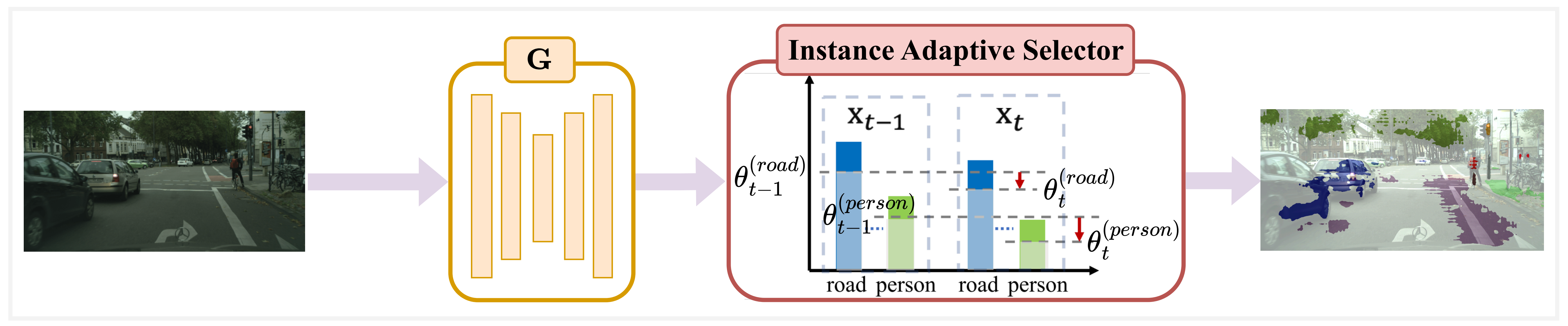}
}
\vspace{-5pt}

\subfigure[Self-training]{
\label{fig:framework_self_training}
\includegraphics[width=0.9\linewidth]{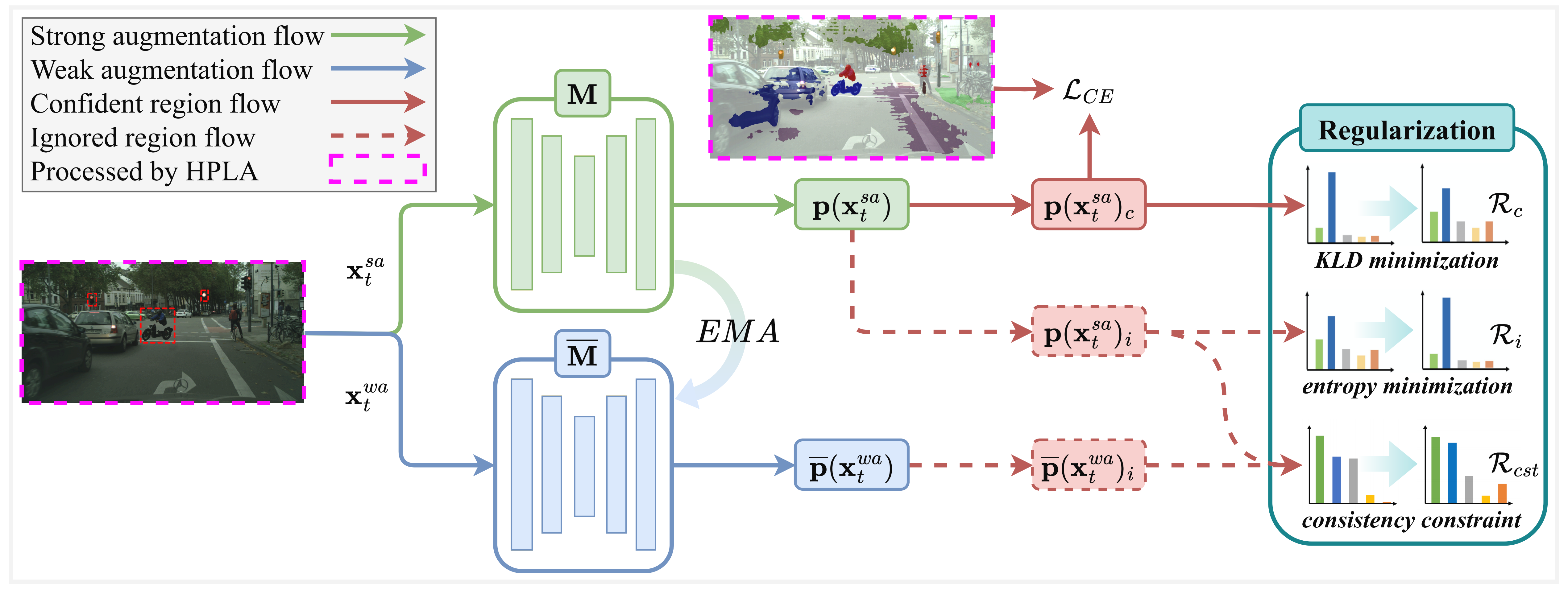}
}
\vspace{-10pt}
\caption{The core flows of HIAST. (a): Before self-training, pseudo-label of the target domain is produced by IAS with $\mathbf{G}$ which is initialized by the warm-up. IAS has combined the global and local information during pseudo-label generation, thus providing adaptive selection thresholds for different classes. (b): During self-training, to enrich the proportions of hard classes, the target domain image and corresponding pseudo-label are first processed by HPLA. Then, the target image is augmented by strong and weak augmentation; following \cite{MeanTeacher_Tarvainen2017MeanTA}, the strong perturbed version is fed into $\mathbf{M}$ for computing segmentation loss, and the weak perturbed version is fed into $\overline{\mathbf{M}}$ for consistency training. Furthermore, the regularization is imposed to avoid model overconfident to the pseudo-labels and sharpen the prediction on ignored regions.
}
\vspace{-12pt}
\label{fig:framework} 
\end{figure*}

We propose a hard class aware instance adaptive self-training framework (HIAST) with an instance adaptive selector (IAS), a hard-aware pseudo-label augmentation (HPLA), and region-adaptive regularization. IAS selects an adaptive pseudo-label threshold for each semantic category in the unit of image and dynamically filters out the tail of each category, to improve the diversity of pseudo-labels and eliminate noise. HPLA performs a pseudo-label augmentation for hard classes with lower {thresholds on the target domain}
during training. The region-adaptive regularization is further designed to smooth the prediction of the confident region and sharpen the prediction of the ignored region. Furthermore, the consistency constraint is proposed to provide stronger supervision signals for the ignored region by momentum model, forcing the consistency on model predictions between different perturbed versions of the same image. Our overall objective function of self-training is as follows:
\begin{equation}
\begin{aligned}
    &\min \limits_{\mathbf{w}}  \mathcal{L}_{CE}(\mathbf{w},\hat{\mathbb{Y}}_{T}) + \mathcal{L}_{R}(\mathbf{w}, \mathbf{\overline{w}}) \\ 
    & =\mathcal{L}_{CE}(\mathbf{w},\hat{\mathbb{Y}}_{T}) 
    + \lambda_{i}\mathcal{R}_{i}(\mathbf{w})  
    + \lambda_{c}\mathcal{R}_{c}(\mathbf{w})
    + \lambda_{cst}\mathcal{R}_{cst}(\mathbf{w}, \mathbf{\overline{w}}),
\end{aligned}
\label{eq:all}
\end{equation}
where $\mathcal{L}_{CE}$ is the cross-entropy loss, which is different from Eq.\eqref{eq:1} and only calculated on confident regions of target domain images. $\hat{\mathbb{Y}}_{T}$ is the set of pseudo-labels, and the detailed generation process is described in Section \ref{subsection:p}. $\mathcal{R}_{i}$, $\mathcal{R}_{c}$, and $\mathcal{R}_{cst}$ are regular terms described in Section \ref{subsection: r}, where $\mathcal{R}_{c}$ is performed on the confident region, while $\mathcal{R}_{i}$ and $\mathcal{R}_{cst}$ are deployed on the ignored region; $\lambda_{i}$, $\lambda_{c}$, and $\lambda_{cst}$ are corresponding weights.

The HIAST training process consists of three phases:
\begin{itemize}[leftmargin=1.75em]
\item[(a)] In the warm-up phase, an adversarial training based method uses both the source and target images to train a segmentation model $\mathbf{M}_0$ as the initial pseudo-label generator $\mathbf{G}$.
\item[(b)] In the pseudo-label generation phase, $\mathbf{G}$ is used to obtain predicted results of the target domain images, and the pseudo-label is generated by IAS, as shown in Fig. \ref{fig:framework_generate_pseudo_label}.
\item[(c)] In the self-training phase, the target domain image and corresponding pseudo-label are firstly processed by HPLA, after that the results are perturbed by strong and weak augmentation respectively, and then according to Eq.\eqref{eq:all}, the segmentation model is trained using the augmented target data, as shown in Fig. \ref{fig:framework_self_training}.

\end{itemize}

\noindent\textbf{Why Warm-up?} Before self-training, we expect to have a stable pre-trained model so that HIAST can be trained in the right direction and avoid disturbances caused by constant fitting the noise of pseudo-labels. We use the adversarial training method described in Section 3.3 to obtain a stable model by roughly aligning model outputs of the source and target domains. In addition, in the warm-up phase, we can optionally apply any other UDA semantic segmentation scheme as the basic method, and it can be retained even in the (c) phase. In fact, we can use HIAST as a decorator to decorate other basic methods.

\noindent\textbf{Multi-round Self-training.} Performing (b) and (c) phases once counts as one round. As with other self-training tasks, in this experiment, we perform a total of three rounds. At the end of each round, the parameters of momentum model $\overline{\mathbf{M}}$ will be copied into pseudo-label generator $\mathbf{G}$ to generate better target domain prediction results in the next round.

\begin{figure*}[htb]
\centering
\subfigure[Constant threshold]{
\label{fig:pseudo-methods_ct}
\hspace{0.025\linewidth}
\includegraphics[width=0.25\linewidth]{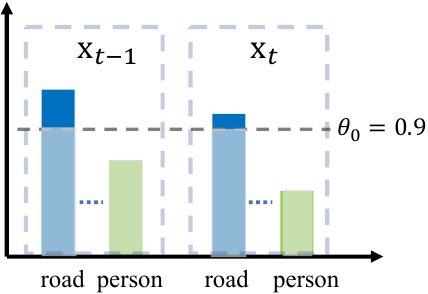}
}
\subfigure[Class-balanced threshold]{
\label{fig:pseudo-methods_cbst}
\hspace{0.025\linewidth}
\includegraphics[width=0.25\linewidth]{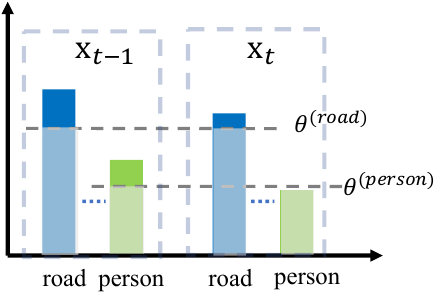}
}
\subfigure[Instance adaptive threshold]{
\label{fig:pseudo-methods_iast}
\hspace{0.025\linewidth}
\includegraphics[width=0.31\linewidth]{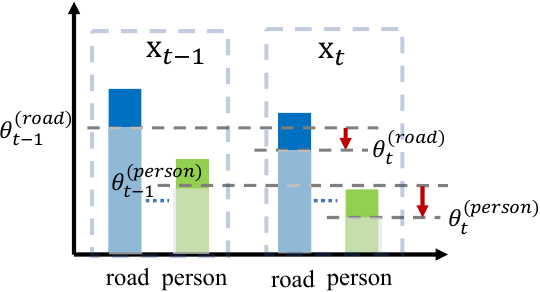}
}
\vspace{-10pt}
\caption{Illustration of three different threshold methods. $\mathbf{x}_{t-1}$ and $\mathbf{x}_{t}$ represent two consecutive instances, and the bars approximately represent the probabilities of each class. (a): A constant threshold is used for all instances. (b): Class-balanced thresholds are used for all instances. (c): Our method adaptively adjusts the threshold of each class based on the instance.}
\vspace{-10pt}
\label{fig:pseudo-methods}
\end{figure*}

\subsection{Pseudo-label Generation Strategy with an Instance Adaptive Selector}
\label{subsection:p}

Pseudo-labels $\hat{\mathbb{Y}}_{T}$ have a decisive effect on the quality of self-training. The generic pseudo-label generation strategy can be simplified to the following form when segmentation model parameter $\mathbf{w}$ is fixed:
\begin{equation}
\begin{aligned}
    \min \limits_{\hat{\mathbb{Y}}_{T}} & -  \frac{1}{\left | \mathbb{X}_{T} \right |} \sum_{\mathbf{x_{t}} \in \mathbb{X}_{T}} \sum_{c=1}^{C}\hat{y}_{t}^{(c)}
    \log\frac{p(c|\mathbf{x}_{t}, \mathbf{w})}
    {\theta_t^{(c)}}\\
    & s.t. \ \hat{\mathbf{y}}_{t} \in \{[onehot]^{C}\}\cup \mathbf{0} \ , \ \forall  \hat{\mathbf{y}}_{t} \in \hat{\mathbb{Y}}_{T},
\end{aligned}
\end{equation}
{where $\theta_t^{(c)}$ indicates the confidence threshold of class $c$ for instance $\mathbf{x}_t$, and $\mathbf{\hat{y}_{t}}=[\hat{y}_{t}^{(1)},...,\hat{y}_{t}^{(C)}]$ is required to be a one-hot vector or an all-zero vector. Therefore, $\hat{y}_{t}^{(c)}$ can be solved by Eq.\eqref{eq:y}.}
{\begin{equation}
\label{eq:y}
    \hat{y}_{t}^{(c)}=\left\{
\begin{array}{rl}
1,       &   if \ c=\mathop{\arg\max} \limits_{c}{{p(c|\mathbf{x}_{t},\mathbf{w})}}\ , \\
& \ \ \ \ \  and 
            \ p(c|\mathbf{x}_{t},\mathbf{w})> \theta_t^{(c)};\\
0,       & otherwise.       
\end{array} \right.
\end{equation}}

\vspace{-10pt}
For class $c$, when predicted probability $p(c|\mathbf{x}_{t},\mathbf{w})> \theta_t^{(c)}$, these pixels are regarded as confident regions (pseudo-label regions), and the rest are ignored regions (non-pseudo-label regions). Therefore, $\theta_t^{(c)}$ becomes the key to the pseudo-label generation process. As shown in Fig. \ref{fig:pseudo-methods_ct}, the traditional pseudo-label generation strategy based on a constant confidence threshold neglects differences between classes; the class-balanced threshold strategy designs a threshold ${\theta}^{(c)}$ for each class $c$, but it is still prone to ignore classes with lower predicted probabilities, as shown in Fig. \ref{fig:pseudo-methods_cbst}; unlike these two methods, we propose a data diversity-driven pseudo-label generation strategy with an instance adaptive selector (IAS), which can produce adaptive thresholds for each class, as shown in Fig. \ref{fig:pseudo-methods_iast}.

IAS maintains two thresholds $\{\boldsymbol{\theta}_t,\boldsymbol{\theta}_{\mathbf{x}_t} \}$, where $\boldsymbol{\theta}_t$ indicates the historical threshold and $\boldsymbol{\theta}_{\mathbf{x}_t}$ indicates the threshold of current instance $\mathbf{x}_t$. During the generation process, IAS dynamically updates  $\boldsymbol{\theta}_t$ based on  $\boldsymbol{\theta}_{\mathbf{x}_t}$, so each instance gets an adaptive threshold, combining global and local information. 
Specifically, for each class $c$ in an instance $\mathbf{x}_t$, we sort confidence probabilities of class $c$ in descending order, and then take the $(\alpha \times 100)$-th percentile as the local threshold $\theta_{\mathbf{x}_{t}}^{(c)}$ for class $c$ in instance $\mathbf{x}_t$. 
Finally, we use the exponentially weighted moving average strategy to update the threshold $\boldsymbol{\theta}_{t}$ for containing historical information as the global threshold. Consequently, the diversity of each category is improved by selecting pseudo-labels from every instance. The details are summarized in Algorithm \ref{alg:IAS} and will be described in the following subsections.

\begin{algorithm}[htb]
\caption{Pseudo-label generation}
\label{alg:IAS}
\begin{algorithmic}[1] 
\REQUIRE model $\mathbf{G}$, target instances $\{\mathbf{x}_{t}\}^{T}$, proportion $\alpha$, momentum $\beta$, weight decay $\gamma$.
\ENSURE target pseudo-labels $\{\hat{\mathbf{y}}_{t}\}^T$.
\STATE \textbf{init} $\boldsymbol{\theta}_{0}=[0.9]_{1 \times C}$
\FOR{$t=1$ \TO $T$} 
    \STATE $\mathbf{P}_{index}=\arg\max(\mathbf{G}(\mathbf{x}_{t}))$
    \STATE $\mathbf{P}_{value}=\max(\mathbf{G}(\mathbf{x}_{t}))$
    \FOR{$c=1$ \TO $C$}
        \STATE $\mathbb{P}_{ \mathbf{x}_t}^{(c)}=\text{sort}(\mathbf{P}_{value}[\mathbf{P}_{index}=c],\text{descending})$
        \STATE $\theta^{(c)}_{\mathbf{x}_{t}}=\Psi ( \mathbf{x}_{t},\theta _{t-1}^{(c)})$, Eq.\eqref{eq:wd}
    \ENDFOR
    \STATE $\boldsymbol{\theta} _{t} = \beta \cdot \boldsymbol{\theta} _{t-1}+(1-\beta)\cdot \boldsymbol{\theta} _{\mathbf{x}_{t}}$, Eq.\eqref{eq:ema}
    \STATE $\hat{\mathbf{y}}_{t}=\text{onehot}(\mathbf{P}_{index}[\mathbf{P}_{value}>\boldsymbol{\theta} _{t}])$
\ENDFOR
\RETURN $\{\hat{\mathbf{y}}_{t}\}^T$
\end{algorithmic}
\end{algorithm}

\subsubsection{Exponential moving average (EMA) threshold}
In the process of generating pseudo-labels, we introduce the EMA threshold to balance the diversity and noise ratio of selected pseudo-labels. If only the global information is employed, as in the approach of \cite{zou2018unsupervised} that utilizes the fixed threshold for each class, instances with overall low confidence will be ignored, thereby compromising diversity.
Conversely, if only the current instance information is used to generate the threshold, by selecting the pseudo-labels within the top $\alpha$\% of the prediction probabilities for each class, numerous noisy pseudo-labels will be selected for instances with overall low confidence.

Specifically, the EMA threshold strategy is shown in Eq.\eqref{eq:ema}, where $\theta _{t}^{(c)}$ represents the smoothed threshold. $\Psi (\mathbf{x}_{t},\theta_{t-1}^{(c)})$ represents the local threshold $\theta_{\mathbf{x}_{t}}^{(c)}$ of instance $\mathbf{x}_{t}$, which will be described in Section \ref{subsection: HWD}. $\beta$ is a momentum factor used to preserve past threshold information. As $\beta$ increases, the threshold $\theta _{t}^{(c)}$ becomes smoother.

\begin{equation}
\label{eq:ema}
    \theta _{t}^{(c)}  = \beta\cdot{\theta _{t-1}^{(c)}} +(1-\beta)\cdot{\Psi( \mathbf{x}_{t},\theta _{t-1}^{(c)})}.
\end{equation}

\subsubsection{Hard classes weight decay (HWD)}
\label{subsection: HWD}

Although the above threshold strategy can improve the diversity of pseudo-labels, more noise is also inevitably introduced, especially for the hard class. To tackle this issue, we design ${\theta_{t-1}^{(c)}}^{\gamma}$ to modify the proportion of pseudo-labels $\alpha$, as shown in Eq.\eqref{eq:wd}, where $\gamma $ is a weight decay parameter used to control the decay degree. 
Given that the noise mainly exists in the tail of each category in pseudo-labels where the predicted probability is relatively smaller, ${\theta_{t-1}^{(c)}}^{\gamma}$ can adaptively filter out a portion of these regions, thus alleviating the noise. 
It should be noted that the thresholds ${\theta_{t-1}^{(c)}} $ of hard classes are usually smaller, so HWD filters out more tail areas for suppressing the noise; on the contrary, the thresholds ${\theta _{t-1}^{(c)}}$ of easy classes is usually larger, so HWD has a weaker impact.

\begin{equation}
\label{eq:wd}
    \Psi (\mathbf{x}_{t},\theta _{t-1}^{(c)}) = \mathbb{P}_{ \mathbf{x}_t}^{(c)}\left [  \alpha  {\theta _{t-1}^{(c)}}^{\gamma}  | \mathbb{P}_{\mathbf{x}_t}^{(c)}  | \right ],
\end{equation}
\added{where $\left [ \cdot \right ] $ indicates the indexing operation, and $\left | \cdot \right | $ denotes the number of elements in the list $\mathbb{P}_{\mathbf{x}_t}^{(c)}$.} 

\subsection{Hard-aware Pseudo-label Augmentation}
\label{subsection: HPLA}

Because hard classes always have smaller proportions, the trained model is prone to bias to the predominant easy classes. Furthermore, due to the interference from noise, the high-quality pseudo-label proportions of hard classes become lower, making the above problem even worse. To alleviate this issue, we design an adaptively hard-aware pseudo-label augmentation in the target domain.

Specifically, {we first detect hard classes with the thresholds after pseudo-label generation, and then perform pseudo-label augmentation for detected hard classes. 

In the hard class detection, we propose to use Eq.\eqref{eq:c_h} to find the $k$ hard classes $\mathbf{C}_{h}$. The thresholds ${\boldsymbol{\theta}=[\theta^{(1)}, ..., \theta^{(C)}]}$ of all classes are sorted by ascending, and {the top $k$ classes with lower threshold are selected, which are regarded as hard classes.
\begin{equation}
\label{eq:c_h}
\mathbf{C}_{h}={\Theta}[:, k], \text{where } {\Theta}=\text{sort}(\boldsymbol{\theta},\text{ascending}).
\end{equation}

Different from other Copy-Paste methods\cite{DACS_Tranheden_2021_WACV,CAMix_zhou2021context} where classes are equally selected, our method assigns higher sampling probability to hard classes. To this end, the sampling probability $\boldsymbol{\boldsymbol{r}}=[r^{(1)}, ...,r^{(C)}]$ is formulated as Eq.\eqref{eq:r} and it means that the smaller the threshold $\theta^{(c)}$, the larger the sampling probability $r^{(c)}$.

{\begin{equation}
\label{eq:r}
r^{(c)}=\frac{1-\theta^{(c)}}{\sum_{i=1}^{C}(1-\theta^{(i)})}.
\end{equation}}

In the pseudo-label augmentation, for each image $\mathbf{x}_t$ in $\mathbb{X}_{T}$, we  first randomly select a class ${c}$ according to the sampling probability $\boldsymbol{\boldsymbol{r}}$. Then, an image $\mathbf{x}_i$ containing class ${c}$ is randomly chosen from the target domain images. Finally, we copy pixels of $k$ selected classes $\mathbf{C}_{h}$ in $\mathbf{x}_i$ and paste them onto $\mathbf{x}_t$. This operation is also synchronously performed to the corresponding pseudo-label $\mathbf{\hat{y}}_t$ in $\hat{\mathbb{Y}}_{T}$. The details are summarized in Algorithm \ref{alg:HPLA}, and we show an example of pseudo-label augmentation in Fig. \ref{fig:cp}. More examples are provided in Appendix C of our supplementary.

\begin{algorithm}[htb]
\caption{Hard-aware pseudo-label augmentation}
\label{alg:HPLA}
\begin{algorithmic}[1]
\REQUIRE instances $\{\mathbf{x}_{t}\}^{T}$, pseudo-labels $\{\hat{\mathbf{y}}_{t}\}^{T}$, thresholds $\boldsymbol{\theta}$, sampling probability $\boldsymbol{r}$, number of hard classes $k$.
\ENSURE augmented $\{\mathbf{x}_{t}^{a}\}^{T}$ and $\{\hat{\mathbf{y}}_{t}^{a}\}^{T}$.
\STATE $\mathbf{C}_{h}=\{\text{sort}({\boldsymbol{\theta}}, \text{ascending})[:, k]\}$, Eq.\eqref{eq:c_h}
\FOR{$t=1$ \TO $T$} 
    \STATE $\mathbf{c}=\text{RandomSelect}(\boldsymbol{r})$
    \STATE $\mathbf{x}_{i}, \hat{\mathbf{y}}_{i}=\text{RandomSelect}(\{\mathbf{x}_{t}\}^{T}, \{\hat{\mathbf{y}}_{t}\}^{T},\mathbf{c})$
    \STATE $\mathbf{x}_{t}^{a}, \hat{\mathbf{y}}_{t}^{a}=\text{CopyPaste}(\mathbf{x}_{t}, \hat{\mathbf{y}}_{t},\mathbf{x}_{i}, \hat{\mathbf{y}}_{i}, \mathbf{C}_{h})$, Fig. \ref{fig:cp}
\ENDFOR
\RETURN $\{\mathbf{x}_{t}^{a}\}^{T}, \{\hat{\mathbf{y}}_{t}^{a}\}^{T}$
\end{algorithmic}
\end{algorithm}

\begin{figure}[htp]     
\centering 
\includegraphics[width=1.0\linewidth]{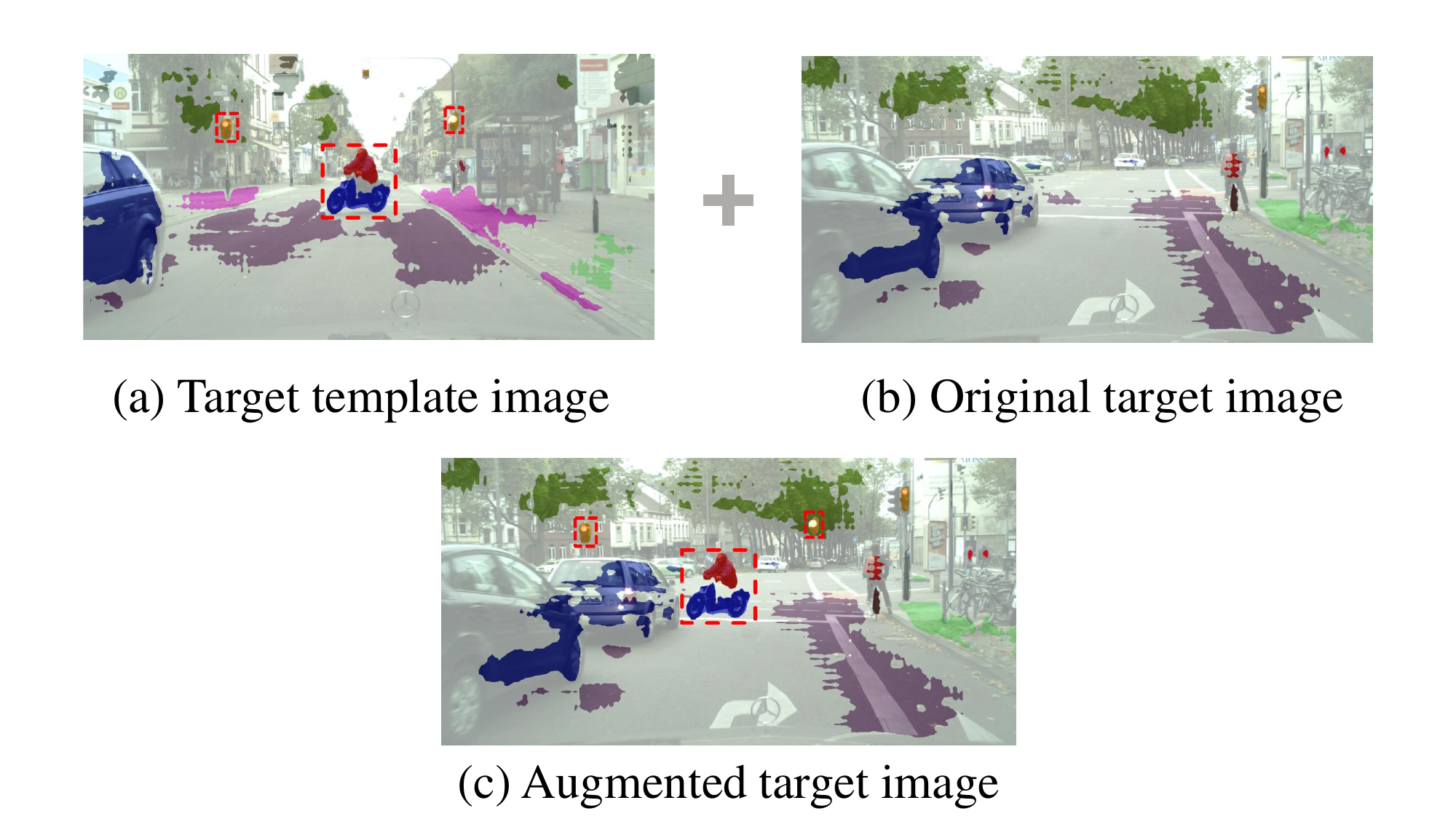}
\vspace{-10.5pt}
\caption{Proposed hard-aware pseudo-label augmentation. (a): The randomly selected target domain images for copying pixels of hard classes. (b): The original target image in the training batch. (c): The result of HPLA, which has copied the data of traffic light, rider, and motorcycle from (a), and thus the diversity of hard classes in pseudo-labels can be further enriched.}
\label{fig:cp} 
\end{figure}

\subsection{Region-adaptive Constraints}
\label{subsection: r}
\subsubsection{Confident region Kullback-Leibler divergence (KLD) minimization}

For the confident region $\mathbb{I}_{\mathbf{x}_{t}}=\{\mathbf{1}\ | \ \mathbf{\hat{y}}_{t}^{(h,w)}>\mathbf{0} \}$, there are pseudo-labels as supervision signals to supervise the model training. However, the noise is inevitably introduced during pseudo-label generation, and being overconfident with pseudo-labels will misguide model training, thus ruining the domain adaptation. How to reduce the impact of noise becomes a key issue. An available way is to smooth model outputs by fitting a uniform distribution, thus avoiding overfitting pseudo-labels \cite{zou2019confidence}. Hence, we introduce the Kullback-Leibler divergence (KLD) as a regularization term and deploy it on the confident region:
\begin{equation} \label{eq:kld}
    \begin{aligned}
        \mathcal{R}_{c}= - \frac{1}{\left | \mathbb{X}_{T} \right |} \sum_{\mathbf{x}_{t} \in \mathbb{X}_{T}}\mathbb{I}_{\mathbf{x}_{t}} \sum_{c=1}^{C}\frac{1}{C}\log p(c|\mathbf{x}_{t}, \mathbf{w}).
 \end{aligned}
\end{equation}

As shown in Eq.\eqref{eq:kld}, when the predicted result $\log p(c|\mathbf{x}_{t}, \mathbf{w})$ is approximately close to the uniform distribution (the probability of each class is $ \frac{1}{C} $), $\mathcal{R}_{c}$ gets smaller. KLD minimization promotes the smoothing of confident regions and avoids the model blindly trusting pseudo-labels.

\subsubsection{Ignored region entropy minimization}
On the other hand, for the ignored region $\mathbb{I}_{\mathbf{x}_{t}}^{\complement}=\{\mathbf{1}\ | \ \mathbf{\hat{y}}_{t}^{(h,w)}=\mathbf{0} \}$, there is no supervision signal during the training process. Because the predicted result of region $\mathbb{I}_{\mathbf{x}_{t}}^{\complement}$ is smooth and has low confidence, we minimize the entropy of the ignored region to prompt the model to predict low entropy results, which makes the prediction result look sharper.
\begin{equation}
\label{eq:ent}
\begin{aligned}
    \mathcal{R}_{i}= - \frac{1}{\left | \mathbb{X}_{T} \right |} \sum_{\mathbf{x}_{t} \in \mathbb{X}_{T}}\mathbb{I}_{\mathbf{x}_{t}}^{\complement} \sum_{c=1}^{C}p(c|\mathbf{x}_{t}, \mathbf{w})\log p(c|\mathbf{x}_{t}, \mathbf{w}).
\end{aligned}
\end{equation}

As shown in Eq.\eqref{eq:ent}, sharpening the predicted result of the ignored region by minimizing $\mathcal{R}_{i}$ can promote the model to learn more useful features from the ignored region without any supervision signal, which has also been proved to be effective for UDA in the work \cite{vu2019advent}.

\begin{figure*}[htb]
\centering
{
\includegraphics[width=0.9\linewidth]{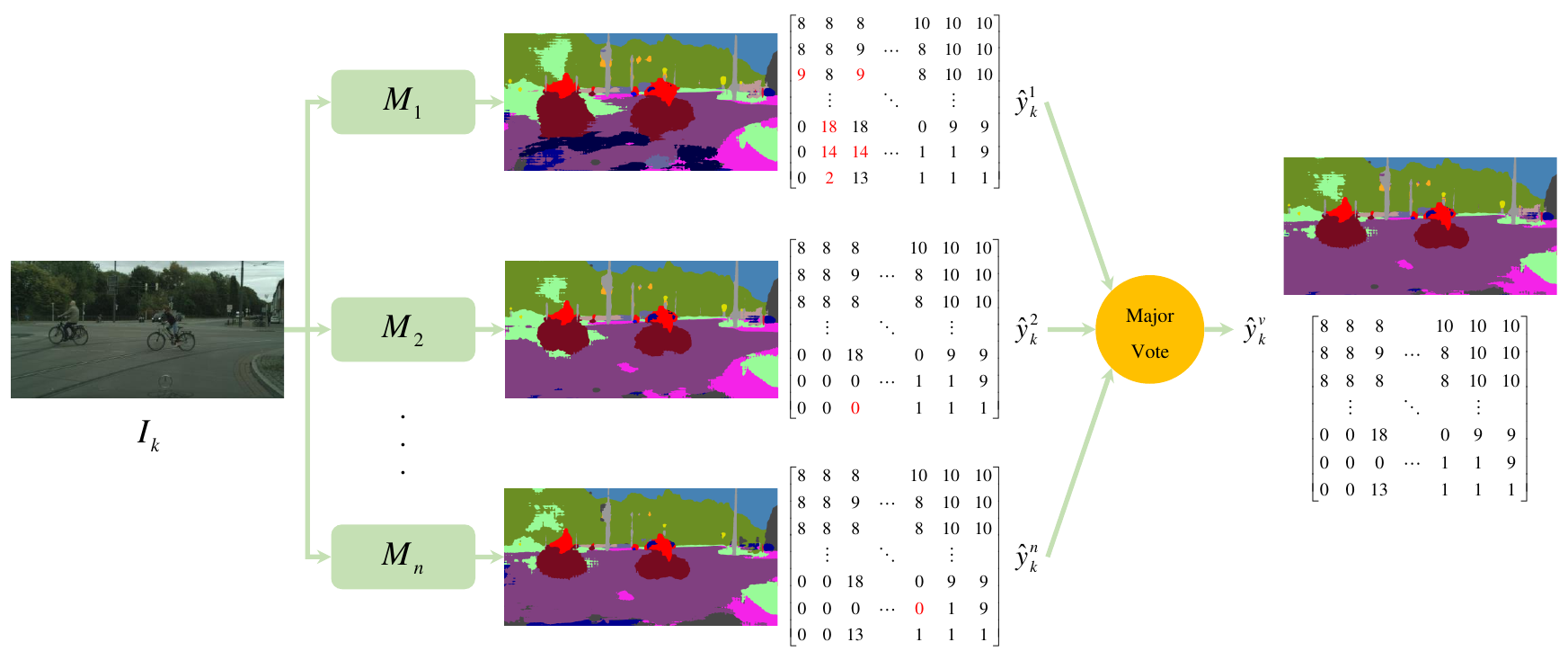}
}
\vspace{-4pt}
\caption{{The specific voting process is as follows. For a given target image $I_k$, the models $[M_{1}, ..., M_{n}]$ trained with different parameter values will generate the pseudo-labels $[\hat{y}_k^{1},...,\hat{y}_k^{n}]$ for $I_k$. Then, at each pixel position, the majority voting method is adopted to obtain the corresponding class, thereby obtaining the fused pseudo-label $\hat{y}_k^v$. The red numbers indicate the values that are different from the $\hat{y}_k^v$ for convenience.}}
\vspace{0pt}
\label{fig:vote}
\end{figure*}

\subsubsection{Ignored region consistency constraint}

{We further utilize consistency training based on the widely spread Mean-Teacher\cite{MeanTeacher_Tarvainen2017MeanTA} framework to promote model training on the ignored region. As shown in Fig. \ref{fig:framework_self_training}, it contains a momentum model $\overline{\mathbf M}$, which is the copy of $\mathbf M$ and slowly updated by EMA strategy with parameter $\tau_{\overline{\mathbf{w}}}$ after each iteration, as shown in Eq.\eqref{eq:momentum_model}, where $\overline{\mathbf{w}}$ indicates the weights of $\overline{\mathbf{M}}$. With the assistance of EMA, $\mathbf M$ can be averaged at multiple moments, thus providing more stable predictions for the consistency training.{ Our method is performed on the ignored region, while ProDA\cite{ProDA_zhang2021prototypical} is conducted on the whole image. In addition, our method is applied in pixel-level, which is more concise and general without relying on the prototypes.}}

Firstly, one target domain image $\mathbf{x}_t^a$ processed by HPLA is perturbed by weak and strong augmentation to get $\mathbf{x}_t^{wa}$ and $\mathbf{x}_t^{sa}$ respectively. 
Then, $\mathbf{x}_t^{sa}$ is fed into $\mathbf{M}$ to get prediction $\mathbf{p}(\mathbf{x}_{t}^{sa}, \mathbf{w})$, and $\mathbf{x}_t^{wa}$ is fed into $\mathbf{\overline{M}}$ to obtain prediction $\mathbf{p}(\mathbf{x}_{t}^{wa}, \mathbf{\overline{w}})$. Finally, soft cross-entropy loss as shown in Eq.\eqref{eq:consistency_loss} is calculated on ignored regions to force $\mathbf{p}(\mathbf{x}_{t}^{sa}, \mathbf{w})$ consistent with $\mathbf{p}(\mathbf{x}_{t}^{wa}, \mathbf{\overline{w}})$.

\begin{equation}
\label{eq:momentum_model}
    \mathbf{\overline{w}}_{t}=
    \tau_{\overline{\mathbf{w}}} \cdot \mathbf{\overline{w}}_{t-1} + 
    (1-\tau_{\overline{\mathbf{w}}})\cdot \mathbf{w}_{t}.
\end{equation}

\begin{equation}
\label{eq:consistency_loss}
    \mathcal{R}_{cst}= - \frac{1}{\left | \mathbb{X}_{T} \right |} 
    \sum_{\mathbf{x}_{t} \in \mathbb{X}_{T}}\mathbb{I}_{\mathbf{x}_{t}}^{\complement} 
    \sum_{c=1}^{C}
    p(c|\mathbf{x}_{t}^{wa}, \mathbf{\overline{w}})
    \log p(c|\mathbf{x}_{t}^{sa}, \mathbf{w}).
\end{equation}

In some sense, $\overline{\mathbf{M}}$ is the ensemble of $\mathbf{M}$ at multiple moments, which can provide more stable predictions, therefore $\mathbf{\overline{M}}$ is used as $\mathbf{G}$ at the end of each self-training round to generate pseudo-labels for the next round.

\subsection{Parameter Selection}
\label{subsection: param_select}

In the setting of UDA tasks, we can only obtain images of the target domain training set. However, most of the existing methods select parameters on the validation set of the target domain, which actually violates the setting of UDA. Therefore, in this paper, we have designed a novel parameter selection algorithm that only utilizes the images of the target domain training set, thus being more in line with the setting of UDA tasks.

The detailed parameter selection is as follows.
Firstly, we randomly select 500 images of the target domain training set to obtain $I_{subset}$.
Next, the discrete candidate set (such as $\boldsymbol{\lambda}=[\lambda_{1}, ..., \lambda_{n}]$) is determined by interval sampling for each hyperparameter, and we need to select the best parameter $\lambda_{best}$ from $n$ 
candidate parameters. Then, we utilize the HIAST framework to perform training with each candidate parameter ${\lambda_i}$ ($i$ from 1 to $n$), and thus obtain the corresponding model $M_i$. After that, we use the obtained model $M_i$ to generate the prediction $\hat{y}_k^{i}$ for image $I_k$ in the ${I}_{subset}$. Then, the prediction results $\mathbf{\hat{Y}_k}=[\hat{y}_k^{1},...,\hat{y}_k^{n}]$ can be generated for $I_k$ with $n$ candidate models $[M_{1}, ..., M_{n}]$. Subsequently, the fusion pseudo-label $\hat{\mathbf{y}}^{v}_{k}$ is formed through a major voting fusion scheme, as shown in Fig. \ref{fig:vote}. In this paper, the $\hat{\mathbf{y}}^{v}_{k}$ is treated as the estimation of the ground-truth in the target domain. The detailed process is shown in the Algorithm \ref{alg:template_GT}. 
Finally, after obtaining $\{\hat{\mathbf{y}}^{v}_{k} \}_{k=1}^{500}$ of all images in $I_{subset}$, the mIoU between $\{\hat{y}_k^{i}\}_{k=1}^{500}$ and the fusion $\{\hat{\mathbf{y}}^{v}_{k}\}_{k=1}^{500}$ (F-mIoU) is calculated, evaluating the performance of each model $M_i$. 
We select the parameter with the highest F-mIoU score as the optimal parameter $\lambda_{best}$. 

\begin{table*}[htb]
\centering
\caption{Details of datasets for UDA semantic segmentation.}
\vspace{-7.5pt}
\label{table:dataset_details}
\begin{tabular}{l|cccc}
\toprule
Dataset & GTA5\cite{richter2016playing} & SYNTHIA\cite{ros2016synthia} & Cityscapes\cite{cordts2016cityscapes} & Oxford RobotCar\cite{RobotCarDatasetIJRR} \\
\midrule
Resolution & $1914 \times 1052$ & $1280 \times 760$ & $2048 \times 1024$ & $1280 \times 960$ \\
Number of images for training & 24966 & 9400 & 2975 & 894 \\
Number of images for evaluation & - & - & 500 & 271 \\
Number of categories & 19 & 16 & 19 & 9 \\
Is synthetic? & \ding{51} & \ding{51} & \ding{53} & \ding{53} \\
\bottomrule
\end{tabular}
\end{table*}

\begin{table*}[hbt]
\centering
\caption{{Results of our HIAST and other SOTA methods (GTA5 $\rightarrow$ Cityscapes). The default warm-up model of IAST and HIAST is AdaptSeg, and $\nabla$ indicates that SePiCo is used for warm-up.}}
\label{table:gta5}
\vspace{-7.5pt}
\resizebox{1\linewidth}{!}{ 
\renewcommand\arraystretch{1.3} %
\begin{tabular}{c|l|ccccccccccccccccccc|c}
\toprule
Year & Method & \rotatebox{90}{Road} & \rotatebox{90}{SW} & \rotatebox{90}{Build} & \rotatebox{90}{Wall} & \rotatebox{90}{Fence} & \rotatebox{90}{Pole} & \rotatebox{90}{TL} & \rotatebox{90}{TS} & \rotatebox{90}{Veg.} & \rotatebox{90}{Terrain} & \rotatebox{90}{Sky} & \rotatebox{90}{PR} & \rotatebox{90}{Rider} & \rotatebox{90}{Car} & \rotatebox{90}{Truck} & \rotatebox{90}{Bus} & \rotatebox{90}{Train} & \rotatebox{90}{Motor} & \rotatebox{90}{Bike} & mIoU \\
\midrule
2018 & AdaptSeg\cite{tsai2018learning} &  86.5 & 36.0 & 79.9 & 23.4 & 23.3 & 23.9 & 35.2 & 14.8 & 83.4 & 33.3 & 75.6 & 58.5 & 27.6 & 73.7 & 32.5 & 35.4 & 3.9 & 30.1 & 28.1 & 42.4 \\
2019 & CLAN\cite{luo2019taking} & 87.0 & 27.1 & 79.6 & 27.3 & 23.3 & 28.3 & 35.5 & 24.2 & 83.6 & 27.4 & 74.2 & 58.6 & 28.0 & 76.2 & 33.1 & 36.7 & 6.7 & 31.9 & 31.4 & 43.2 \\
2019 & AdvEnt+MinEnt\cite{vu2019advent}  & 89.4 & 33.1 & 81.0 & 26.6 & 26.8 & 27.2 & 33.5 & 24.7 & 83.9 & 36.7 & 78.8 & 58.7 & 30.5 & 84.8 & 38.5 & 44.5 & 1.7 & 31.6 & 32.4 & 45.5 \\
2018 & CBST\cite{zou2018unsupervised} & 91.8 & 53.5 & 80.5 & 32.7 & 21.0 & 34.0 & 28.9 & 20.4 & 83.9 & 34.2 & 80.9 & 53.1 & 24.0 & 82.7 & 30.3 & 35.9 & 16.0 & 25.9 & 42.8 & 45.9 \\
2020 & MRNet+Pseudo\cite{zheng2019unsupervised} & 90.5 & 35.0 & 84.6 & 34.3 & 24.0 & 36.8 & 44.1 & 42.7 & 84.5 & 33.6 & 82.5 & 63.1 & 34.4 & 85.8 & 32.9 & 38.2 & 2.0 & 27.1 & 41.8 & 48.3 \\
2020 & FADA\cite{Haoran_2020_ECCV} & 91.0 & 50.6 & 86.0 & {43.4} & 29.8 & 36.8 & 43.4 & 25.0 & 86.8 & 38.3 & 87.4 & 64.0 & 38.0 & 85.2 & 31.6 & 46.1 & 6.5 & 25.4 & 37.1 & 50.1 \\
2019 & CAG-UDA\cite{zhang2019category} & 90.4 & 51.6 & 83.8 & 34.2 & 27.8 & 38.4 & 25.3 & 48.4 & 85.4 & 38.2 & 78.1 & 58.6 & 34.6 & 84.7 & 21.9 & 42.7 & \textbf{41.1} & 29.3 & 37.2 & 50.2 \\
2021 & CDGA\cite{CDGA_Kim_Joung_Kim_Park_Kim_Sohn_2021} & 91.1 & 52.8 & 84.6 & 32.0 & 27.1 & 33.8 & 38.4 & 40.3 & 84.6 & 42.8 & 85.0 & 64.2 & 36.5 & 87.3 & 44.4 & 51.0 & 0.0 & 37.3 & 44.9 & 51.5 \\
2021 & DACS\cite{DACS_Tranheden_2021_WACV} & 89.9 & 39.7 & {87.9} & 30.7 & {39.5} & 38.5 & 46.4 & {52.8} & {88.0} & 44.0 & 88.8 & 67.2 & 35.8 & 84.5 & 45.7 & 50.2 & 0.0 & 27.3 & 34.0 & 52.1 \\
2021 & SAC\cite{SAC_araslanov2021self} & 90.4 & 53.9 & 86.6 & 42.4 & 27.3 & {45.1} & 48.5 & 42.7 & 87.4 & 40.1 & 86.1 & 67.5 & 29.7 & 88.5 & {49.1} & 54.6 & 9.8 & 26.6 & 45.3 & 53.8 \\
2021 & DSP\cite{DSP_gao2021dsp} & 92.4 & 48.0 & 87.4 & 33.4 & 35.1 & 36.4 & 41.6 & 46.0 & {87.7} & 43.2 & {89.8} & 66.6 & 32.1 & {89.9} & {57.0} & 56.1 & 0.0 & 44.1 & 57.8 & 55.0 \\
2021 & CAMix\cite{CAMix_zhou2021context} & 93.3 & 58.2 & 86.5 & 36.8 & 31.5 & 36.4 & 35.0 & 43.5 & 87.2 & {44.6} & 88.1 & 65.0 & 24.7 & {89.7} & 46.9 & 56.8 & 27.5 & 41.1 & 56.0 & 55.2 \\
2021 & ProDA\cite{ProDA_zhang2021prototypical} & 87.8 & 56.0 & 79.7 & \textbf{46.3} & 44.8 & 45.6 & 53.5 & 53.5 & 88.6 & 45.2 & 82.1 & 70.7 & 39.2 & 88.8 & 45.5 & 59.4 & 1.0 & 48.9 & 56.4 & 57.5 \\
2022 & CPSL\cite{li2022class} & 92.3 & 59.9 & 84.9 & 45.7 & 29.7 & \textbf{52.8} & \textbf{61.5} & 59.5 & 87.9 & 41.5 & 85.0 & 73.0 & 35.5 & 90.4 & 48.7 & 73.9 & 26.3 & 53.8 & 53.9 & 60.8 \\
2022 & SePiCo\cite{xie2022sepico} & 95.2 & 67.8 & 88.7 & 41.4 & 38.4 & 43.4 & 55.5 & 63.2 & 88.6 & \textbf{46.4} & 88.3 & 73.1 & 49.0 & 91.4 & \textbf{63.2} & 60.4 & 0.0 & 45.2 & 60.0 & 61.0 \\
2023 &
{FREDOM\cite{truong2023fredom}} & 90.9 & 54.1 & 87.8 & 44.1 & 32.6 & 45.2 & 51.4 & 57.1 & 88.6 & 42.6 & 89.5 & 68.8 & 40.0 & 89.7 & 58.4 & 62.6 & 55.3 & 47.7 & 58.1 & 61.3 \\
2023 &
{RTea\cite{zhao2023learning}}  & 95.4 & 67.1 & 87.9 & 46.1 & 44.0 & 46.0 & 53.8 & 59.5 & \textbf{89.7} & 49.8 & 89.8 & 71.5 & 40.5 & 90.8 & 55.0 & 57.9 & 22.1 & 47.7 & 62.5 & 61.9 \\
2022 & DDB\cite{chen2022deliberated} & 95.3 & 67.4 & 89.3 & 44.4 & \textbf{45.7} & 38.7 & 54.7 & 55.7 & 88.1 & 40.7 & \textbf{90.7} & 70.7 & 43.1 & \textbf{92.2} & 60.8 & \textbf{67.6} & 34.2 & 48.7 & 63.7 & 62.7 \\	
2024 &
{RDASS-KD\cite{10766055}}  & 95.1 & 64.0 & 89.7 & 50.4 & 46.3 & 50.9 & 61.1 & 62.4 & 88.9 & 51.6 & 87.7 & 73.0 & 39.4 & 91.8 & 67.8 & 67.0 & 0.0 & 51.7 & 64.5 & 63.3 \\
\midrule
2020 & {IAST} (Ours) & 93.8 & 57.8 & 85.1 & 39.5 & 26.7 & 26.2 & 43.1 & 34.7 & 84.9 & 32.9 & 88.0 & 62.6 & 29.0 & 87.3 & 39.2 & 49.6 & 23.2 & 34.7 & 39.6 & 51.5 \\
 & {HIAST} (Ours) & 94.4 & 63.4 & 87.0 & 43.1 & 31.3 & 35.6 & 50.1 & 37.4 & 86.0 & 29.9 & 88.3 & 68.8 & 34.1 & 87.3 & 41.7 & 43.2 & 37.0 & 50.9 & 60.0 & 56.3 \\
 & {HIAST} (Ours)$^{\nabla}$ & \textbf{95.8} & \textbf{69.3} & \textbf{89.8} & 44.6 & 40.3 & 49.2 & 61.4 & \textbf{67.8} & {89.4} & 43.7 & 90.0 & \textbf{76.0} & \textbf{53.2} & 92.4 & 62.0 & 67.2 & 0.0 & \textbf{58.1} & \textbf{68.0} & \textbf{64.1} \\
 
\bottomrule
\end{tabular}
}
\vspace{4pt}
\end{table*}

\begin{algorithm}[htb]
\caption{{Fusion pseudo-label generation}}
\label{alg:template_GT}
\begin{algorithmic}[1]
\REQUIRE $\boldsymbol{\lambda}=[\lambda_{1}, ..., \lambda_{n}]$, the dataset for parameter adjustment ${I}_{subset}$, instances $\{\mathbf{x}_{t}\}^{T}$, pseudo-labels $\{\hat{\mathbf{y}}_{t}\}^{T}$.
\ENSURE fusion pseudo-labels $\{\hat{\mathbf{y}}^{v}_{k} \}_{k=1}^{500}$ for ${I}_{subset}$.
\FOR{$i=1$ \TO $n$} 
    \STATE ${M_i}=Train(\{\mathbf{x}_{t}\}^{T}, \{\hat{\mathbf{y}}_{t}\}^{T}, \lambda_{i})$
\ENDFOR
\FOR{$k=1$ \TO $500$}
    \STATE $\mathbf{\hat{Y}}_k=[\ \ ]$
    \FOR{$i=1$ \TO $n$}
        \STATE  $\hat{y}_k^i={M_i}(I_k)$
        \STATE $\mathbf{\hat{Y}}_k.append(\hat{y}_k^i)$
    \ENDFOR
    \STATE $\hat{\mathbf{y}}^{v}_{k}=vote$($\mathbf{\hat{Y}}_k$), Fig. \ref{fig:vote}
\ENDFOR
\RETURN $\{\hat{\mathbf{y}}^{v}_{k} \}_{k=1}^{500}$
\end{algorithmic}
\end{algorithm}

\section{Experiment}
\label{sec:experiment}
\subsection{Experimental Settings}

\noindent\textbf{Network Architecture.} Following recent works\cite{tsai2018learning, vu2019advent, Tsai_adaptseg_ICCV19, zou2018unsupervised, zou2019confidence, MetaCorrection_guo2021metacorrection, SAC_araslanov2021self, DACS_Tranheden_2021_WACV, ContenTransfer_Lee2021UnsupervisedDA}, we adopt widely used DeepLab-V2\cite{chen2017deeplab} with ASPP for UDA semantic segmentation, and ResNet-101\cite{he2016deep} pre-trained on ImageNet\cite{deng2009imagenet} is selected as the backbone. All experiments in this work are carried out under this architecture.

\noindent\textbf{Datasets and Metric.} 
We evaluate our UDA method for semantic segmentation on two popular synthetic-to-real scenarios: (a) GTA5 \cite{richter2016playing} $\rightarrow$  Cityscapes \cite{cordts2016cityscapes}, (b) SYNTHIA \cite{ros2016synthia} $\rightarrow$ Cityscapes, and one cross-city scenario: (c) Cityscapes $\rightarrow$ Oxford RobotCar\cite{RobotCarDatasetIJRR}. GTA5 and SYNTHIA datasets are rendered by the virtual engine, while Cityscapes and Oxford RobotCar datasets consist of real images of the street view. Following the standard protocols in \cite{tsai2018learning} for synthetic-to-real adaptation, the Cityscapes dataset is utilized as the unlabeled target domain; similarly for cross-city adaptation following \cite{Tsai_adaptseg_ICCV19, Zheng2021RectifyingPL}, the Oxford RobotCar dataset serves as the unlabeled target domain; our UDA method is evaluated on the validation dataset. All aforementioned datasets have shared categories, and the details are listed in Table \ref{table:dataset_details}. We use mIoU as the metric for evaluation in all experiments.

\begin{table*}[htb]
\centering
\caption{{Results of our HIAST and other SOTA methods (SYNTHIA $\rightarrow$ Cityscapes). The default warm-up model of IAST and HIAST is AdaptSeg, and $\nabla$ indicates that SePiCo is used for warm-up.}}
\label{table:synthia}
\vspace{-7.5pt}
\resizebox{1.0\linewidth}{!}{
\renewcommand\arraystretch{1.3}
\begin{tabular}{c|l|cccccccccccccccc|cc}
\toprule
Year & Method    &               \rotatebox{90}{Road} & \rotatebox{90}{SW}   & \rotatebox{90}{Build} & \rotatebox{90}{Wall*} & \rotatebox{90}{Fence*} & \rotatebox{90}{Pole*}  & \rotatebox{90}{TL}    &  \rotatebox{90}{TS}   & \rotatebox{90}{Veg.} & \rotatebox{90}{Sky}  & \rotatebox{90}{PR}   & \rotatebox{90}{Rider} & \rotatebox{90}{Car}  & \rotatebox{90}{Bus}  & \rotatebox{90}{Motor} & \rotatebox{90}{Bike} & mIoU & mIoU* \\
\midrule
2018 & AdaptSeg\cite{tsai2018learning}  & 84.3 & 42.7 & 77.5 & - & - & - & 4.7 & 7.0 & 77.9 & 82.5 & 54.3 & 21.0 & 72.3 & 32.2 & 18.9 & 32.3 & - & 46.7 \\
2019 & CLAN\cite{luo2019taking} &  81.3 & 37.0 & 80.1 & - & - & - & 16.1 & 13.7 & 78.2 & 81.5 & 53.4 & 21.2 & 73.0 & 32.9 & 22.6 & 30.7 & - & 47.8 \\
2019 & AdvEnt+MinEnt\cite{vu2019advent} &  85.6 & 42.2 & 79.7 & 8.7 & 0.4 & 25.9 & 5.4 & 8.1 & 80.4 & 84.1 & 57.9 & 23.8 & 73.3 & 36.4 & 14.2 & 33.0 & 41.2 & 48.0 \\
2018 & CBST\cite{zou2018unsupervised} & 68.0 & 29.9 & 76.3 & 10.8 & 1.4 & 33.9 & 22.8 & 29.5 & 77.6 & 78.3 & 60.6 & 28.3 & 81.6 & 23.5 & 18.8 & 39.8 & 42.6 & 48.9 \\
2020 & FADA\cite{Haoran_2020_ECCV} & 84.5 & 40.1 & 83.1 & 4.8 & 0.0 & 34.3 & 20.1 & 27.2 & 84.8 & 84.0 & 53.5 & 22.6 & 85.4 & 43.7 & 26.8 & 27.8 & 45.2 & 52.5 \\
2019 & CAG-UDA\cite{zhang2019category} & 84.7 & 40.8 & 81.7 & 7.8 & 0.0 & 35.1 & 13.3 & 22.7 & 84.5 & 77.6 & 64.2 & 27.8 & 80.9 & 19.7 & 22.7 & 48.3 & 44.5 & 52.6 \\
2021 & CDGA\cite{CDGA_Kim_Joung_Kim_Park_Kim_Sohn_2021} & 90.7 & 49.5 & 84.5 & - & - & - & 33.6 & {38.9} & 84.6 & 84.6 & 59.8 & 33.3 & 80.8 & 51.5 & 37.6 & 45.9 & - & 54.1 \\
2021 & DACS\cite{DACS_Tranheden_2021_WACV} & 80.6 & 25.1 & 81.9 & 21.5 & {2.9} & 37.2 & 22.7 & 24.0 & 83.7 & \textbf{90.8} & 67.6 & {38.3} & 82.9 & 38.9 & 28.5 & 47.6 & 48.4 & 54.8 \\
2021 & SAC\cite{SAC_araslanov2021self} & 89.3 & 47.2 & {85.5} & {26.5} & 1.3 & {43.0} & {45.5} & 32.0 & {87.1} & {89.3} & 63.6 & 25.4 & 86.9 & 35.6 & 30.4 & 53.0 & 52.6 & 59.3 \\
2021 & DSP\cite{DSP_gao2021dsp} & 86.4 & 42.0 & 82.0 & 2.1 & 1.8 & 34.0 & 31.6 & 33.2 & {87.2} & 88.5 & 64.1 & 31.9 & 83.8 & {65.4} & 28.8 & 54.0 & 51.0 & 59.9 \\
2021 & CAMix\cite{CAMix_zhou2021context} & \textbf{91.8} & \textbf{54.9} & 83.6 & - & - & - & 23.0 & 29.0 & 83.8 & 87.1 & 65.0 & 26.4 & 85.5 & 55.1 & 36.8 & 54.1 & - & 59.7 \\
2021 & ProDA\cite{ProDA_zhang2021prototypical} & 87.8 & 45.7 & 84.6 & 37.1 & 0.6 & 44.0 & 54.6 & 37.0 & \textbf{88.1} & 84.4 & 74.2 & 24.3 & 88.2 & 51.1 & 40.5 & 45.6 & 55.0 & 62.0 \\
2022 & CPSL\cite{li2022class} & 87.2 & 43.9 & 85.5 & \textbf{33.6} & 0.3 & \textbf{47.7} & \textbf{57.4} & 37.2 & 87.8 & 88.5 & \textbf{79.0} & 32.0 & \textbf{90.6} & 49.4 & 50.8 & 59.8 & 57.9 & 65.3 \\
2023 & RTea\cite{zhao2023learning} & 93.2 & 59.6 & 86.3 & 31.3 & 4.8 & 43.1 & 41.8 & 44.0 & 88.6 & 90.5 & 70.4 & 42.6 & 89.5 & 56.7 & 40.2 & 59.9 & 58.9 & 66.4\\
2022 & SePiCo\cite{xie2022sepico} & 77.0 & 35.3 & 85.1 & 23.9 & 3.4 & 38.0 & 51.0 & 55.1 & 85.6 & 80.5 & 73.5 & 46.3 & 87.6 & 69.7 & 50.9 & 66.5 & 58.1 & 66.5 \\
2024 & RDASS-KD\cite{10766055} & 85.0 & 51.3 & 80.3 & 16.3 & 1.7 & 47.6 & 52.4 & 40.9 & 88.3 & 90.9 & 73.8 & 31.5 & 90.4 & 71.8 & 52.8 & 63.3 & 58.6 & 67.1 \\
2023 & FREDOM\cite{truong2023fredom} & 86.0 & 46.3 & 87.0 & 33.3 & 5.3 & 48.7 & 53.4 & 46.8 & 87.1 & 89.1 & 71.2 & 38.1 & 87.1 & 54.6 & 51.3 & 59.9 & 59.1 & 66.0 \\

\midrule
2020 & IAST (Ours) & 81.9 & 41.5 & 83.3 & 17.7 & \textbf{4.6} & 32.3 & 30.9 & 28.8 & 83.4 & 85.0 & 65.5 & 30.8 & 86.5 & 38.2 & 33.1 & 52.7 & 49.8 & 57.0 \\
 & HIAST (Ours) & 75.9 & 37.5 & {81.3} & {29.3} & 2.0 & 40.6 & {44.4} & {39.9} & 86.0 & 88.2 & {68.4} & {30.8} & 81.5 & 40.7 & {48.7} & {60.6} & {53.5} & {60.3} \\
& {HIAST} (Ours)$^{\nabla}$ & 70.8 & 30.7 & \textbf{85.6} & 21.4 & 3.7 & 43.6 & 56.5 & \textbf{58.4} & 85.8 & 86.6 & 75.7 & \textbf{48.2} & 88.5 & \textbf{72.1} & \textbf{55.4} & \textbf{70.7} & \textbf{59.6} & \textbf{68.1} \\
\bottomrule
\end{tabular}
}
\end{table*}

\begin{table*}[htb]
\centering
\caption{Results of our HIAST and other SOTA methods (Cityscapes $\rightarrow$ Oxford RobotCar).}
\vspace{-7.5pt}
\resizebox{0.75\linewidth}{!}{
\renewcommand\arraystretch{1.3}
\begin{tabular}{c|l|ccccccccc|c}
\toprule
Year & Method & \rotatebox{90}{Road} & \rotatebox{90}{Sidewalk} & \rotatebox{90}{Building} & \rotatebox{90}{Light} & \rotatebox{90}{Sign} & \rotatebox{90}{Sky} & \rotatebox{90}{Person} & \rotatebox{90}{Automobile} & \rotatebox{90}{Two-Wheel} & mIoU \\ \midrule
2018 & AdaptSeg\cite{tsai2018learning} & {95.1} & 64.0 & 75.7 & 61.3 & 35.5 & 63.9 & 58.1 & 84.6 & 57.0 & 69.5 \\ \midrule
2019 & PatchAlign\cite{Tsai_adaptseg_ICCV19} & 94.4 & 63.5 & 82.0 & 61.3 & 36.0 & 76.4 & 61.0 & 86.5 & 58.6 & 72.0 \\
2020 & MRNet\cite{zheng2019unsupervised}  & \textbf{95.9} & {73.5} & 86.2 & 69.3 & 31.9 & 87.3 & 57.9 & {88.8} & \textbf{61.5} & 72.5 \\
2021 & MRNet+Rectifying\cite{Zheng2021RectifyingPL}  & \textbf{95.9} & \textbf{73.7} & 87.4 & {72.8} & \textbf{43.1} & 88.6 & \textbf{61.7} & \textbf{89.6} & 57.0 & 74.4 \\ \midrule
 & IAST (Ours) & 94.8 & 70.8 & \textbf{93.1} & 69.3 & 33.9 & \textbf{96.1} & 57.1 & 86.9 & 56.9 & 73.2 \\
 & HIAST (Ours)  & 94.9 & 71.7 & {92.7} & \textbf{75.0} & {40.5} & {95.6} & {61.0} & 87.1 & {58.5} & \textbf{75.2} \\
\bottomrule
\end{tabular}
}
\label{table:oxford}
\end{table*}

\begin{figure*}[!htb]
\centering
\subfigure{
\begin{minipage}[t]{.02\linewidth}
    \scriptsize
    \vspace{-4.5em}
	\rotatebox{90}{GTA5 $\rightarrow$ Cityscapes}\vspace{8em}
	\rotatebox{90}{SYNTHIA $\rightarrow$ Cityscapes}\vspace{5.25em}
	\rotatebox{90}{Cityscapes $\rightarrow$ Oxford RobotCar}
\end{minipage}
}
\hspace{-1.2em}
\setcounter{subfigure}{0}
\subfigure[Target image]{
\begin{minipage}[t]{0.225\linewidth}
    \includegraphics[width=\linewidth]{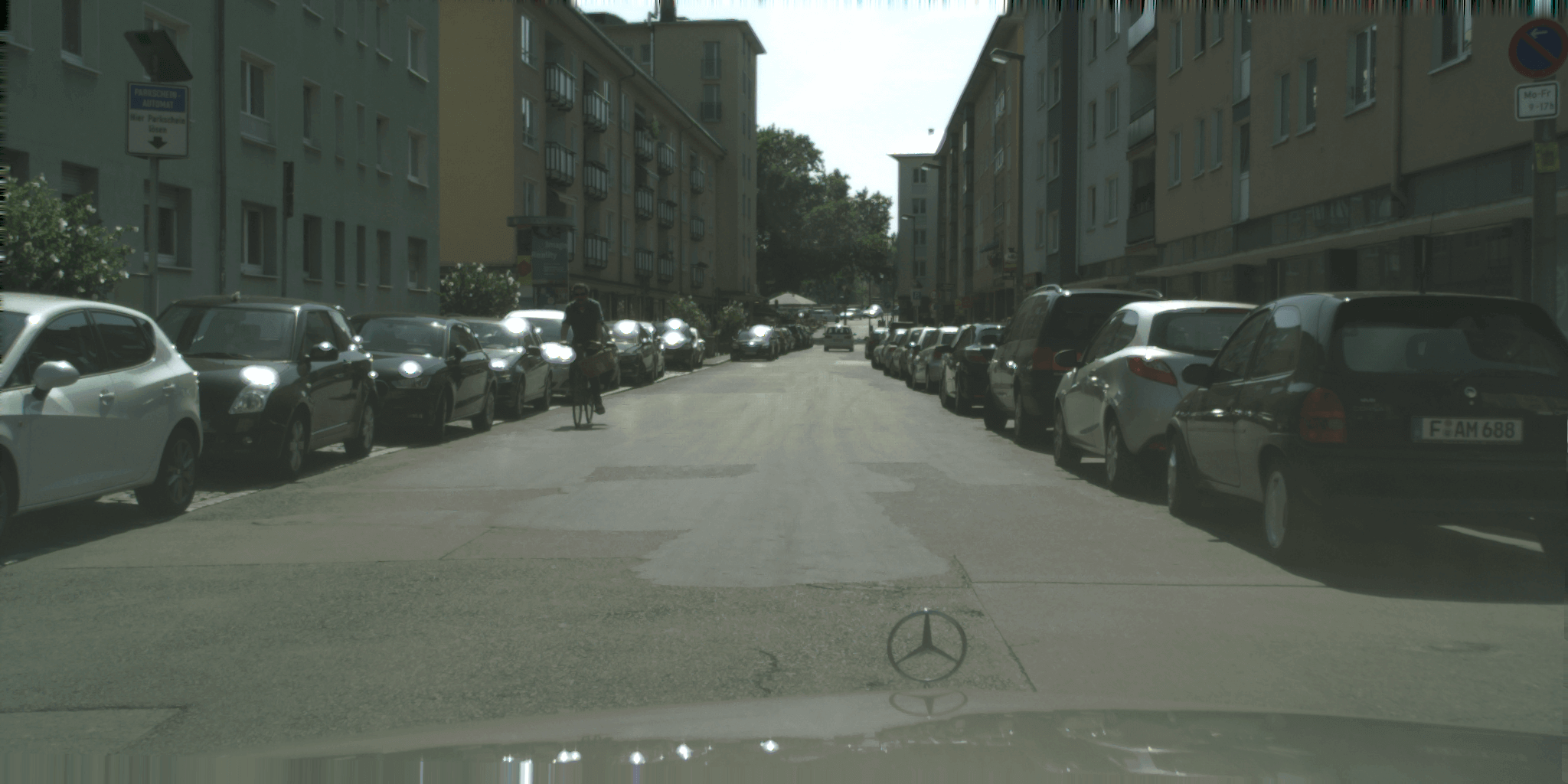}\vspace{3pt}
    \includegraphics[width=\linewidth]{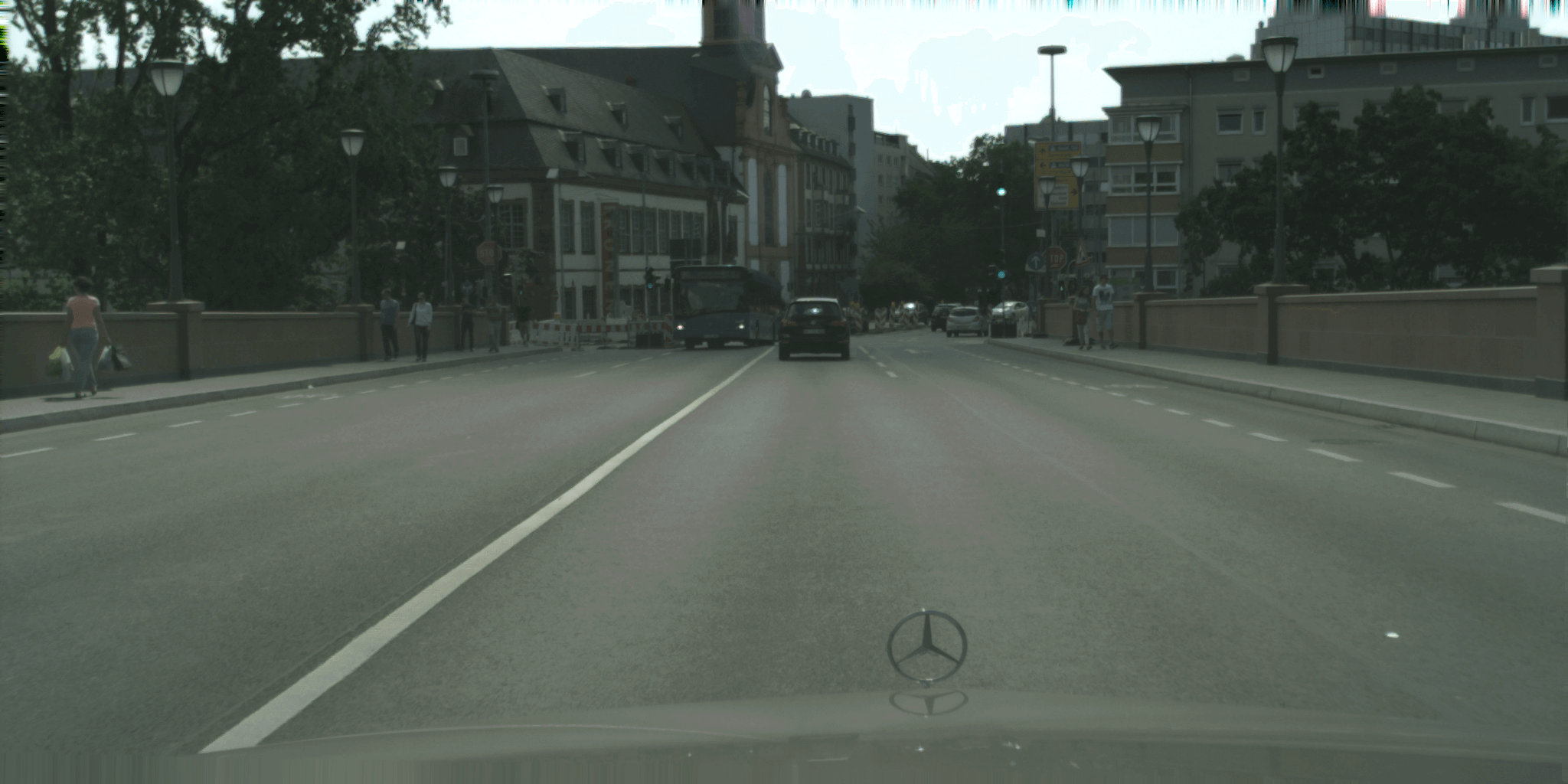}\vspace{6pt}
    \includegraphics[width=\linewidth]{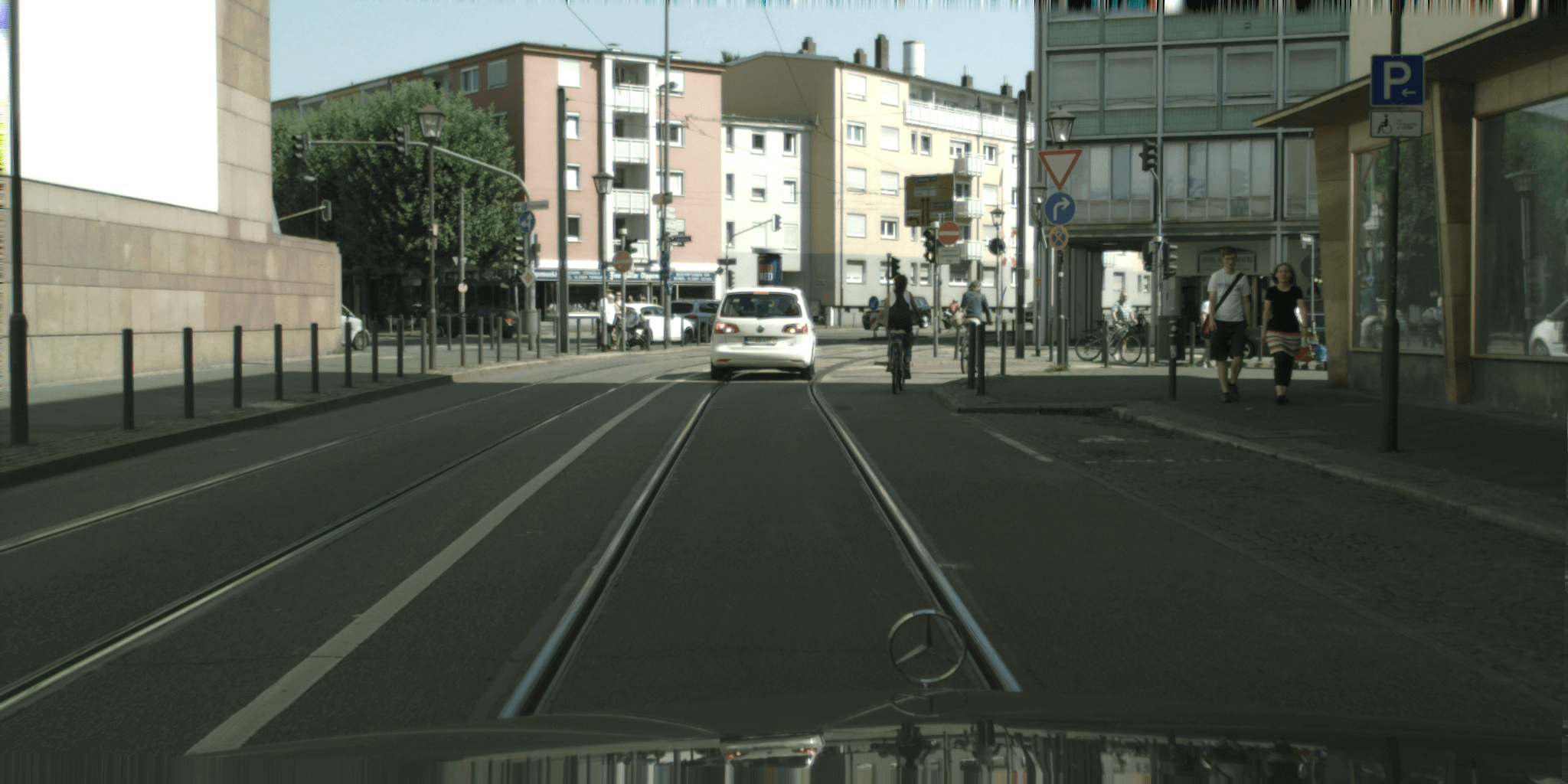}\vspace{3pt}
    \includegraphics[width=\linewidth]{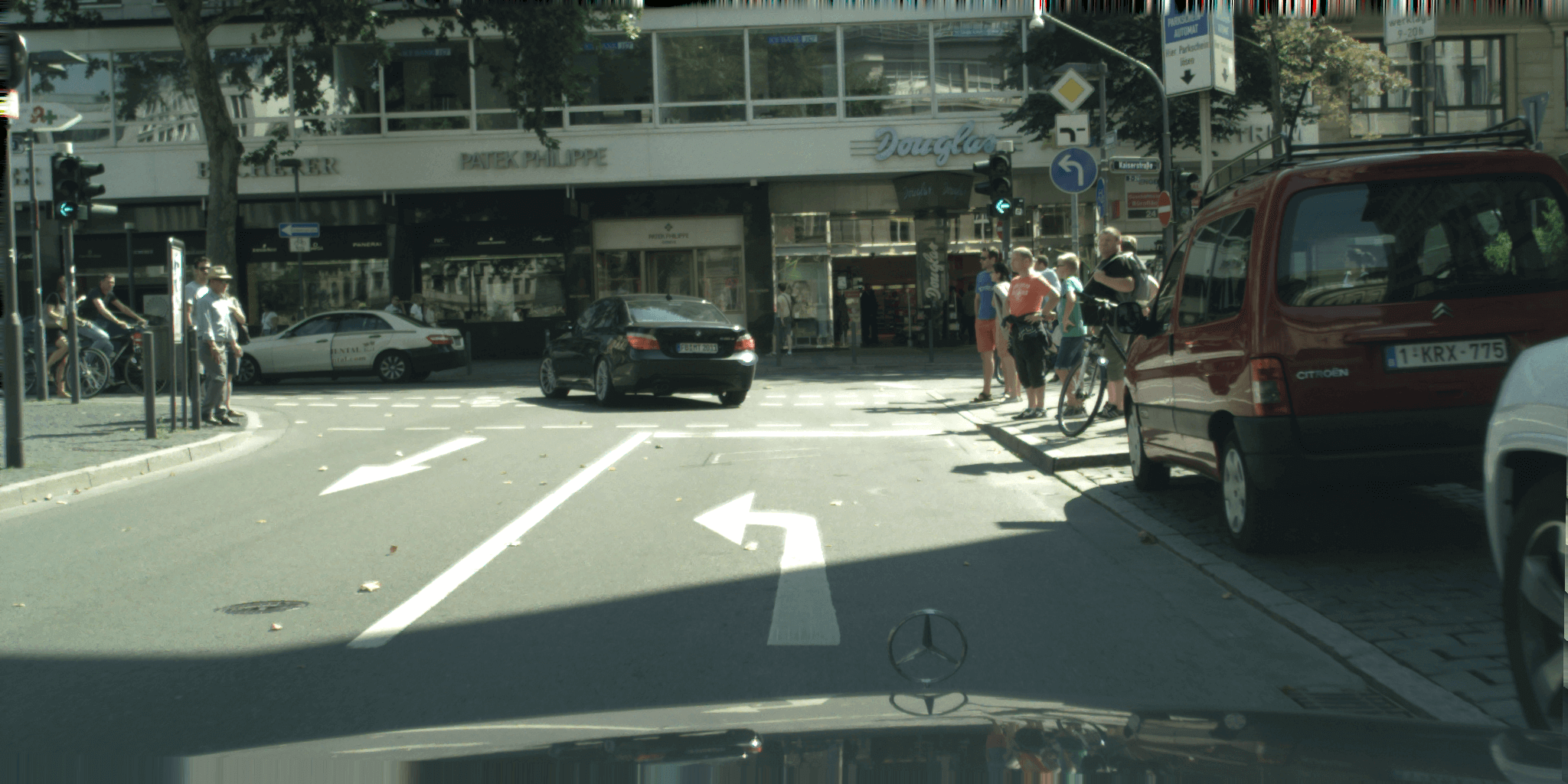}\vspace{6pt}
    \includegraphics[width=\linewidth,height=0.5\linewidth]{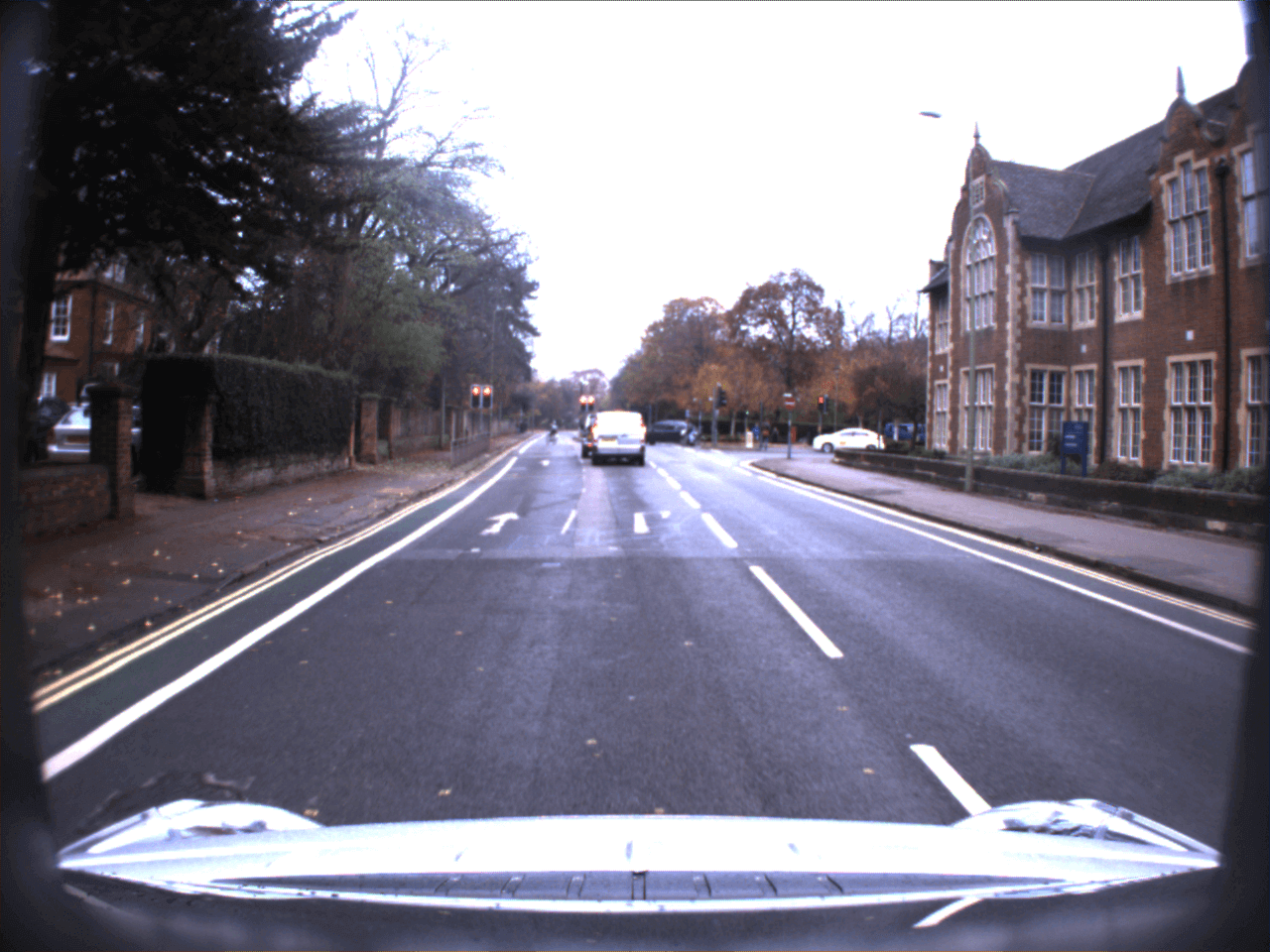}\vspace{3pt}
    \includegraphics[width=\linewidth,height=0.5\linewidth]{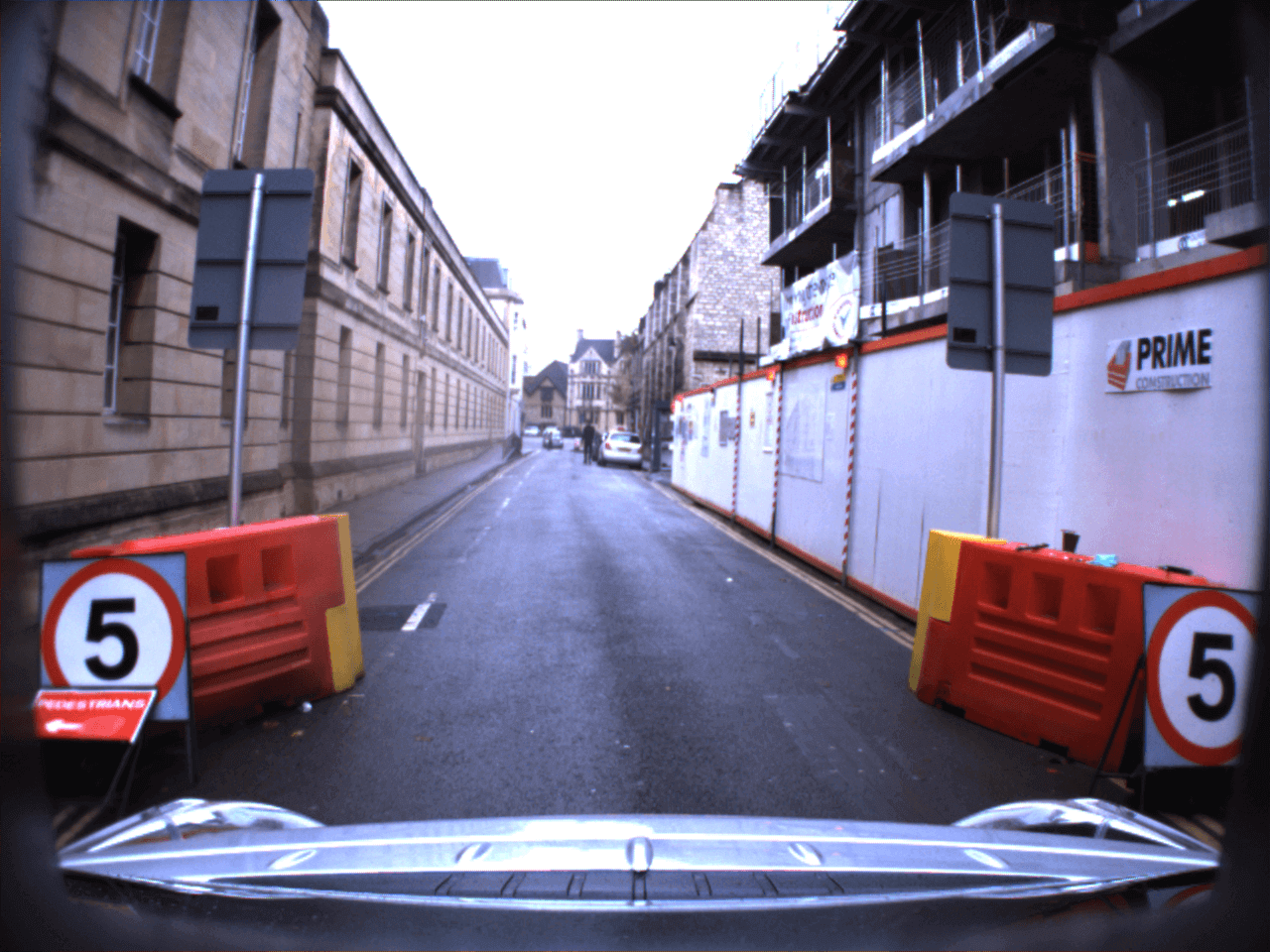}\vspace{5pt}
\end{minipage}
}
\subfigure[Ground truth]{
\begin{minipage}[t]{0.225\linewidth}
    \includegraphics[width=\linewidth]{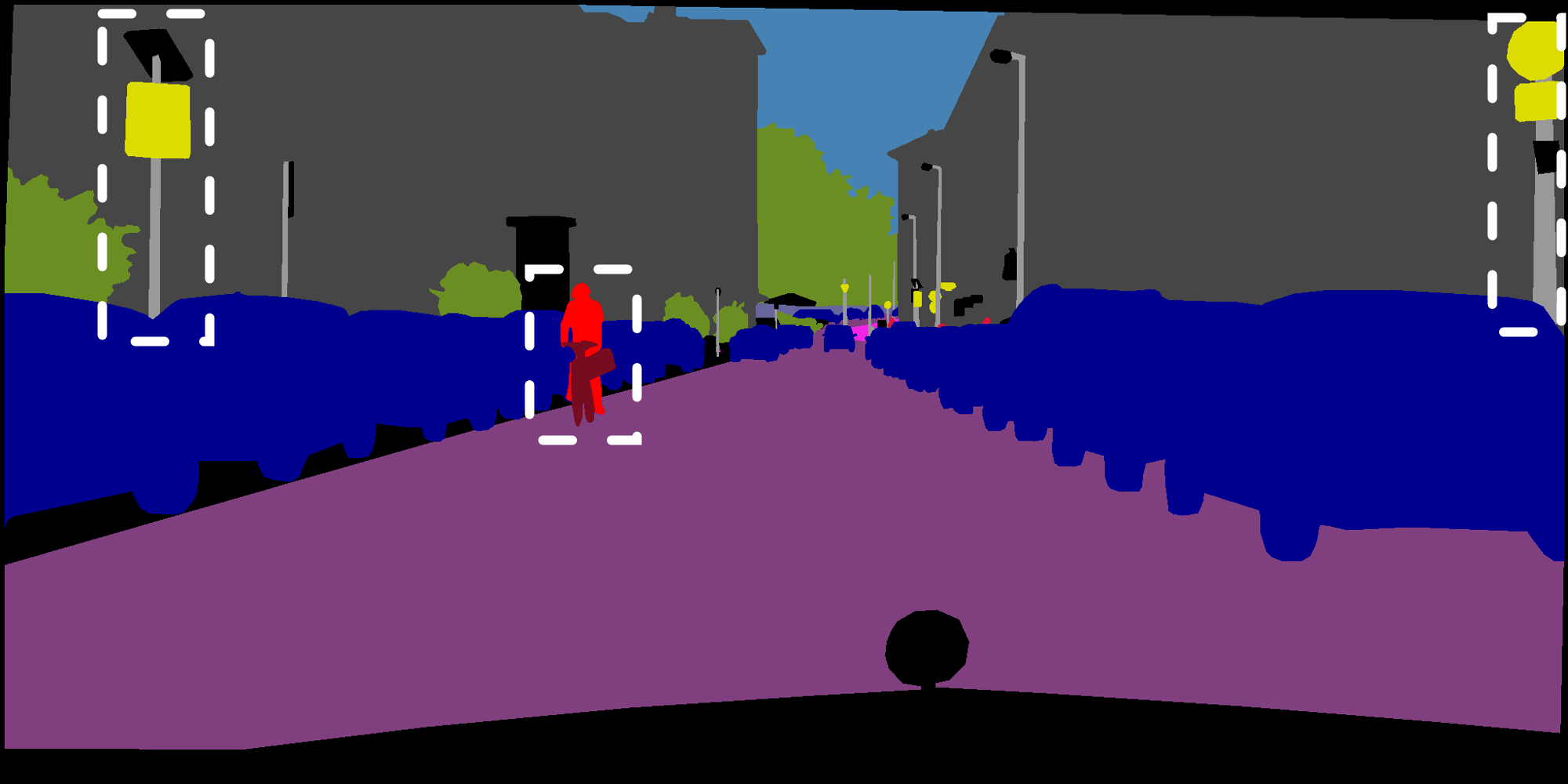}\vspace{3pt}
    \includegraphics[width=\linewidth]{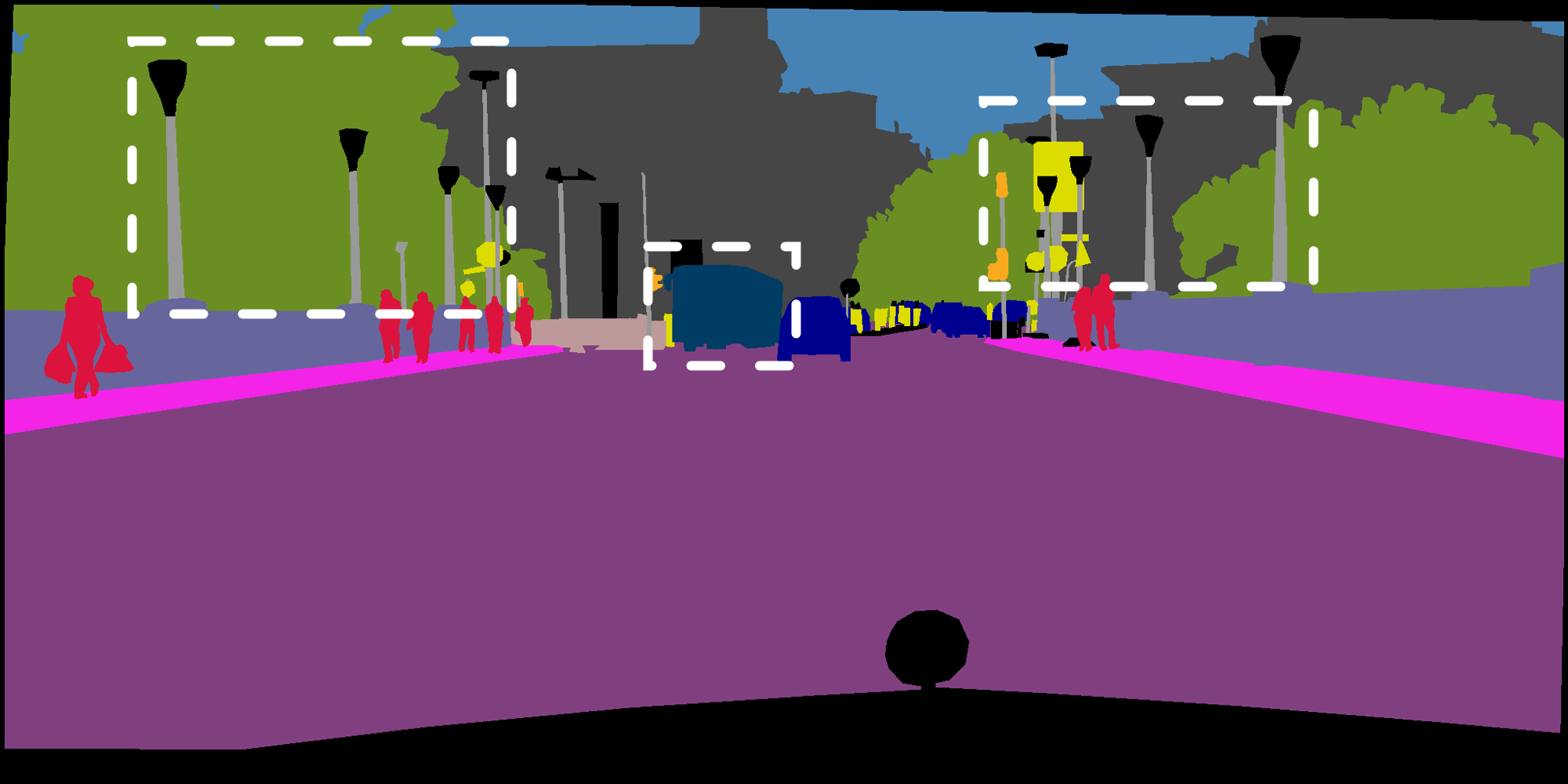}\vspace{6pt}
    \includegraphics[width=\linewidth]{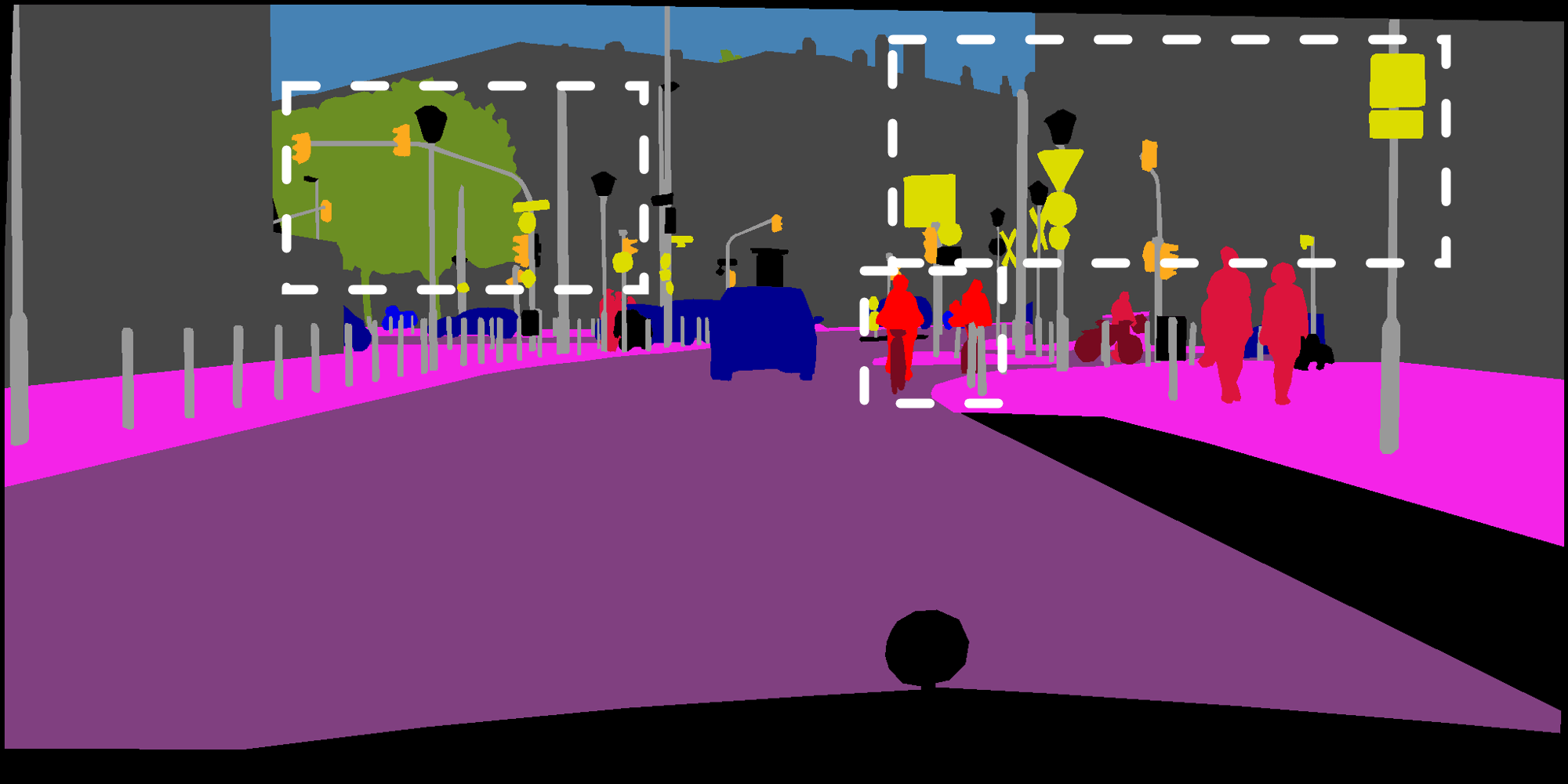}\vspace{3pt}
    \includegraphics[width=\linewidth]{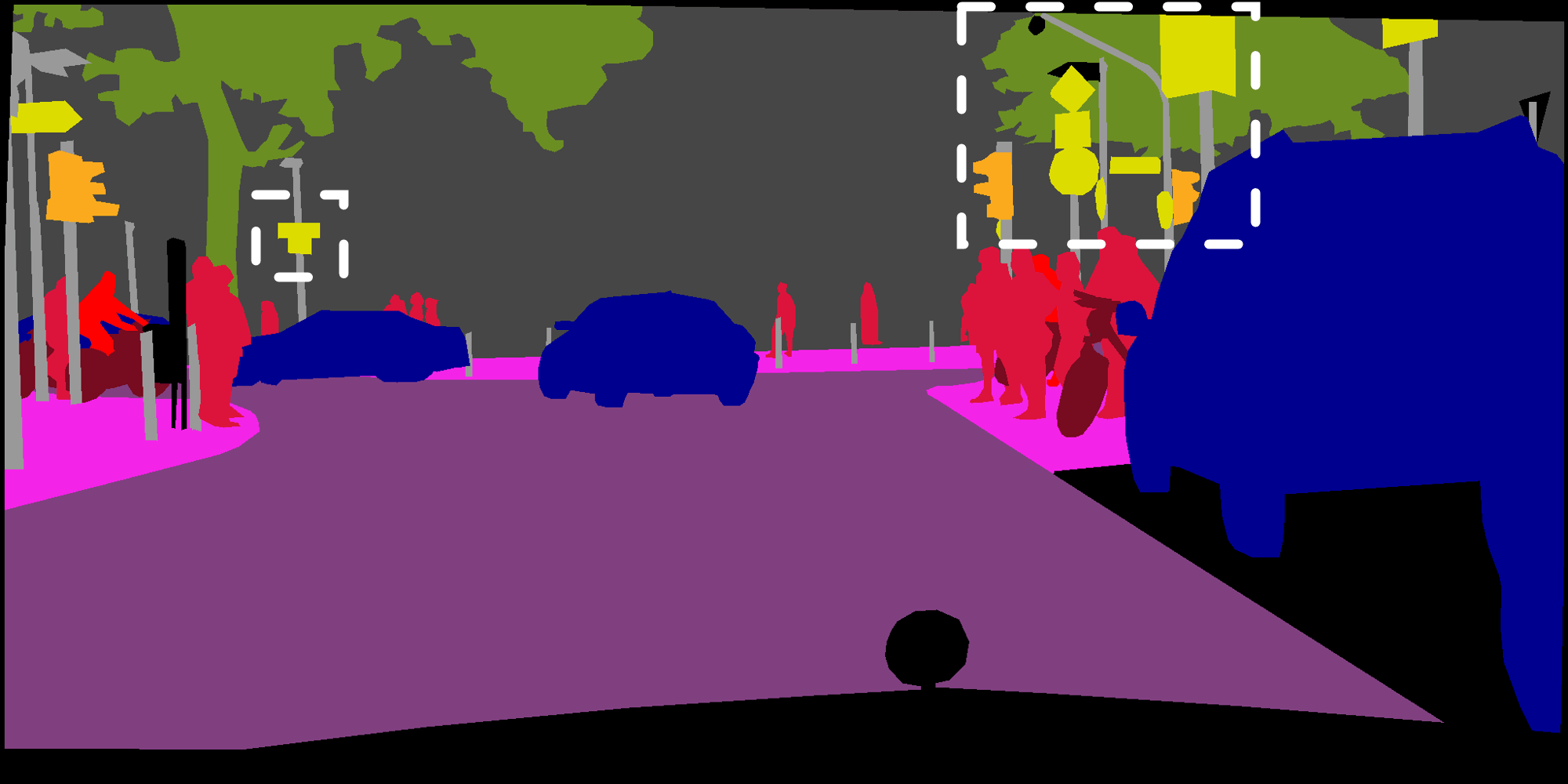}\vspace{6pt}
    \includegraphics[width=\linewidth,height=0.5\linewidth]{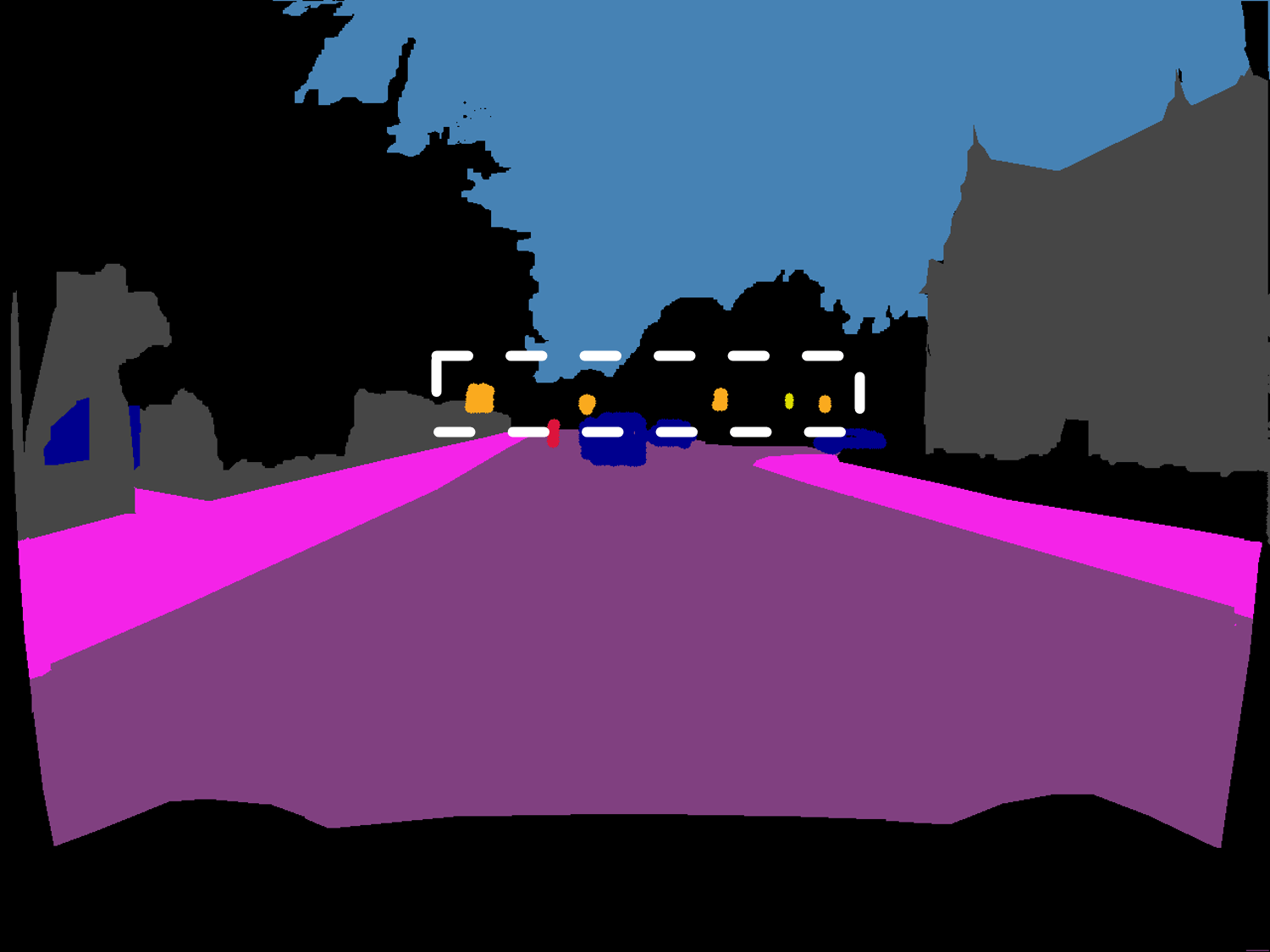}\vspace{3pt}
    \includegraphics[width=\linewidth,height=0.5\linewidth]{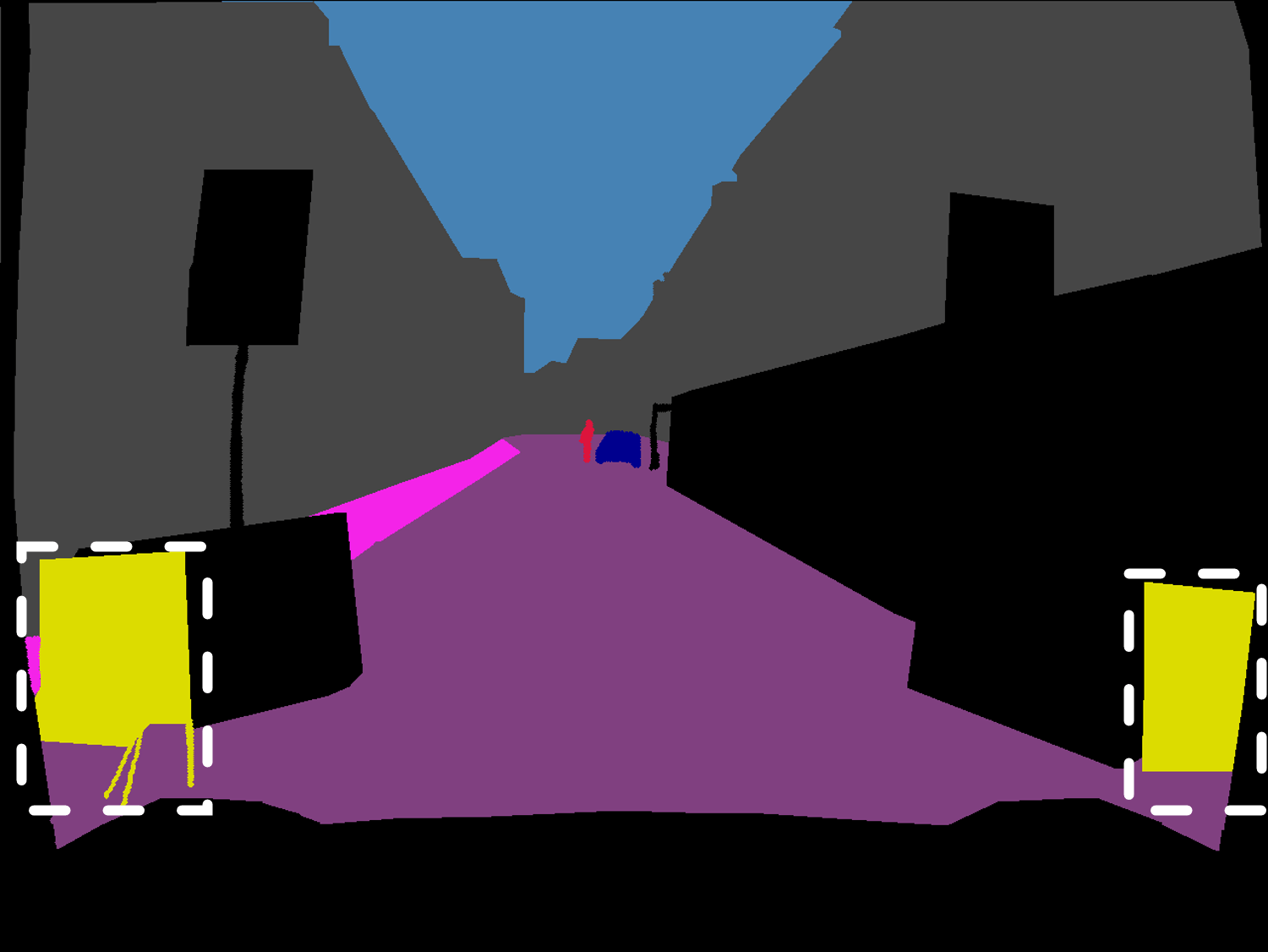}\vspace{5pt}
\end{minipage}
}
\subfigure[IAST (Ours)]{
\begin{minipage}[t]{0.225\linewidth}
    \includegraphics[width=\linewidth]{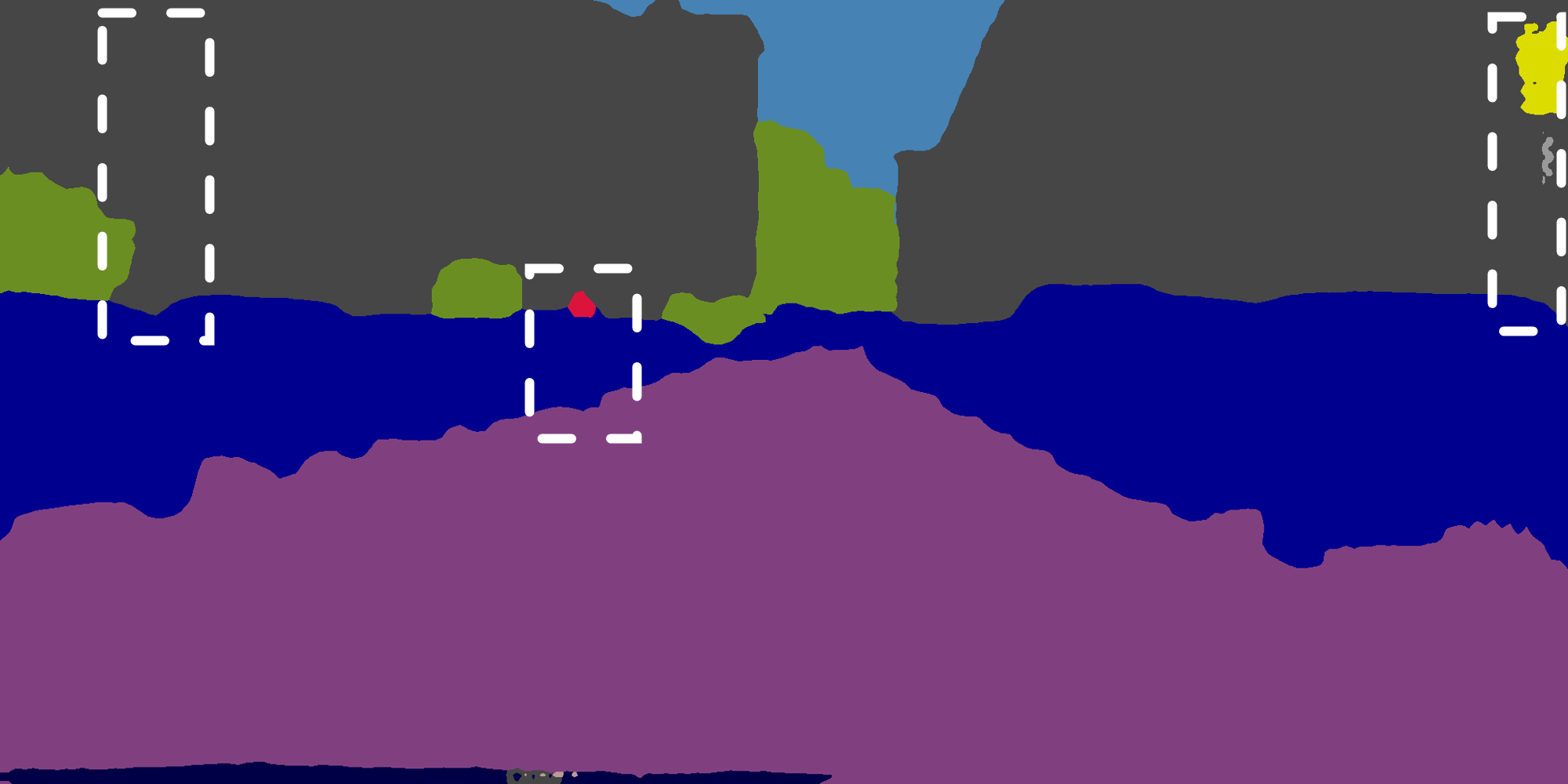}\vspace{3pt}
    \includegraphics[width=\linewidth]{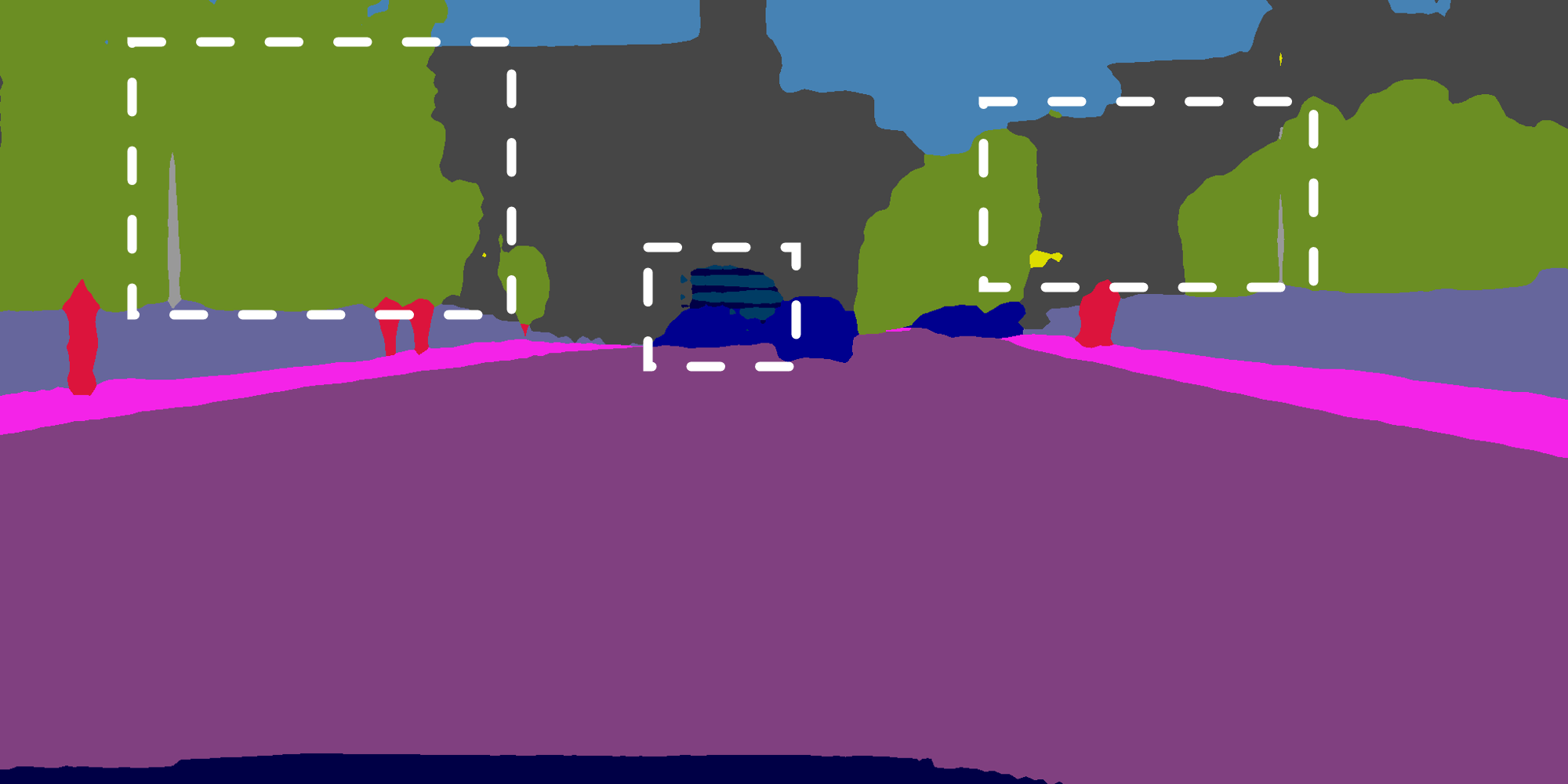}\vspace{6pt}
    \includegraphics[width=\linewidth]{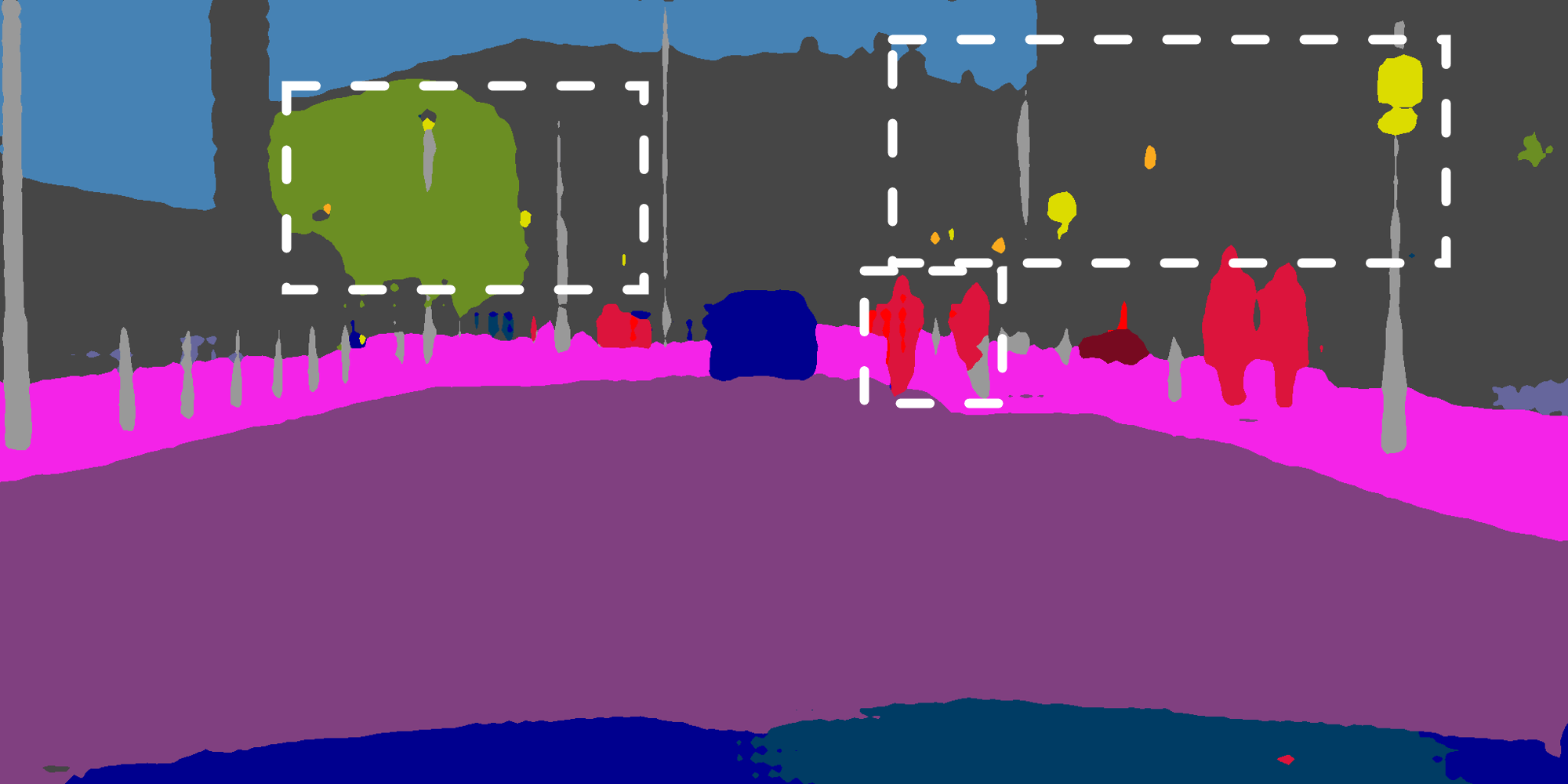}\vspace{3pt}
    \includegraphics[width=\linewidth]{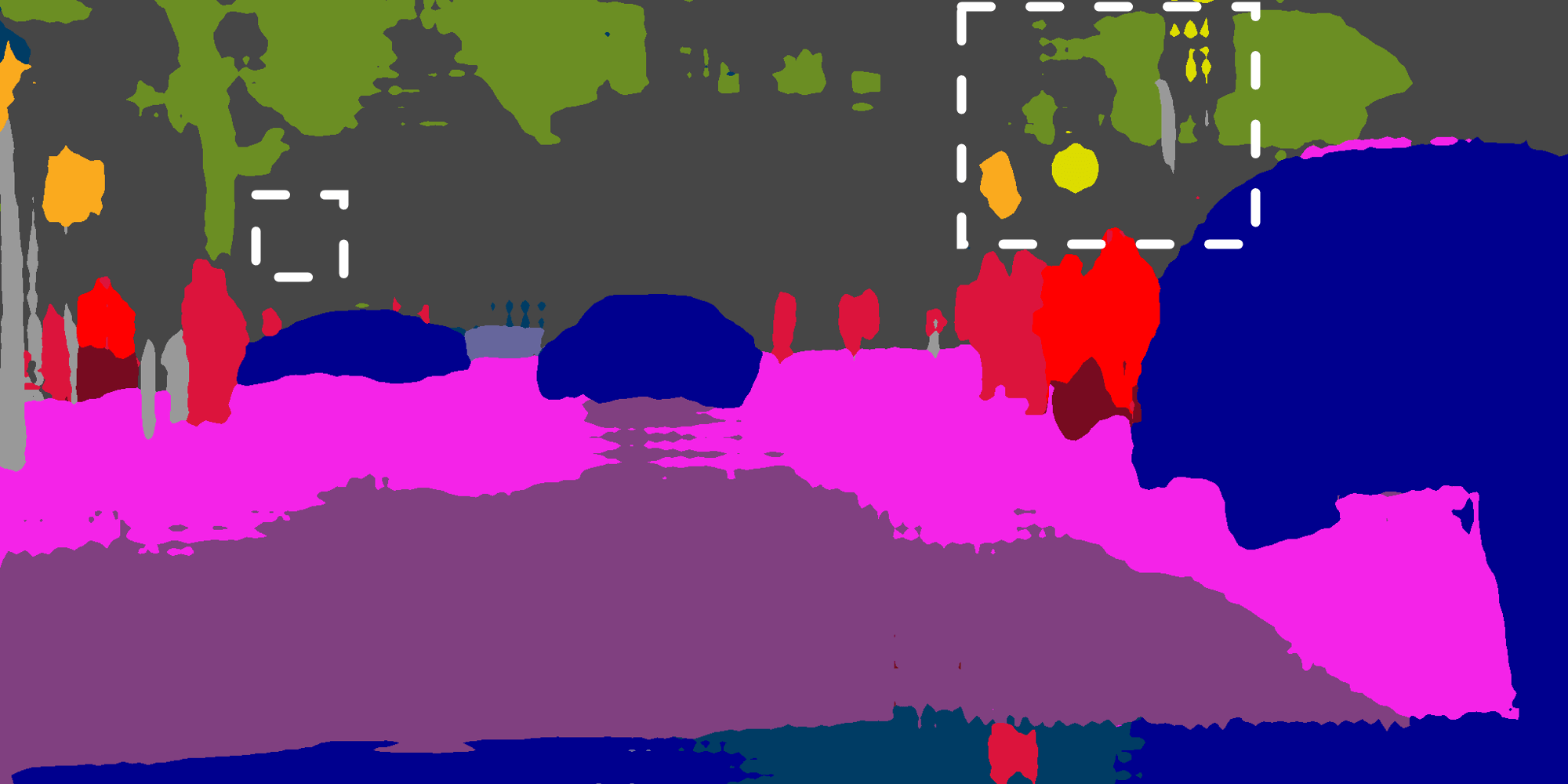}\vspace{6pt}
    \includegraphics[width=\linewidth,height=0.5\linewidth]{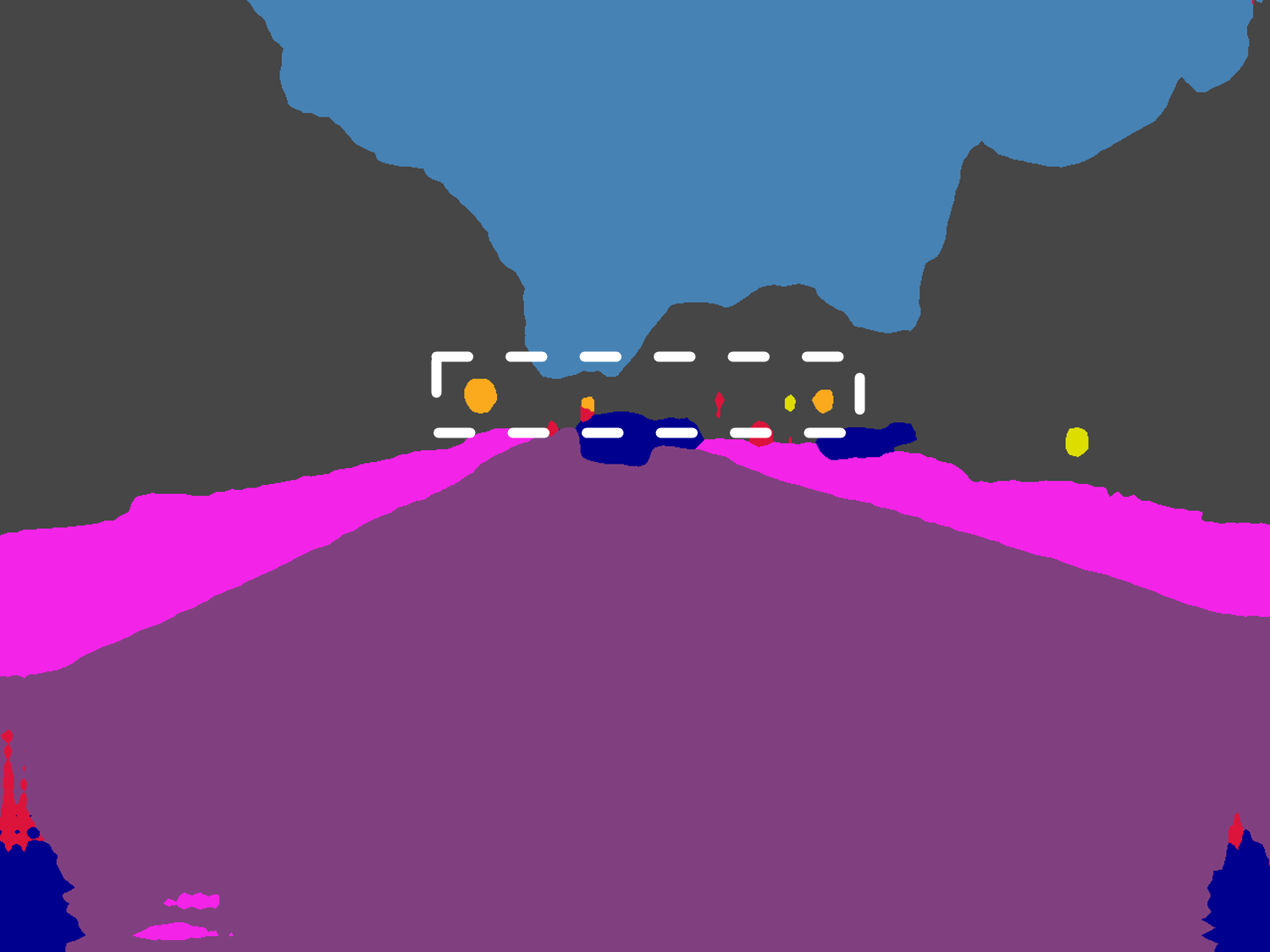}\vspace{3pt}
    \includegraphics[width=\linewidth,height=0.5\linewidth]{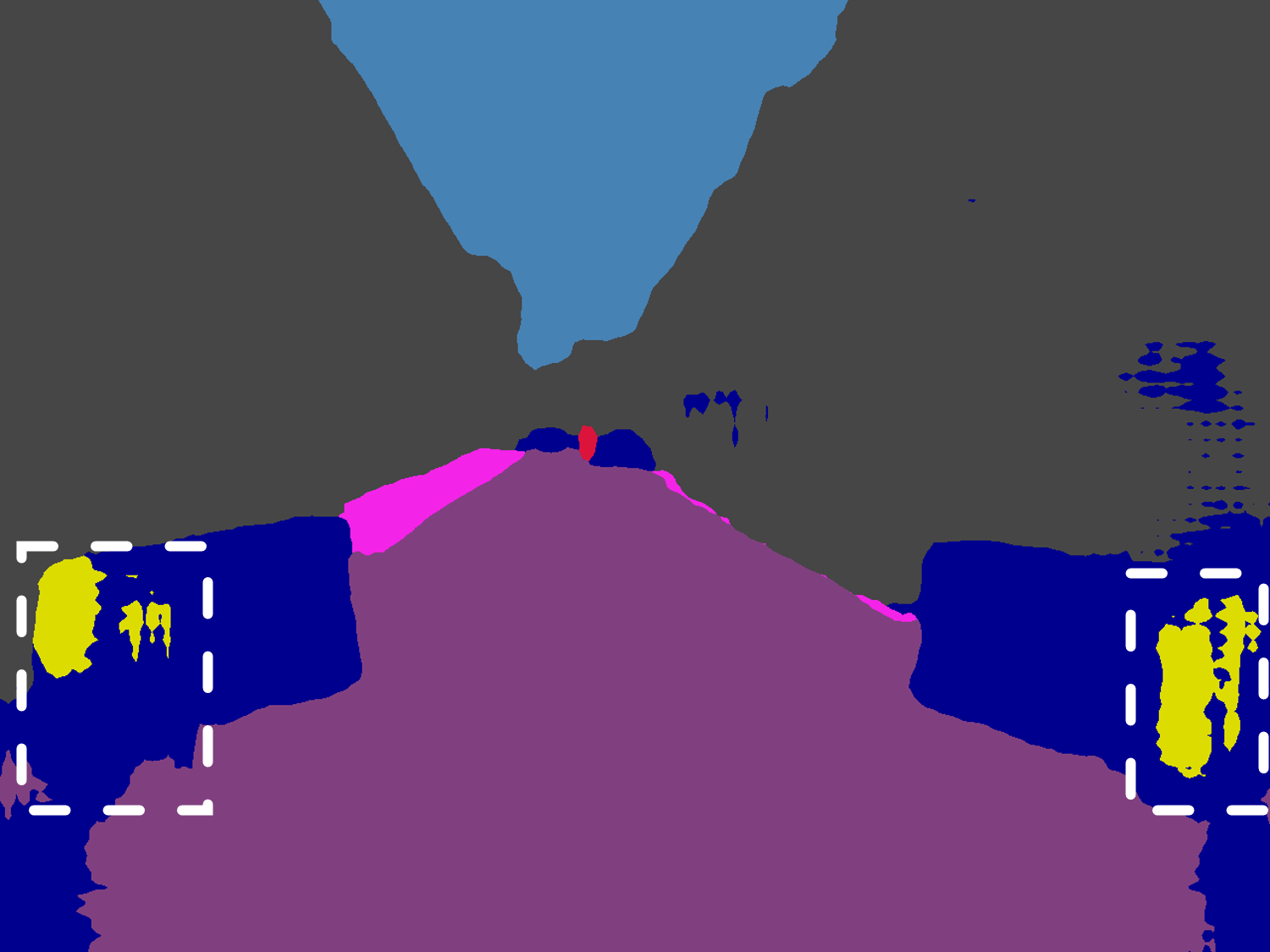}\vspace{5pt}
\end{minipage}
}
\subfigure[HIAST (Ours)]{
\begin{minipage}[t]{0.225\linewidth}
    \includegraphics[width=\linewidth]{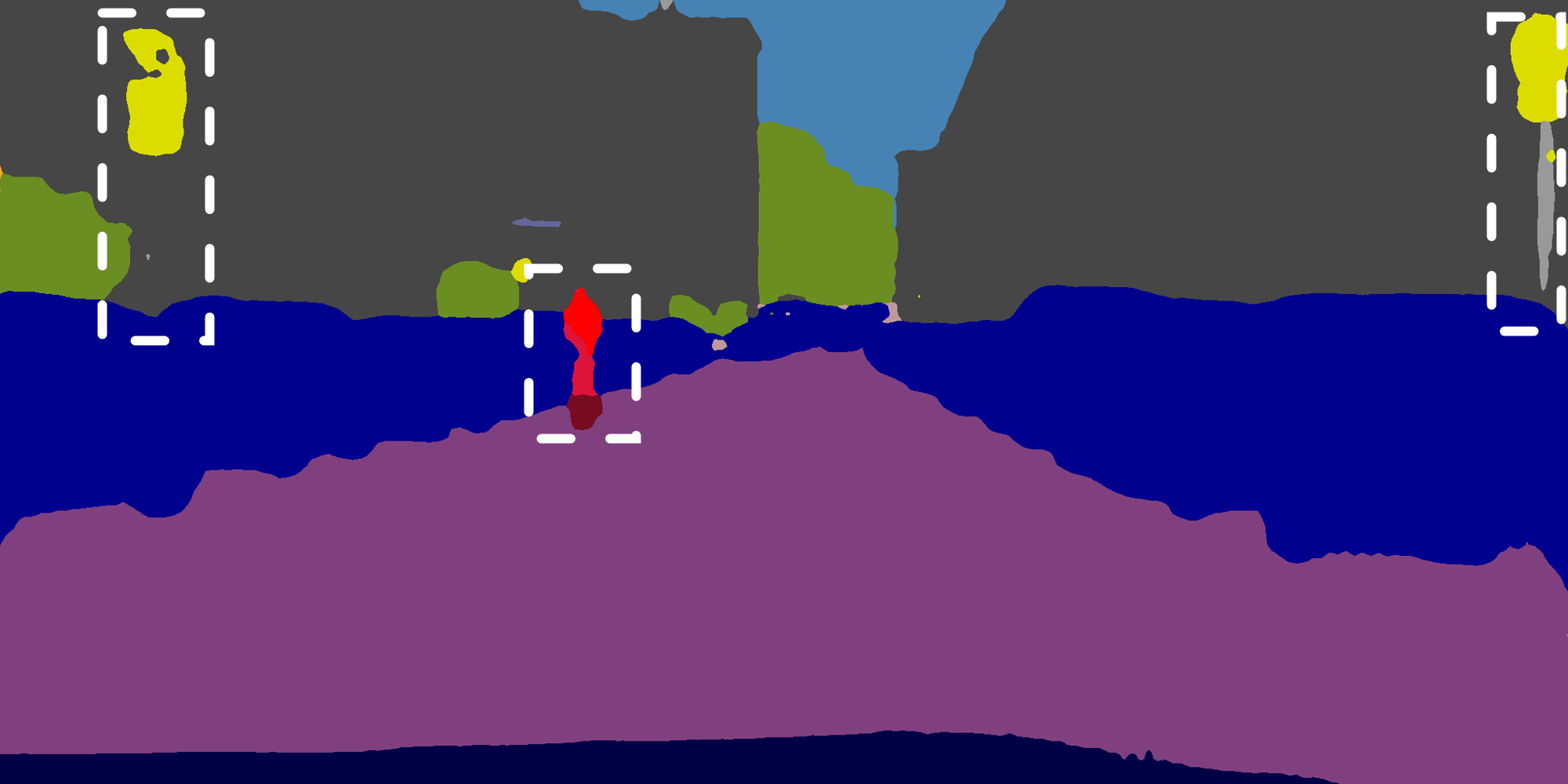}\vspace{3pt}
    \includegraphics[width=\linewidth]{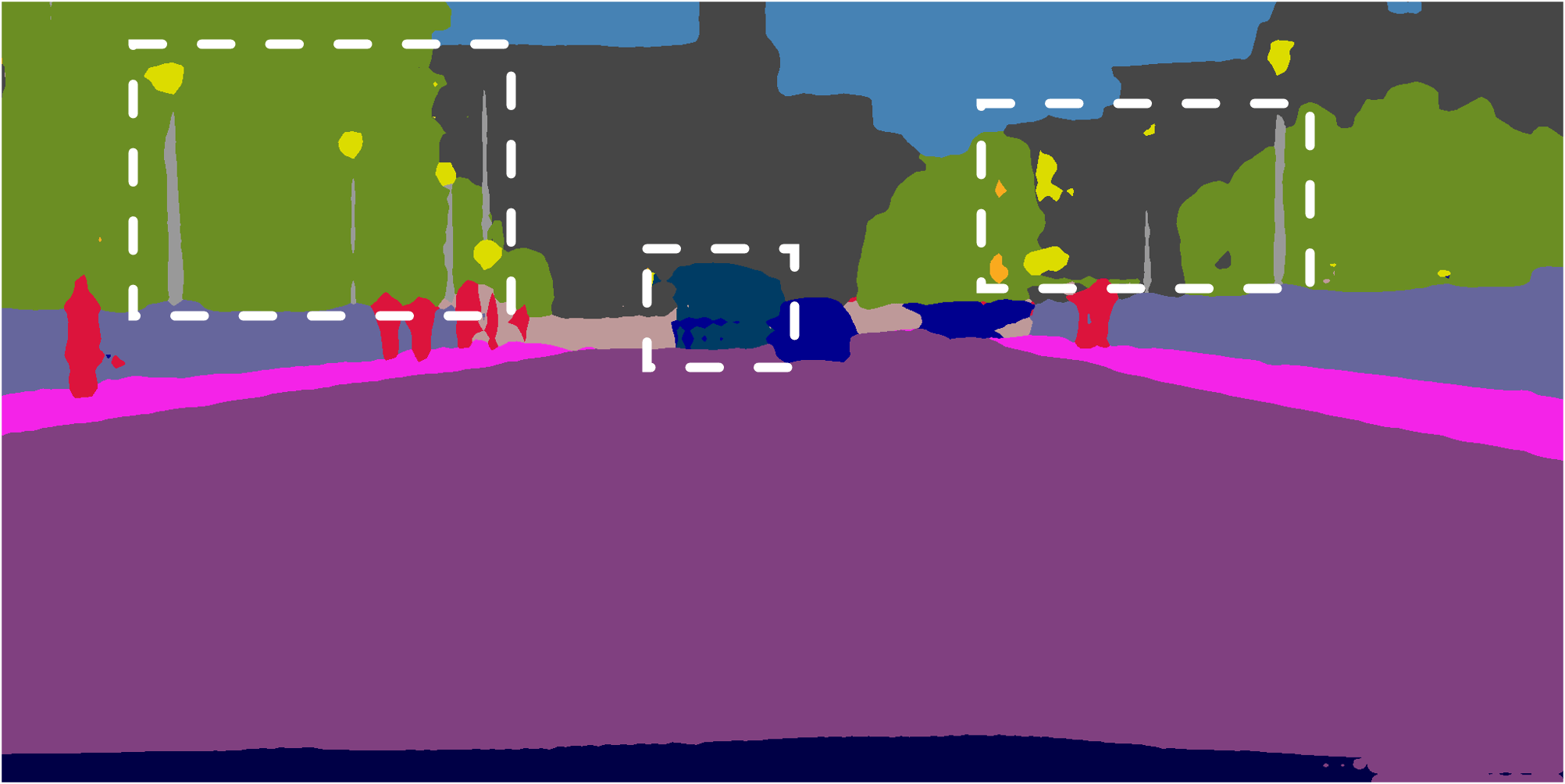}\vspace{6pt}
    \includegraphics[width=\linewidth]{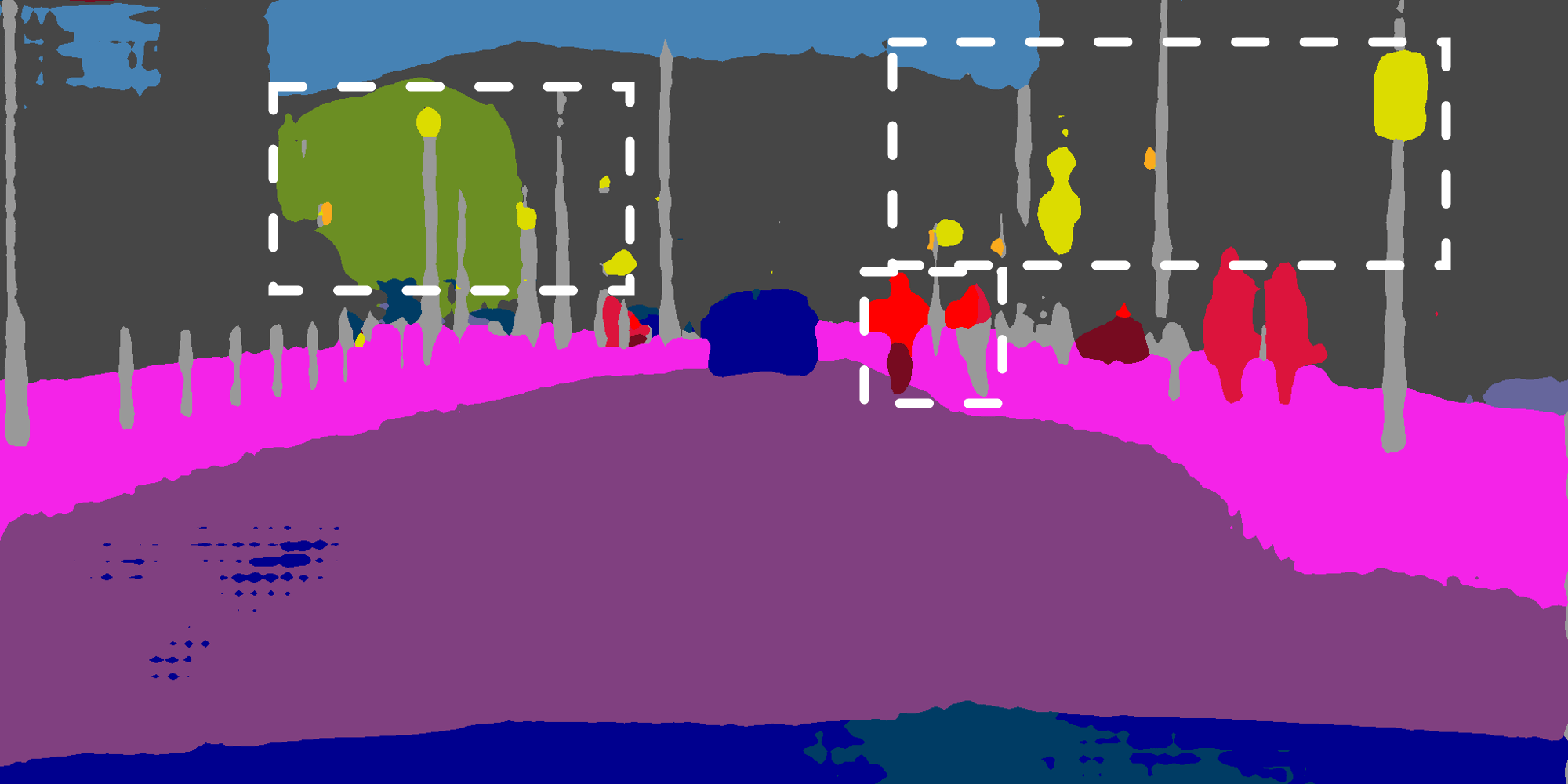}\vspace{3pt}
    \includegraphics[width=\linewidth]{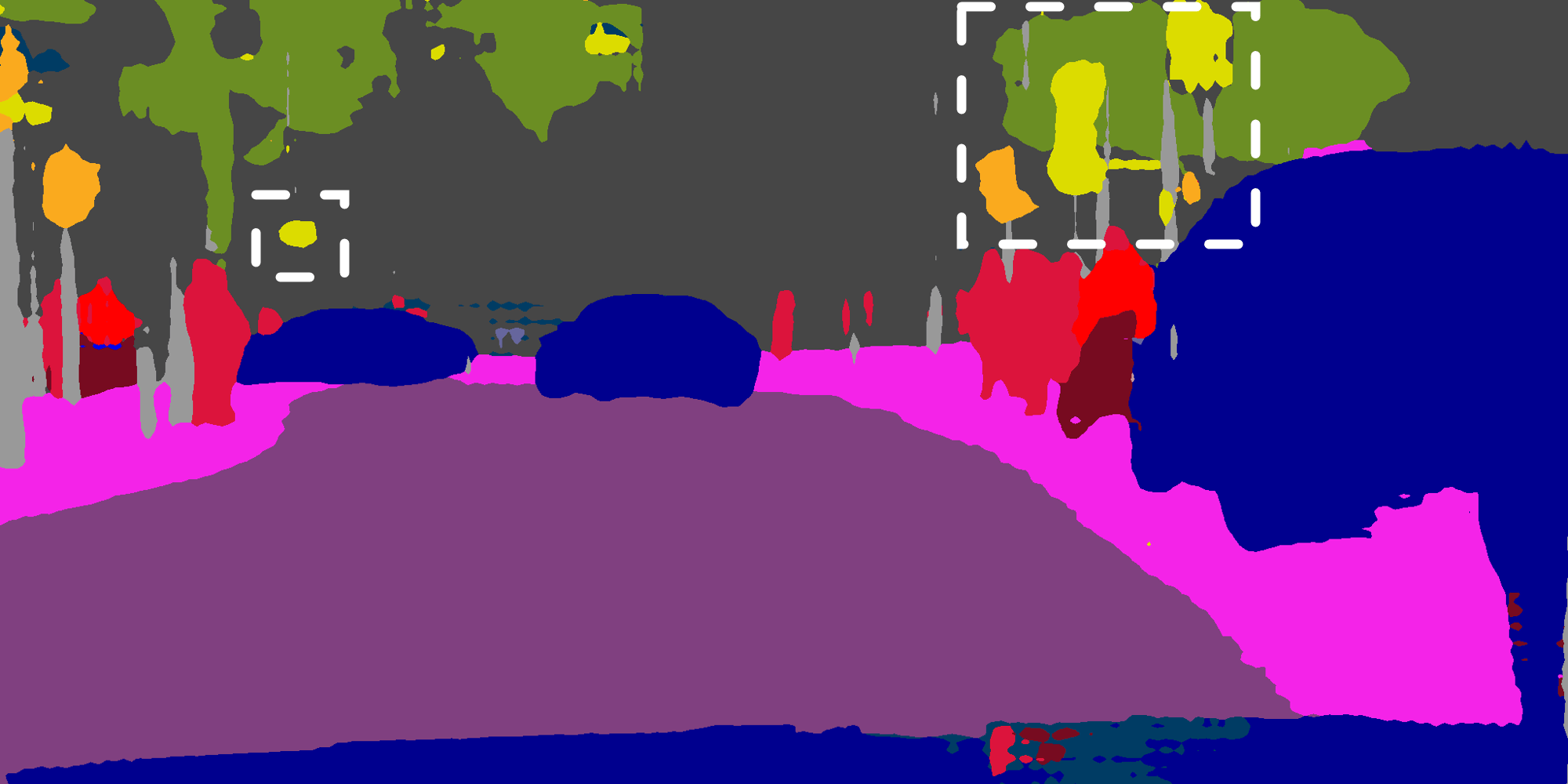}\vspace{6pt}
    \includegraphics[width=\linewidth,height=0.5\linewidth]{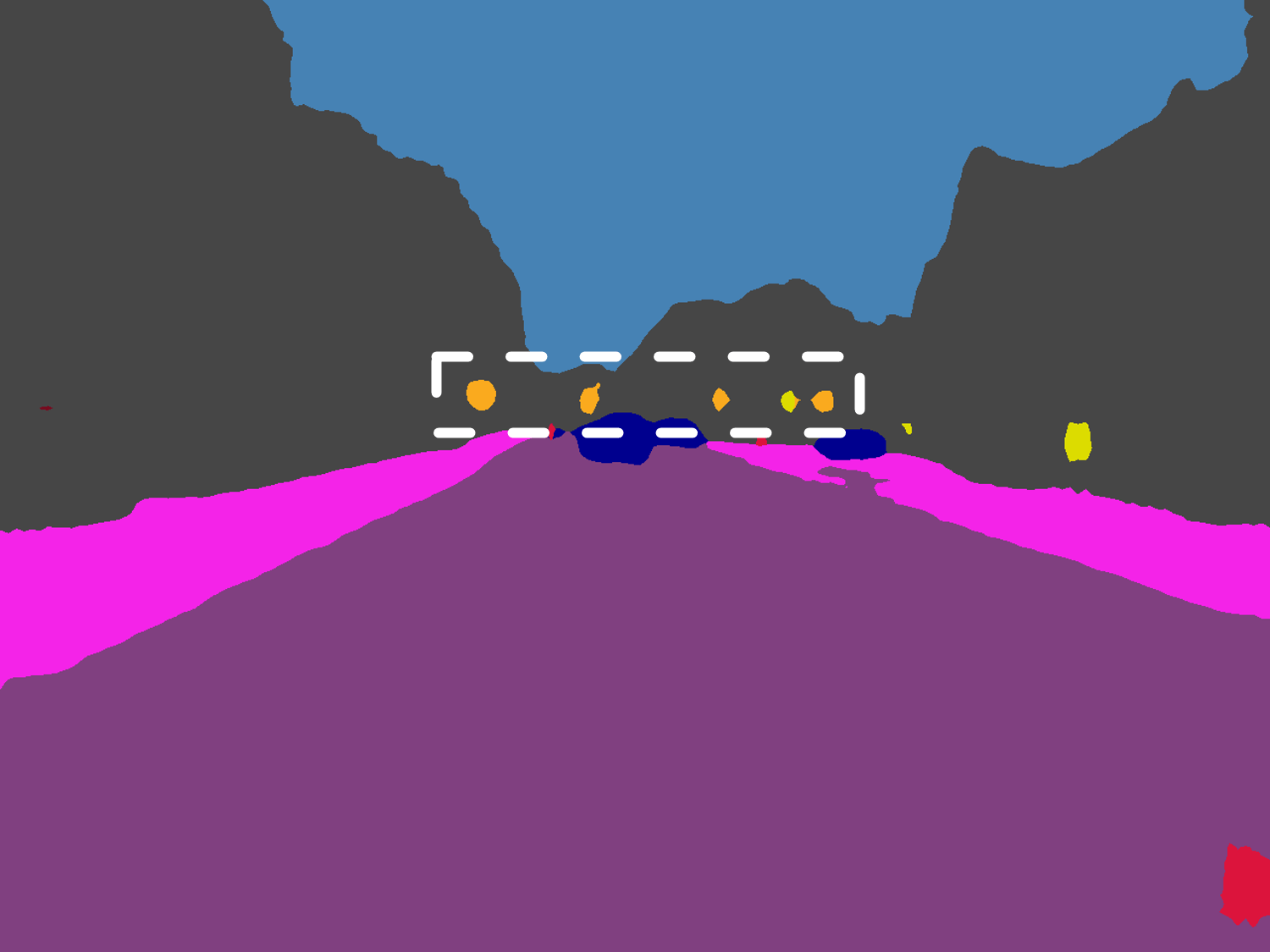}\vspace{3pt}
    \includegraphics[width=\linewidth,height=0.5\linewidth]{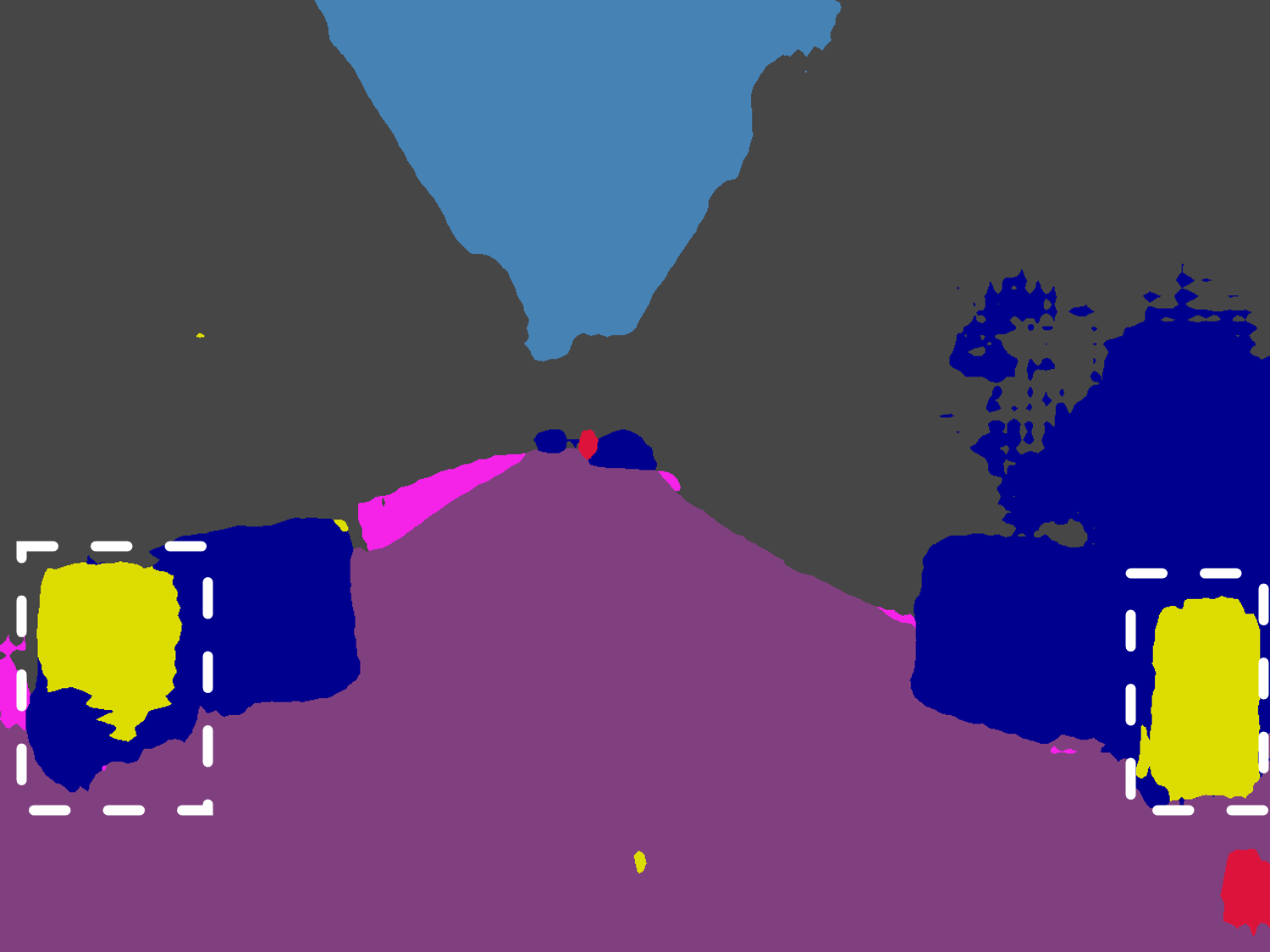}\vspace{5pt}
\end{minipage}
}
\vspace{-10pt}
\caption{Qualitative results of UDA semantic segmentation on GTA5 $\rightarrow$ Cityscapes, SYNTHIA $\rightarrow$ Cityscapes, and Cityscapes $\rightarrow$ Oxford RobotCar. For each UDA scenario, we show the results of IAST and HIAST. IAST can distinguish parts of hard classes, such as pole, traffic light, and traffic sign. HIAST further makes significant improvements on the segmentation results.
}
\vspace{-5pt}
\label{fig:visualization_prediction}
\end{figure*}

\begin{figure*}[htb]
\centering
\subfigure[Target image]{\includegraphics[width=0.235\linewidth]{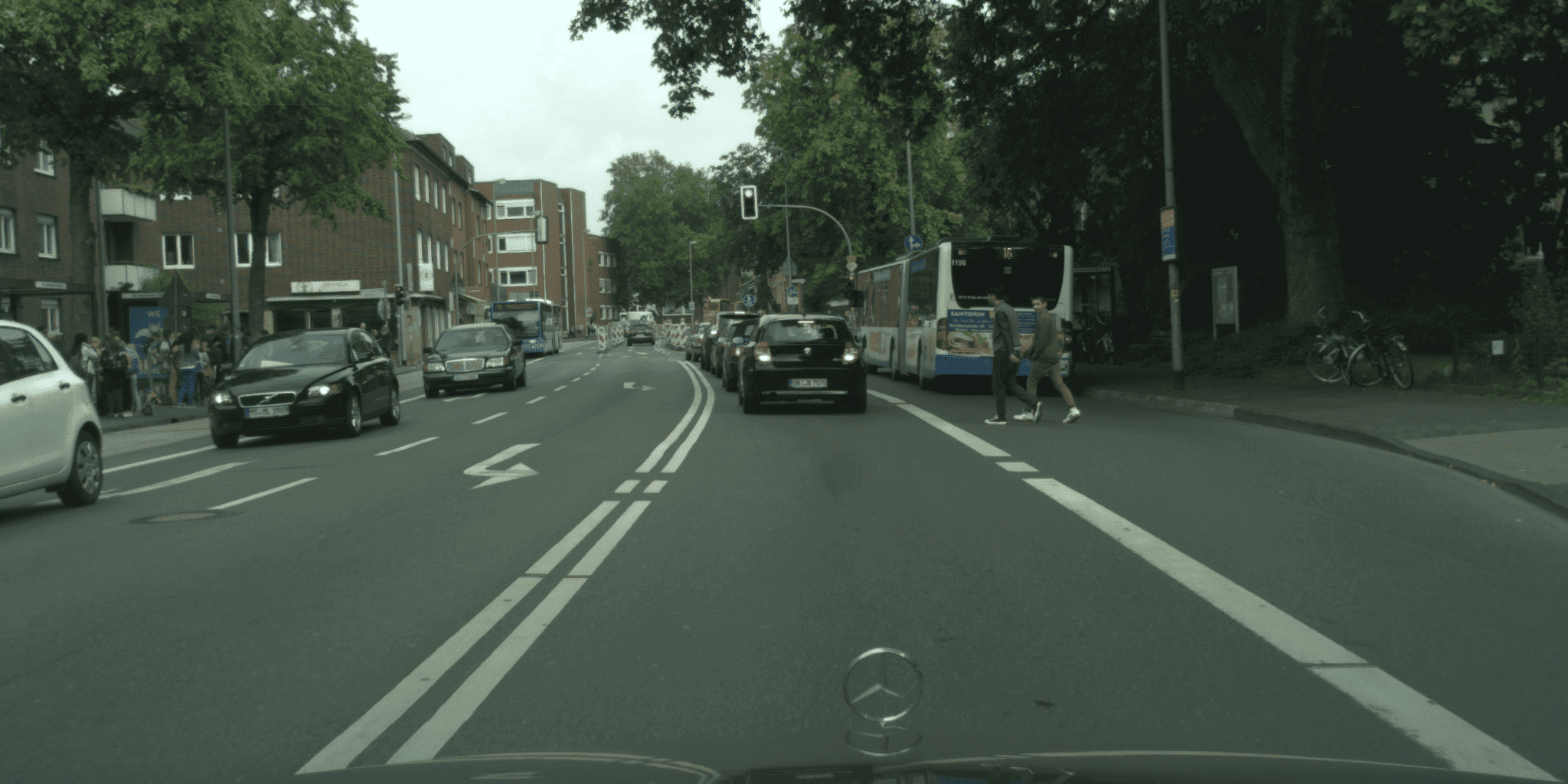}}
\subfigure[Ground truth]{\includegraphics[width=0.235\linewidth]{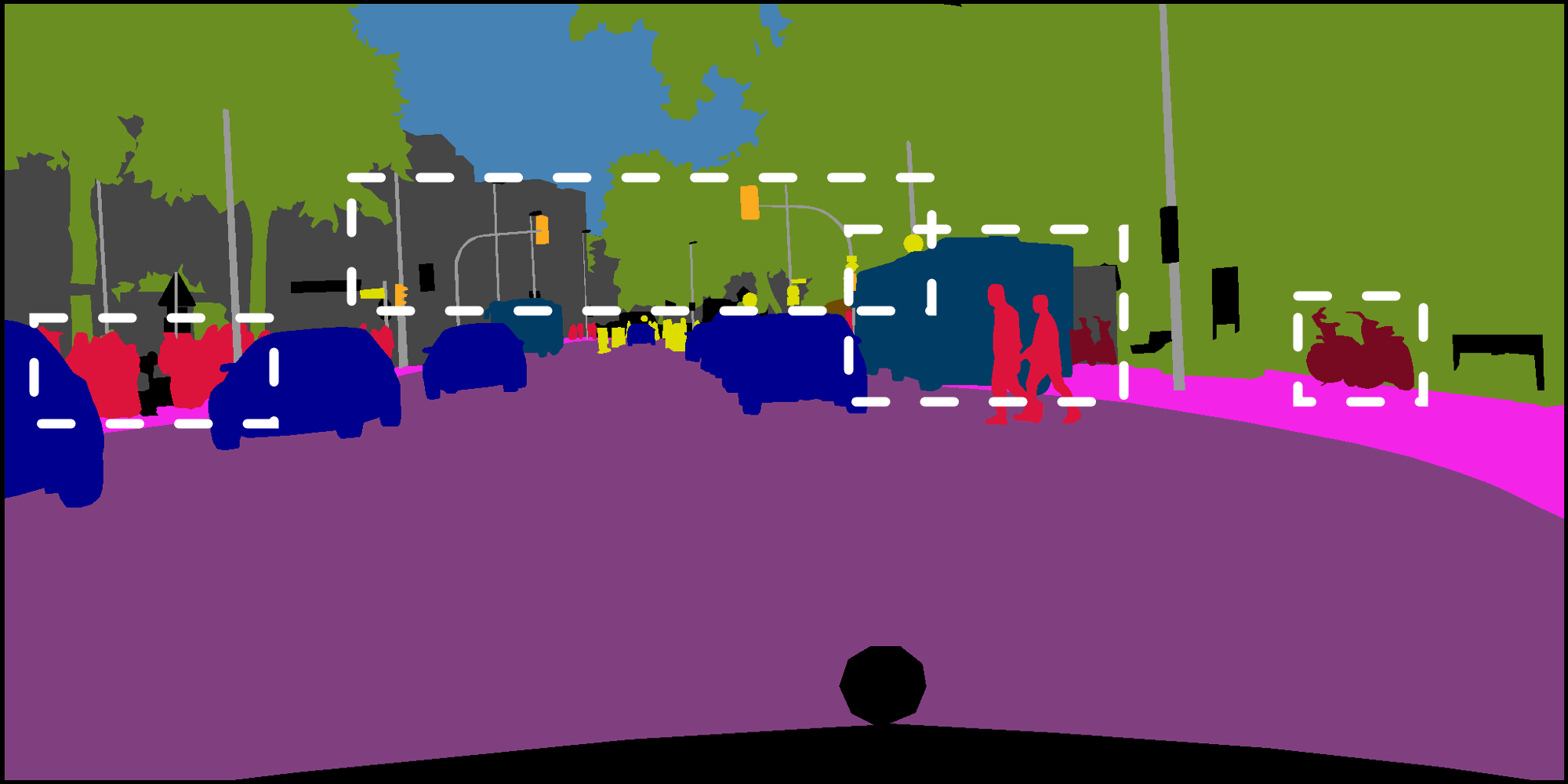}}
\subfigure[Without adaptation]{\includegraphics[width=0.235\linewidth]{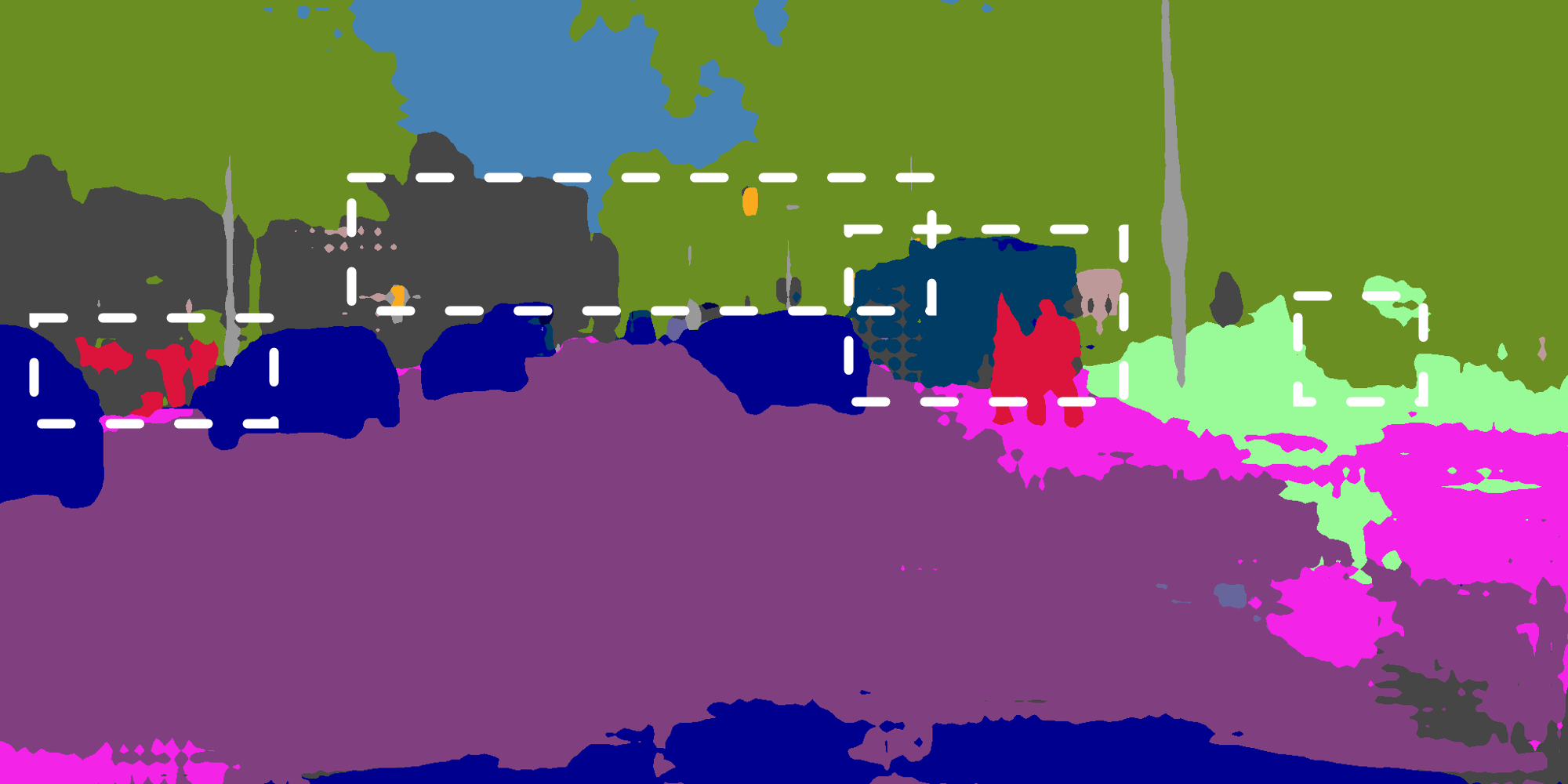}}
\subfigure[MRNet+Rectifing\cite{Zheng2021RectifyingPL}]{\includegraphics[width=0.235\linewidth]{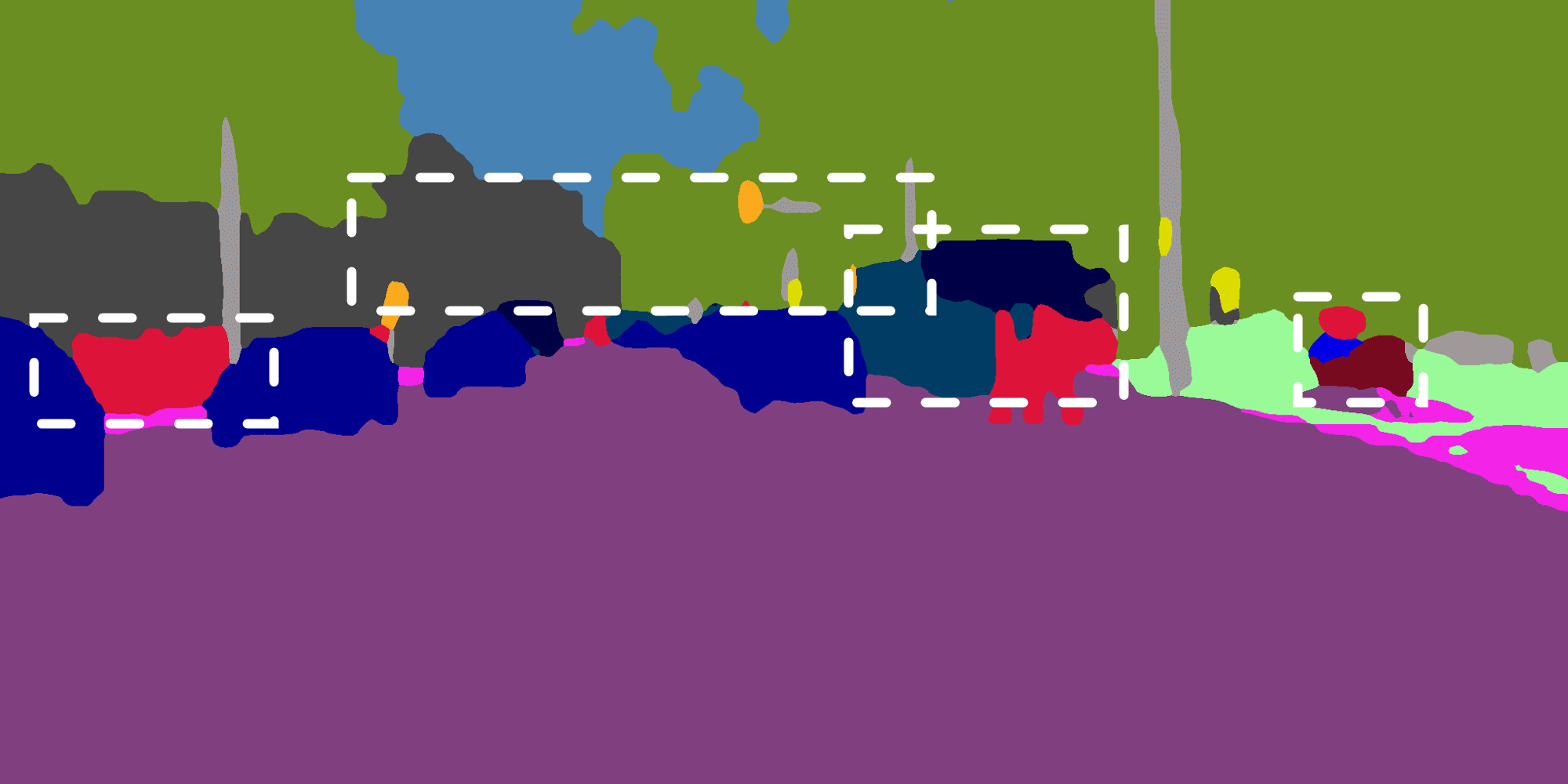}}
\vspace{-5pt}
\subfigure[DPL-Dual\cite{DPL_cheng2021dual}]{\includegraphics[width=0.235\linewidth]{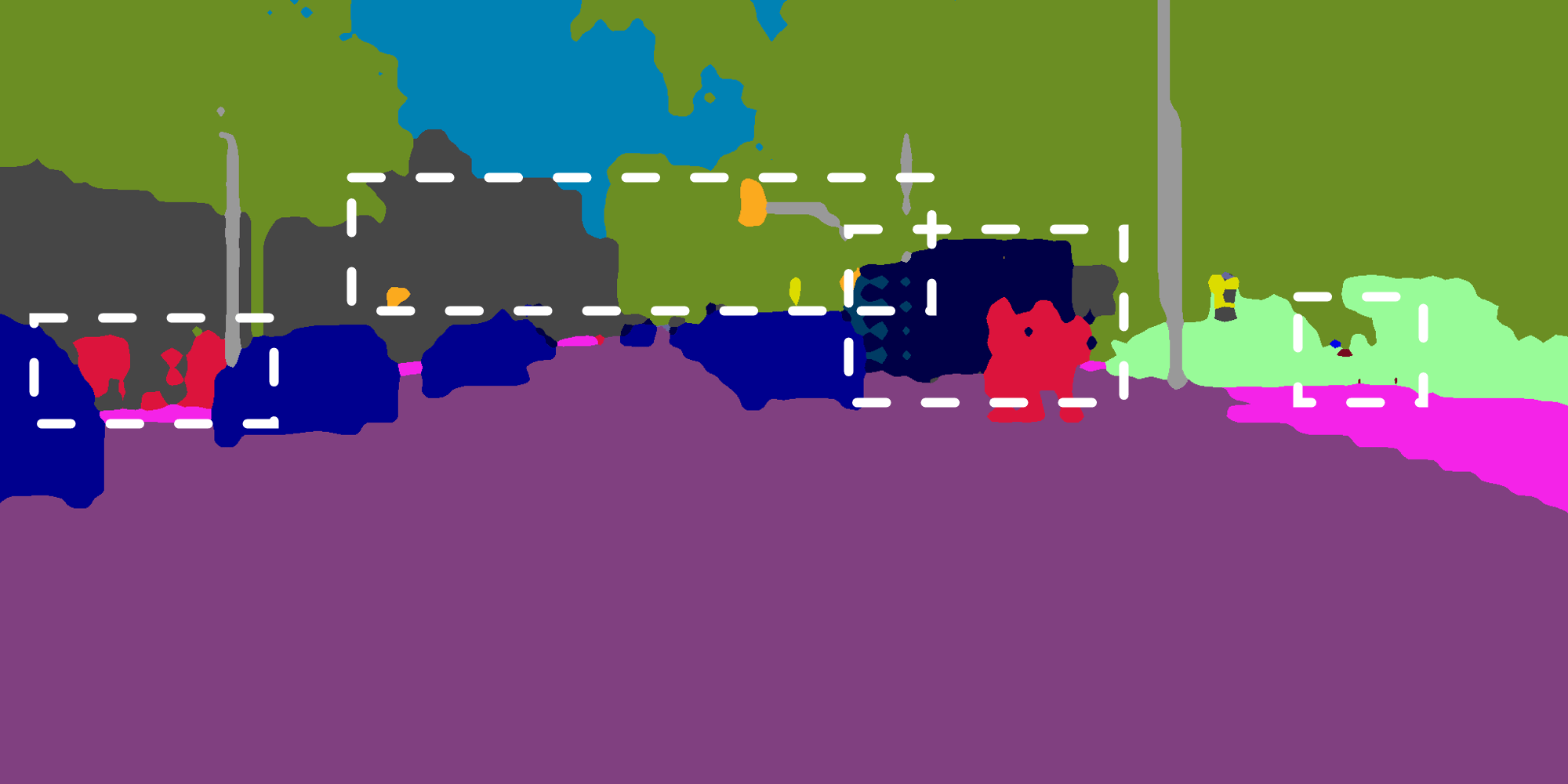}}
\subfigure[SAC\cite{SAC_araslanov2021self}]{\includegraphics[width=0.235\linewidth]{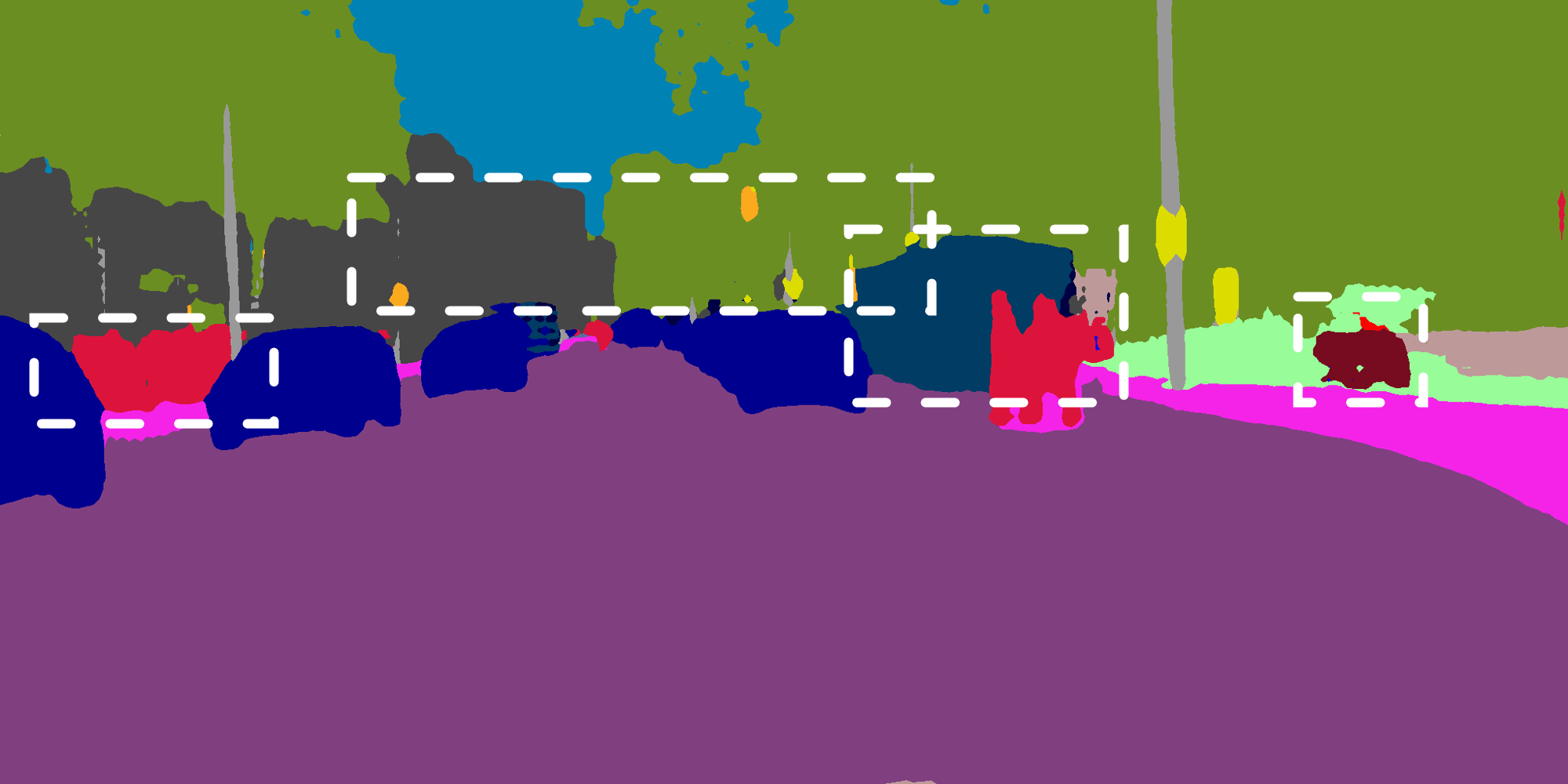}}
\subfigure[DSP\cite{DSP_gao2021dsp}]{\includegraphics[width=0.235\linewidth]{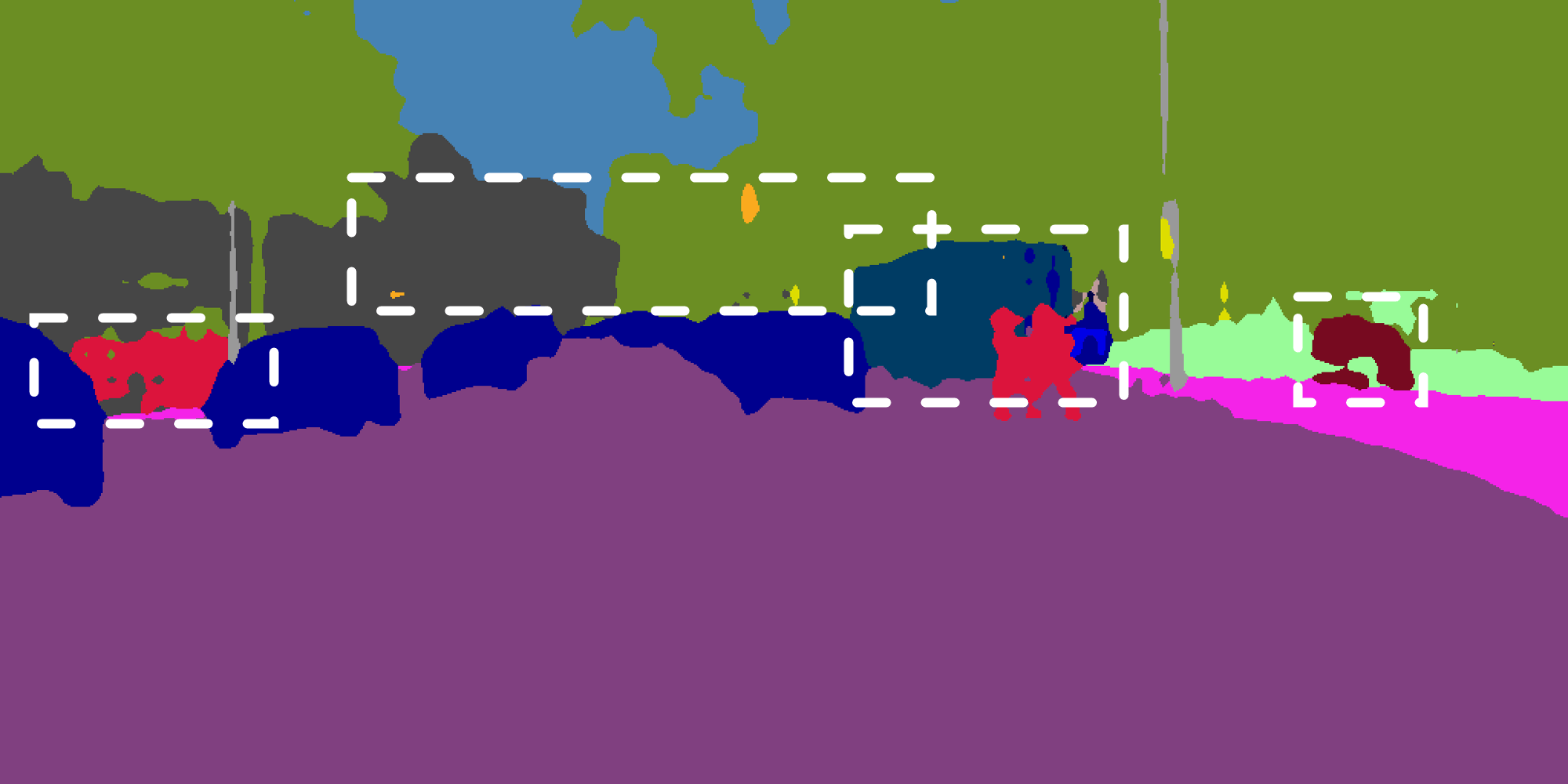}}
\subfigure[ProDA\cite{ProDA_CRA_wang2021cross}]{\includegraphics[width=0.235\linewidth]{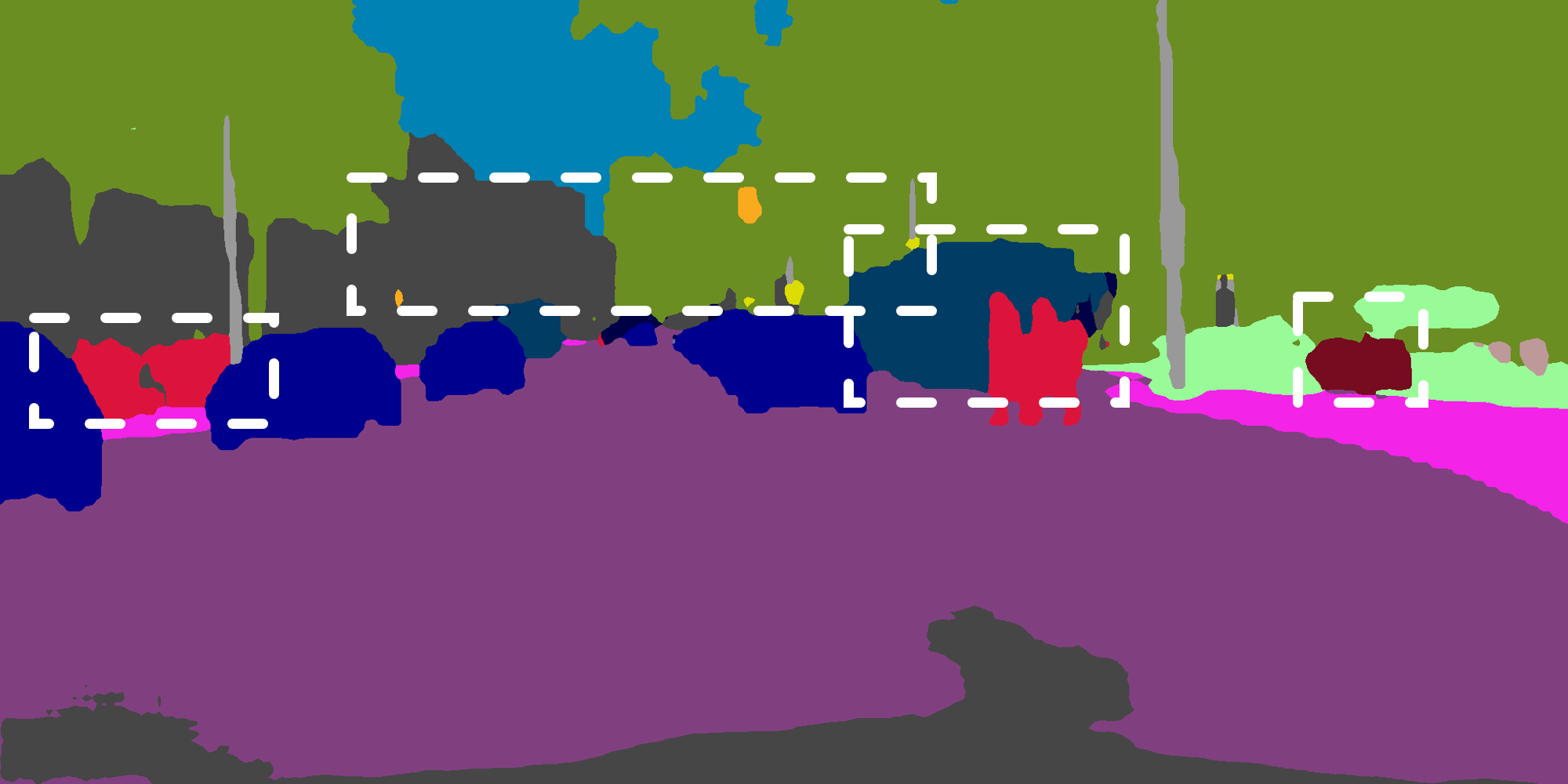}}
\subfigure[CPSL\cite{li2022class}]{\includegraphics[width=0.235\linewidth]{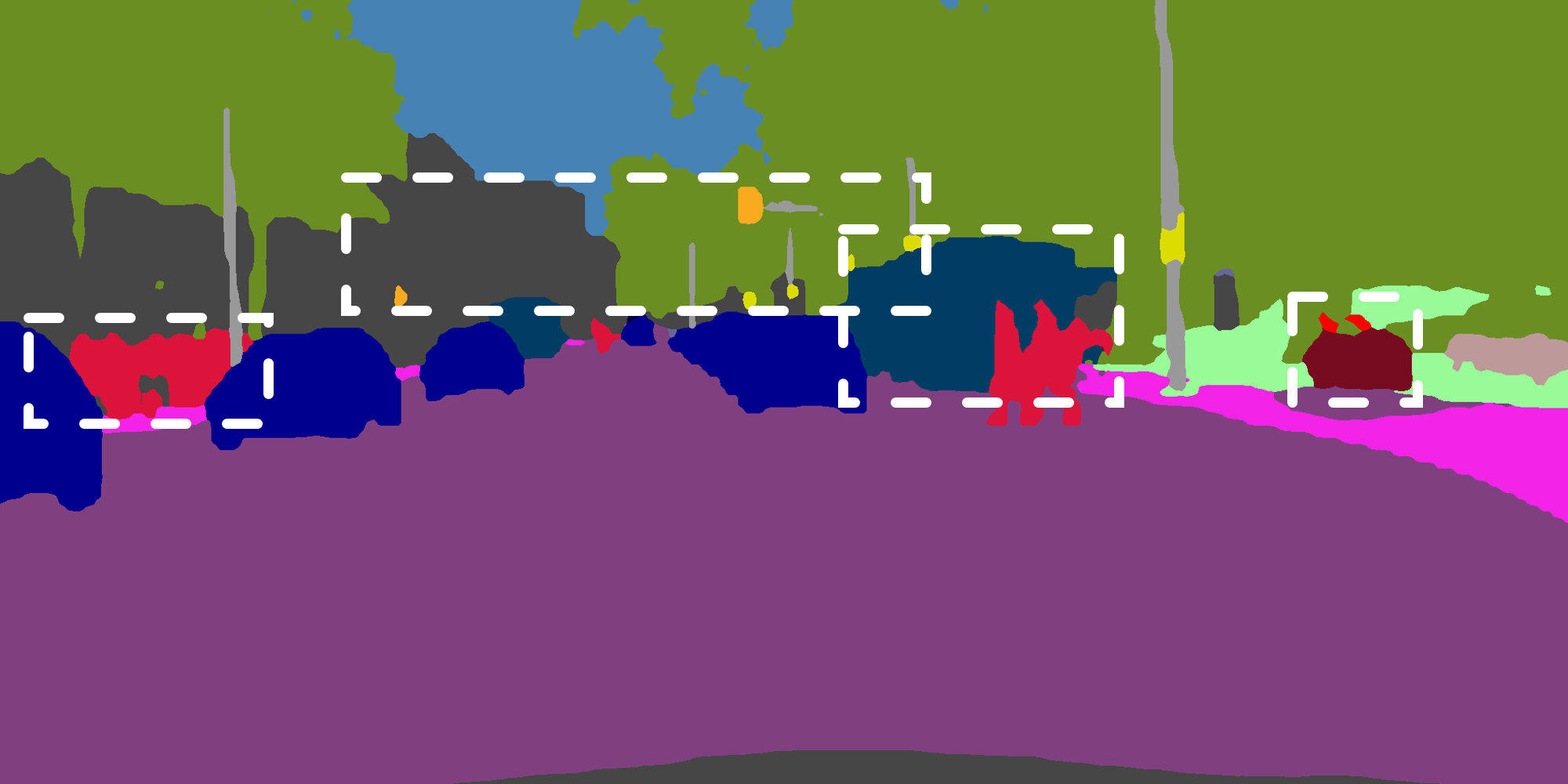}}
\subfigure[SePiCo\cite{xie2022sepico}]{\includegraphics[width=0.235\linewidth]{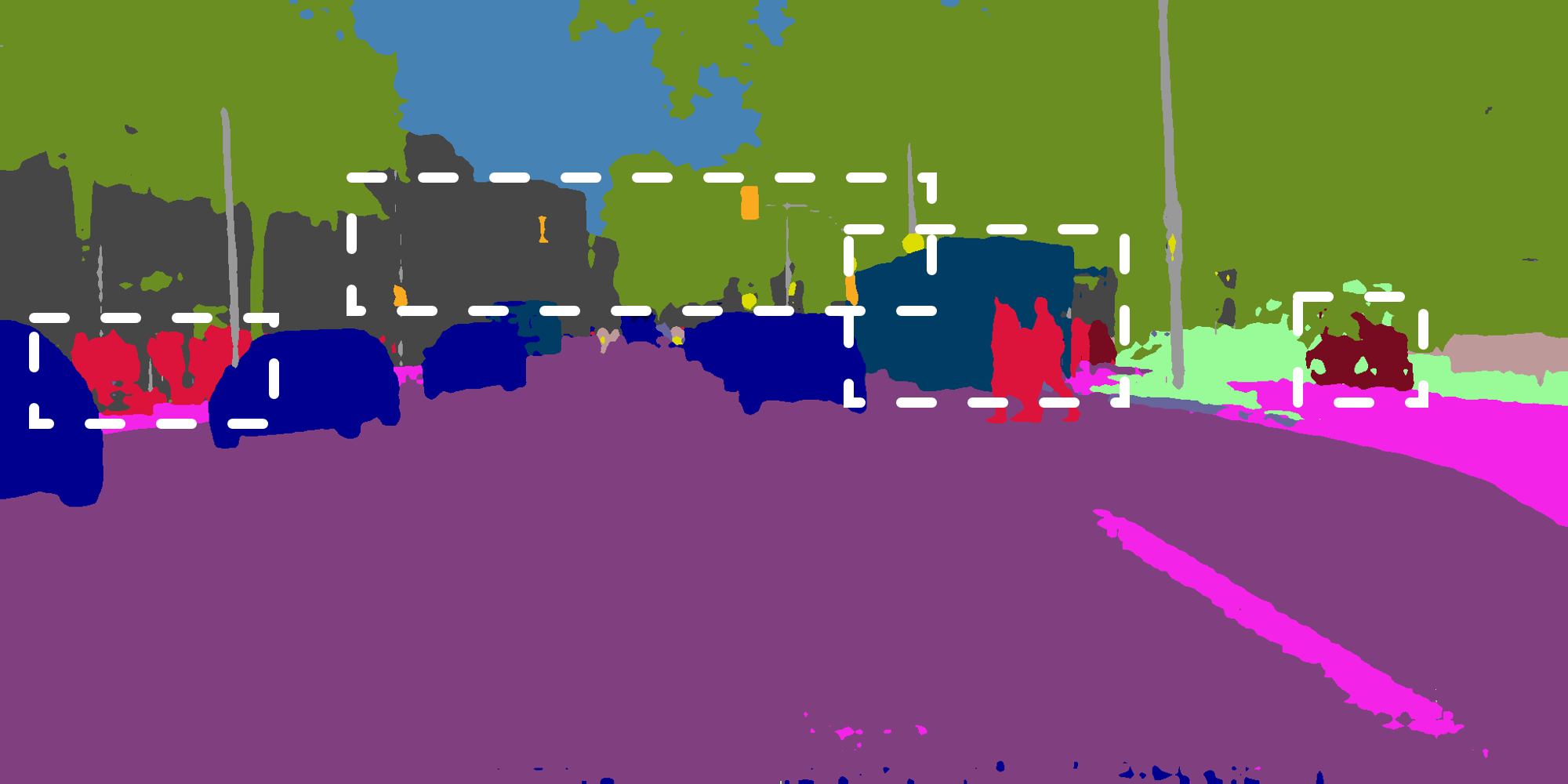}}
\subfigure[HIAST (Ours)]{\includegraphics[width=0.235\linewidth]{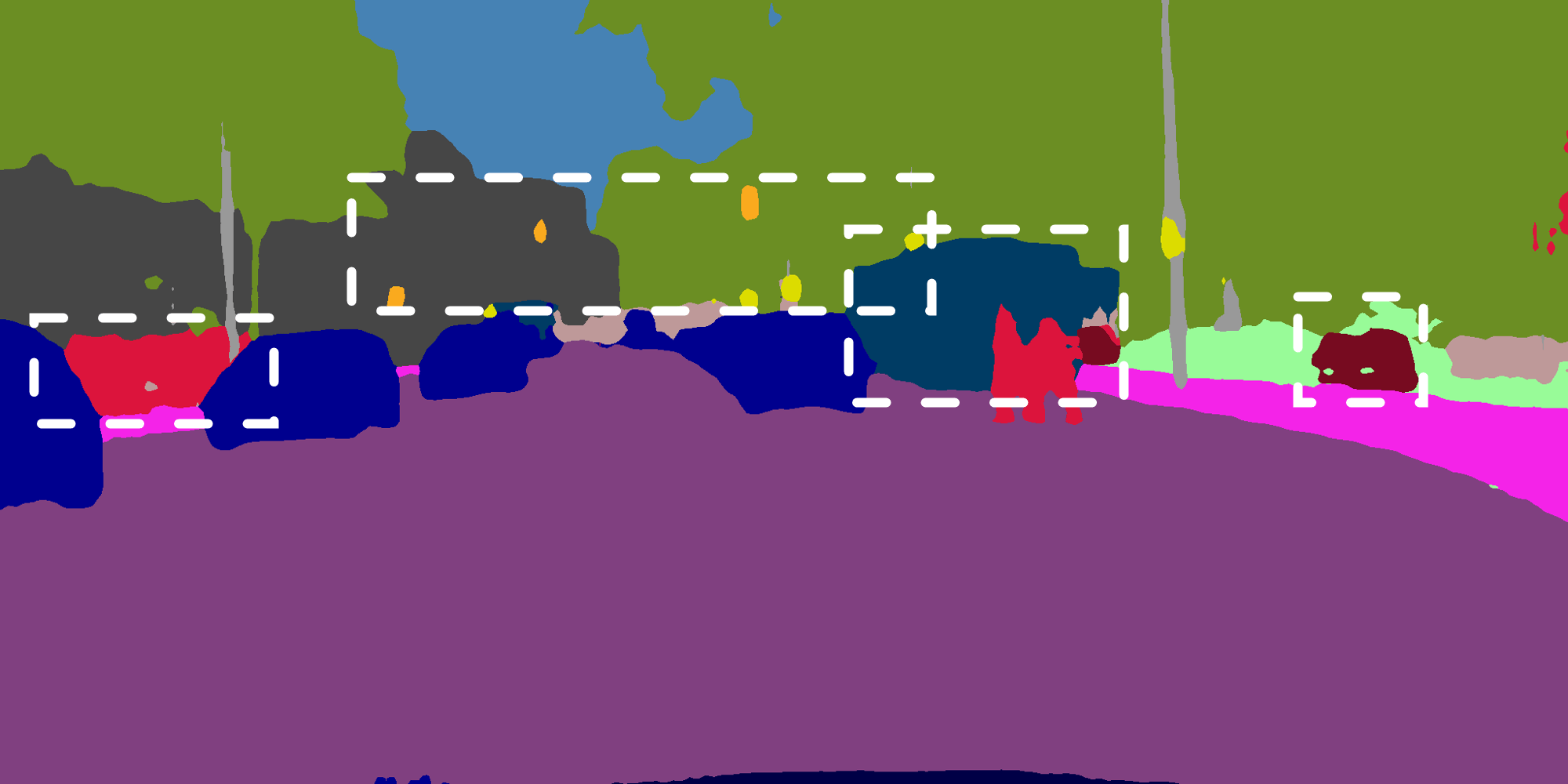}}
\subfigure[HIAST (Ours)$^{\nabla}$]{\includegraphics[width=0.235\linewidth]{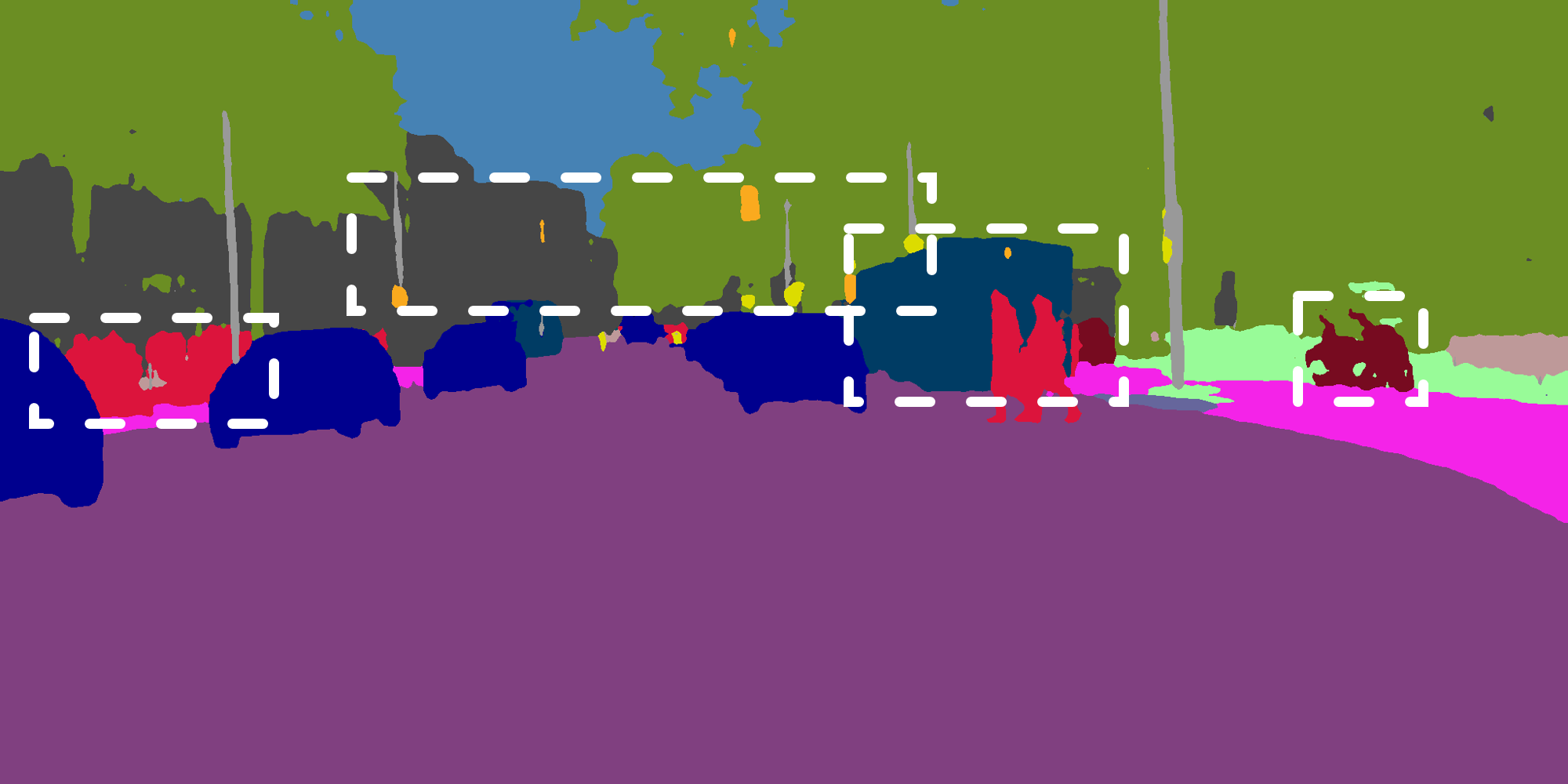}}
\vspace{-10pt}
\caption{{Qualitative comparisons of different methods on GTA5 $\rightarrow$ Cityscapes. The default warm-up model of HIAST is AdaptSeg, and $\nabla$ indicates that SePiCo is used for warm-up. Compared to other methods, HIAST has better segmentation performance for the hard class with small scale, such as traffic sign and traffic light. Furthermore, for the hard class which has a low occurrence frequency in the target domain, such as bus and bike, HIAST also has better performance.}}
\vspace{-5pt}
\label{fig:visualization_comparison_0}
\end{figure*}

\noindent\textbf{Implementation Details.} In our experiments, we implement HIAST by using PyTorch on a single NVIDIA Tesla V100 with 32 GB memory. 
During training for the synthetic-to-real scenario, images are randomly cropped and resized to $1024\times512$, and the height ranges for random cropping of GTA5, SYNTHIA, and Cityscapes are set to $[341\sim850]$, $[341\sim640]$, and $[341\sim1000]$, respectively. For the cross-city scenario, we set the resized size to $1024\times768$, and the height ranges for random cropping of Cityscapes and Oxford RobotCar are set to $[341\sim1000]$ and $[341\sim900]$, respectively.
We adopt Adam as the optimizer, and the learning rate is initialized to $3\times10^{-5}$ and modified by the cosine annealing scheduler. During multi-round self-training of HIAST, all weights of batch normalization layers are frozen, and each round lasts 8000 iterations with a batch size of 6. 
We set the pseudo-label parameters $\alpha$, $\beta$, $\gamma$ to $0.5$, $0.9$, $8.0$, and the regularization weights $\lambda_{i}$, $\lambda_{c}$, $\lambda_{cst}$ are set to $1.0$, $0.1$, $0.5$. 
For the synthetic-to-real scenario, $k$ is set to $14$ for GTA5 $\rightarrow$ Cityscapes and $10$ for SYNTHIA $\rightarrow$ Cityscapes. For the cross-city scenario, $k$ is set to $5$. The parameter $\tau_{\overline{\mathbf{w}}}$ for consistency constraint on the ignored region is set to $0.999$. We use standard resizing and random flip as the weak augmentaion. The strong augmentation of consistency constraint on the ignored region is implemented by RandAugment\cite{cubuk2020randaugment}, which randomly selects 3 style transformations from the candidate pool. {The hyper-parameters are set according to the pseudo-labels of randomly selected 500 training images from the target, and the details are shown in Section \ref{subsection: param_select}.} It should be noted that no ground truth labels are used in our hyper-parameters tuning. 

\begin{table}
\centering
\caption{{Results of applying our method to Transformers (GTA5 $\rightarrow$ Cityscapes).}}
\vspace{-7.5pt}
\begin{tabular}{lccc}
\toprule
Method & Baseline & +HIAST & Improvement \\ \midrule
DAFormer\cite{hoyer2022daformer} & 68.3 & 69.5 & +1.2 \\
HRDA\cite{hoyer2022hrda} & 73.8 & 74.5 & +0.7 \\
\bottomrule
\end{tabular}
\label{table:transformer}
\end{table}

\subsection{Comparison with the SOTAs}

\noindent\textbf{Synthetic-to-Real Adaptation.} The results of HIAST and some other state-of-the-art methods on GTA5 $\rightarrow$ Cityscapes are presented in Table \ref{table:gta5}.

With AdaptSeg\cite{tsai2018learning} as the warm-up model, HIAST has the mIoU of $56.3\%$, yielding the competitive performance. Our previous work IAST has achieved the mIoU of $51.5\%$. Based on this, HIAST has the gain of $4.8\%$ and almost has significant gains of performance over all classes, especially on such hard classes with small scale as pole, traffic light, and traffic sign, and also on rider, motorcycle, and bike which always have a low occurrence frequency in target domain dataset, verifying the effect of HPLA based on hard classes and consistency constrain on ignored regions. When the warm-up model is replaced by the latest model SePiCo\cite{xie2022sepico}, HIAST has the mIoU of $64.1\%$, which is higher than all previous methods. 
Additionally, we also evaluate the results under the transformer architecture based on DAFormer\cite{hoyer2022daformer} and HRDA\cite{hoyer2022hrda}. 
As shown in Table \ref{table:transformer}, it can be seen that by introducing our method, the two baselines obtain performance gains of 1.2\% and 0.7\%, respectively. This indicates our method has the potential for improvement with advanced transformer architectures.

Table \ref{table:synthia} shows the results of SYNTHIA $\rightarrow$ Cityscapes. For a comprehensive comparison, as in the previous work, we also report two mIoU metrics: $16$ classes of mIoU and $13$ classes of mIoU* with wall, fence, and pole excluded. The domain gap between SYNTHIA and Cityscapes is much larger than the domain gap between GTA5 and Cityscapes. Many of the methods that perform well on GTA5 $\rightarrow$ Cityscapes have experienced a significant performance degradation on this task. {Our method with AdaptSeg achieves $53.5\%$ mIoU and $60.3\%$ mIoU*.} Moreover, HIAST with SePiCo as warm-up also achieves the best results of $59.6\%$ mIoU and $68.1\%$ mIoU*, which are significantly higher than all recent state-of-the-art methods.

\noindent\textbf{Cross-City Adaptation.} In addition, we also evaluate HIAST on Cityscapes $\rightarrow$ Oxford RobotCar. Under this scenario, both source and target domain images are collected from the real-world scene with a large domain gap on the weather conditions. Concretely, the Cityscapes dataset consists of sunny scenes, however, images in the Oxford RobotCar dataset are almost rainy, making it challenging for cross-city adaptation. Following \cite{Tsai_adaptseg_ICCV19}, the results of 9 shared classes are reported in Table \ref{table:oxford}. HIAST achieves the best mIoU of $75.2\%$ and has a gain of $2.0\%$ compared with IAST, with almost significant improvement over all classes, especially on light and sign which are classes with small scale and low occurrence frequency in the Oxford RobotCar dataset.

\noindent\textbf{Visualization.} Besides, we have provided the visualization results of all three UDA scenarios for qualitative analysis. As shown in Fig. \ref{fig:visualization_prediction}, HIAST has better performance on pole, traffic light, and traffic sign which always have a small scale, however, IAST always confuses them with their nearby classes. In terms of such hard classes with low occurrence frequency as bike, rider, and bus, HIAST also has better performance, but IAST has fewer correct predictions for these classes.

\begin{table*}[htb]
\centering
\caption{Ablation study (GTA5 $\rightarrow$ Cityscapes). The self-training phase contains 3 rounds and pseudo-labels are re-generated at each round.}
\vspace{-7.5pt}
\begin{tabular}{c|l|ccccc|cc}  
\toprule
Phase & Method & IAS & $\mathcal{R}_c$ & $\mathcal{R}_i$ & HPLA & $\mathcal{R}_{cst}$ & mIoU & $\Delta$ \\
\midrule
\multirow{2}{*}{Initialization} & Source-only &   &   &   &   &   & 35.6 & - \\
 & Warm-up &   &   &   &   &   & 43.8 & +8.2 \\
\midrule
\multirow{5}{*}{\shortstack{Self-training\\(3 rounds)}} & Instance adaptive selector (IAS) & \ding{51} &   &   &   &    & 49.5 & +5.7\\
 & Confident region KLD minimization ($\mathcal{R}_c$) & \ding{51} & \ding{51} &   &   &   & 50.5 & +1.0\\
 & Ignored region entropy minimization ($\mathcal{R}_i$) & \ding{51} & \ding{51} & \ding{51} &   &   & 51.5 & +1.0\\
 & Hard-aware pseudo-label augmentation (HPLA) & \ding{51} & \ding{51} & \ding{51} & \ding{51} &   & 55.7 & +4.2\\
 & Ignored region consistency constraint ($\mathcal{R}_{cst}$) & \ding{51} & \ding{51} & \ding{51} & \ding{51} & \ding{51} & 56.3 & +0.6\\
\bottomrule
\end{tabular}
\label{table:ablation}
\end{table*}

Furthermore, the qualitative comparisons between HIAST and other recent methods are also provided in Fig. \ref{fig:visualization_comparison_0}. Overall, HIAST has more accurate segmentation performance for all classes, especially on such hard classes as traffic light, traffic sign, bike, and bus, however, other methods suffer from seriously incorrect predictions on these classes.

\subsection{Ablation Study}
\label{subsection: ablation_study}
Results of the ablation study are reported in Table \ref{table:ablation}. We add the methods proposed in Section \ref{subsection:p}, Section \ref{subsection: HPLA}, and Section \ref{subsection: r} sequentially to study their performance in the validation set under the UDA scenario of GTA5 $\rightarrow$ Cityscapes.

\begin{table}[htb]
\centering
\caption{{Ablation study based on the switching-off policy (GTA5 $\rightarrow$ Cityscapes), where proposed modules are deactivated independently. $\star$ indicates the EMA in IAS module.}}
\vspace{-7.5pt}
\setlength\fboxsep{0pt}
\begin{tabular}{l|c|l}
\toprule
Configuration & mIoU & \multicolumn{1}{>{\centering\arraybackslash}m{18mm}}{$\Delta$} \\
\midrule
No IAS & 51.6 & \cbar{36}{5} -4.7 \\
{No EMA$^\star$} & 55.2 & \cbar{8}{5} -1.1 \\
No HPLA & 52.0 & \cbar{30}{5} -4.3 \\
No $\mathcal{R}_i$ & 53.9 & \cbar{18}{5} -2.4 \\
No $\mathcal{R}_c$ & 55.0 & \cbar{9}{5} -1.3 \\
No $\mathcal{R}_{cst}$ & 55.7 & \cbar{5.6}{5} -0.6 \\
\midrule
Full HIAST & 56.3 & - \\
\bottomrule
\end{tabular}
\vspace{-10pt}
\label{table:ablation_switch_off}
\end{table}

As shown in Table \ref{table:ablation}, after gradually adding IAS, $\mathcal{R}_c$, $\mathcal{R}_i$, HPLA, and $\mathcal{R}_{cst}$, the performance is progressively improved. The IAS has a significant gain of $5.7\%$, verifying the effectiveness of the proposed pseudo-label generation strategy, which will be beneficial for the subsequent self-training phase. Regularization $\mathcal{R}_c$, $\mathcal{R}_i$ have equal gains of $1.0\%$, indicating that smoothing predictions of the confident region and sharpening predictions of the ignored region are efficient. Furthermore, HPLA has a great gain of $4.2\%$, suggesting that our copy-and-paste strategy concentrated on hard classes can significantly improve corresponding performance. Besides, the gain of $0.6\%$ brought by consistency constraint on the ignored region has proved that consistency constraint can promote model learning on the ignored region, thus producing better adaptation performance.
Finally, HIAST achieves a great result of $56.3\%$ mIoU.

Furthermore, to more fairly demonstrate the separate contributions of the proposed modules, we conduct another ablation study based on the switching-off policy, where the proposed modules are deactivated independently. As shown in Table \ref{table:ablation_switch_off}, disabling IAS leads to a significant mIoU decrease of 4.7\%, also indicating the largest contribution to our method. Moreover, we further ablate the EMA strategy in IAS, which results in a 1.1\% decrease in performance. Abolishing HPLA has brought the mIoU decrease of 4.3\%, demonstrating the next largest contribution to our method. Canceling $\mathcal{R}_i$, $\mathcal{R}_{c}$, and $\mathcal{R}_{cst}$, have resulted in mIoU degenerations of 2.4\%, 1.3\%, and 0.6\% respectively.

\begin{table*}[htb]
\centering
\caption{{Results of parameters selection (GTA5 $\rightarrow$ Cityscapes). F-mIoU means the result by fusion pseudo-labels. The parameters corresponding to bold F-mIoU mean the selected parameters by our method, and the parameters corresponding to bold mIoU mean the optimal parameters.}}
\vspace{-7.5pt}
\label{table:params_results}
\begin{tabular}{ccc|ccc|ccc|ccc|ccc}
\toprule
${\alpha}$ & F-mIoU & mIoU & ${\gamma}$ & F-mIoU & mIoU & ${\lambda}_i$& F-mIoU & mIoU & ${\lambda}_{cst}$ & F-mIoU & mIoU & $k$ & F-mIoU & mIoU   \\
\midrule
0.10 & 85.4 & 53.5 & 1 & 88.4& 56.0 & 0.1& 84.5& 55.6 & 0.1& 95.8& 56.1 & 12& 91.1 & 55.7 \\
0.30 & 92.4 & 55.9 & 4 & 93.9& 56.3 & 0.5& 92.9& 55.8 & 0.5& \textbf{96.5}& \textbf{56.3} & 14& \textbf{93.9} & \textbf{56.3} \\
0.50 & \textbf{95.2} & 56.3 & 8 & \textbf{96.2}& 56.3 & 1.0& \textbf{96.6}& 56.3 & 1.0& 94.8& 56.2 & 16& 89.6 & 56.0 \\
0.70 & 83.5 & \textbf{56.5} & 16 & 91.8& \textbf{56.5} &  5.0& 91.5 & \textbf{56.4} & 5.0& 91.4& 56.0 & & \\
0.90 & 71.3 & 56.2 & 32 & 86.0& 56.2 & 10.0& 87.7& 56.1 & 10.0& 83.6& 55.6 & & \\
\bottomrule
\end{tabular}
\vspace{-8pt}
\end{table*}

\begin{table}[htb]
\centering
\caption{$\alpha$ selection with different sizes $N$ of $I_{subset}$ (GTA5 $\rightarrow$ Cityscapes). The parameters corresponding to bold F-mIoU mean the selected parameters.}
\vspace{-7.5pt}
\label{table:different_size}
\begin{tabular}{c|cccc|c}
\toprule 
\multirow{2}{*}{$\alpha$\,$\backslash$\,{N}} & \multicolumn{4}{c|}{F-mIoU} & \multirow{2}{*}{mIoU} \\
 \cline{2 - 5} & \rule{0pt}{1.0em}50 & 100 & 500 & 1000 & \\
\midrule
0.1 & 83.1 & 85.4 & 85.4 & 84.7 & 53.5 \\
0.3 & 90.1 & 92.1 & 92.4 & 92.1 & 55.9 \\
0.5 & \textbf{95.5} & \textbf{95.4} & \textbf{95.2} & \textbf{95.3} & 56.3 \\
0.7 & 79.0 & 82.4 & 83.5 & 83.1 & \textbf{56.5} \\
0.9 & 67.0 & 69.6 & 71.3 & 70.9 & 56.2 \\
\bottomrule 
\end{tabular}
\vspace{-5pt}
\end{table}

\begin{table}[htb]
\centering
\caption{$\alpha$ selection with different 500 images generated from four random seeds (GTA5 $\rightarrow$ Cityscapes). The parameters corresponding to bold F-mIoU mean the selected parameters.}
\vspace{-7.5pt}
\label{table:different_seed}
\begin{tabular}{c|cccc|c}
\toprule 
\multirow{2}{*}{$\alpha$\,$\backslash$\,\text{Seed}}  & \multicolumn{4}{c|}{F-mIoU} & \multirow{2}{*}{mIoU} \\
 \cline{2 - 5} & \rule{0pt}{1.0em}Seed-1 & Seed-2 & Seed-3 & Seed-4 & \\
\midrule
0.1 & 85.4 & 84.9 & 84.6 & 83.4 & 53.5 \\
0.3 & 92.4 & 92.3 & 92.6 & 91.5 & 55.9 \\
0.5 & \textbf{95.2} & \textbf{95.6} & \textbf{95.4} & \textbf{94.8} & 56.3 \\
0.7 & 83.5 & 84.1 & 83.3 & 82.2 & \textbf{56.5} \\
0.9 & 71.3 & 72.5 & 71.7 & 69.2 & 56.2 \\
\bottomrule 
\end{tabular}
\vspace{-5pt}
\end{table}

\subsection{Parameter Analysis}
\noindent{\textbf{Parameter selection for our method.} As shown in Table \ref{table:params_results}, each parameter has a series of candidate values, and we have selected each parameter using the method mentioned in Section \ref{subsection: param_select}. It can be seen that our selected parameters are the optimal or sub-optimal ones, which demonstrates the effectiveness of our parameter selection proposed in this paper.} 
To further explore the impact of different $I_{subset}$ on the parameter selection, we have conducted the following two experiments: (1) parameter selection with different sizes; (2) parameter selection with different random seeds. The results of $\alpha$ selection are shown in Table \ref{table:different_size} and Table \ref{table:different_seed}, and the results of other parameters can be found in Appendix A. It can be seen that the results of our parameter selection are stable under different configurations.

\noindent\textbf{Parameter Sensitivity of IAS.}{ Table \ref{table:alpha} shows a sensitivity analysis on parameter $\alpha$. It has been shown that values between 0.3 and 0.9 give consistently fantastic performance, which is a fairly wide range for robust parameter selection.} 

\begin{table}[htb]
\centering
\caption{$\alpha$ and $\gamma$ sensitivity analysis (GTA5 $\rightarrow$ Cityscapes). P-proportion means the proportion of selected pseudo-labels. \underline{Underline} means the results selected by our hyper-parameters method.}
\vspace{-7.5pt}
\begin{tabular}{cccc}
\toprule
$\alpha$ & $\gamma$ & P-proportion & mIoU 
        \\\midrule
\cellcolor{lightgray!50}.10                          & 8                          & 20.6                               & 53.5                         \\
\cellcolor{lightgray!50}.30                          & 8                          & 35.2                               & 55.9                  \\
\cellcolor{lightgray!50}.50                          & 8                          & 44.1                               & \underline{56.3}
        \\
\cellcolor{lightgray!50}.70                          & 8                          & 49.7                               & \textbf{56.5}  
        \\
\cellcolor{lightgray!50}.90                          & 8                          & 53.5                               & 56.2  
        \\ \midrule
.50                          & \cellcolor{lightgray!50}1                         & 52.8                               & 56.0                         \\
.50                          & \cellcolor{lightgray!50}4                          & 47.7                               & 56.3                         \\
.50                          & \cellcolor{lightgray!50}8                         & 44.1                               & \underline{56.3}                   \\
.50                          & \cellcolor{lightgray!50}16                         & 39.6                               & \textbf{56.5}    \\
.50                          & \cellcolor{lightgray!50}32                            & 34.9                               & 56.2          
\\ \bottomrule
\end{tabular}
\label{table:alpha}
\end{table}

\begin{figure}[htb]
\centering
\includegraphics[width=\linewidth]{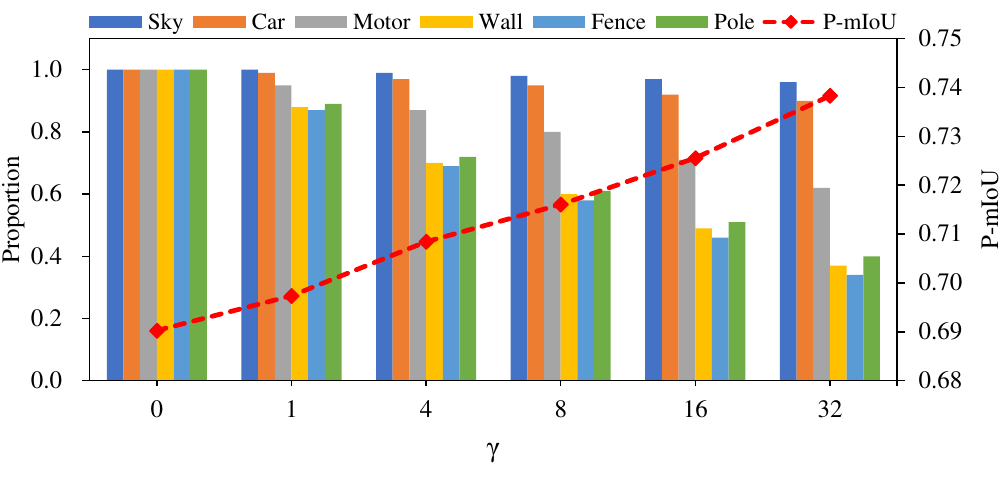}
\vspace{-20pt}
\caption{Relationship between the class proportions of selected pseudo-labels, mIoU of selected pseudo-labels (P-mIoU) and $\gamma$.}
\vspace{-10pt}
\label{fig:bar_IAS}
\end{figure}

Then, the influence of $\gamma$ has been studied by fixing other parameters. With class proportions of pseudo-labels when $\gamma=0$ as unit one, Fig. \ref{fig:bar_IAS} shows that as $\gamma$ increases, the proportions of some easy classes (sky, car) that have a high predicted score do not decrease significantly, while the proportions of some hard classes (motorcycle, wall, fence, and pole) that have a low predicted score decrease sharply, meanwhile the mIoU of selected pseudo-labels (P-mIoU) is gradually improved. This proves that the noise mainly exists in hard classes, and Eq.\eqref{eq:wd} can effectively suppress noise interference on pseudo-labels. Table \ref{table:alpha} also shows a sensitivity analysis on parameter $\gamma$, indicating that the performance is not sensitive to this parameter.

\noindent\textbf{Parameter Sensitivity of HPLA.} We have also investigated the influence of parameter $k$ based on the best setting of HIAST. {According to previous researches\cite{zou2018unsupervised,tsai2018learning,mei2020instance}, we found that there are 14 classes whose performance is poor, so we set the range of $k$ between 12 and 16. As shown in Table \ref{table:HPLA_params_sensitivity_analysis}, it is sensitive because the number of hard classes is an important parameter in the pseudo-label augmentation.}

\begin{table}[htb]
\centering
\caption{$k$ sensitivity analysis of HPLA (GTA5 $\rightarrow$ Cityscapes). \underline{Underline} means the results selected by our hyper-parameters method.}
\vspace{-7.5pt}
\label{table:HPLA_params_sensitivity_analysis}
\begin{tabular}{ccc}
\toprule
$k$ & mIoU     \\
\midrule
12 & 55.7          \\
14 & \underline{\textbf{56.3}} \\
16 & 56.0 \\
\bottomrule
\end{tabular}
\end{table}

\noindent\textbf{Parameter Sensitivity of Region-adaptive Constraints.} {For KLD minimization on the confident region we fellow the setting in CRST\cite{zou2019confidence} and set it to 0.1.} For entropy minimization on the ignored region, Table \ref{table:lambda} shows a sensitivity analysis of parameters $\lambda_i$. We perform multiple sets of experiments with fixed $\lambda_c$ respectively. {When $\lambda_i$ is gradually increased, the overall model performance tends to be improved until $\lambda_i$ is between 1 and 10.} It can be shown that when the low entropy prediction is insufficiently performed on the non-pseudo-label region, the influence of noise will not be suppressed and the model training will be damaged.

The weight of consistency constraint $\lambda_{cst}$ has also been studied. It is robust to a wide numerical range if $\lambda_{cst}$ is not too large as shown in Table \ref{table:lambda}. 

\begin{table}[htb]
\centering
\caption{$\lambda_i$, and $\lambda_{cst}$ sensitivity analysis (GTA5 $\rightarrow$ Cityscapes). \underline{Underline} means the results selected by our hyper-parameters method.}
\vspace{-7.5pt}
\begin{tabular}{ccc}
\toprule
$\lambda_i$ &  $\lambda_{cst}$ & mIoU\ \\ \midrule
\cellcolor{lightgray!50}0.1                               & 0.5        & 55.6                         \\
\cellcolor{lightgray!50}0.5                               & 0.5        & 55.8                         \\
\cellcolor{lightgray!50}1.0                           & 0.5        & \underline{56.3}                         \\
\cellcolor{lightgray!50}5.0                           & 0.5        & \textbf{56.4}                        \\
\cellcolor{lightgray!50}10.0                          & 0.5        & 56.1                         \\

\midrule
1.0                   & \cellcolor{lightgray!50}0.1  & 56.1                         \\ 
1.0                 & \cellcolor{lightgray!50}0.5  & \underline{\textbf{56.3}}                \\ 
1.0                      & \cellcolor{lightgray!50}1.0  & {56.2}                        
\\ 
1.0             & \cellcolor{lightgray!50}5.0  & 56.0                     
\\ 
1.0            & \cellcolor{lightgray!50}10.0  & 55.6                         \\ 
\bottomrule
\end{tabular}
\label{table:lambda}
\end{table}

\subsection{Extensions and Limitations}
\label{subsection:Extensions}
{\noindent\textbf{Apply to Other UDA Methods.} Because HIAST has no special structure or model dependency, it can be directly used to decorate other UDA methods. We have deployed our method on recent adversarial training methods\cite{tsai2018learning, Haoran_2020_ECCV, BCDM_Li2021BiClassifierDM} and self-training methods\cite{ProDA_zhang2021prototypical,li2022class,xie2022sepico}. As shown in Table \ref{table:warmup_gta5}, these methods have been significantly improved under our HIAST framework. 
{However, it is worth noting that some stronger self-training baselines, such as ProDA~\cite{ProDA_zhang2021prototypical} and CPSL~\cite{li2022class}, select all pseudo-labels for model training and use additional modules to correct the noisy pseudo-labels; but our method is to avoid selecting these pseudo-label areas with serious noise to ensure the quality of model training. Therefore, it is different that the two types of methods deal with noisy pseudo-labels, which could be considered as a potential conflict, limiting the further improvement of our method.
}}

\noindent\textbf{Extend to SSL.} The proposed method can also be applied to the semi-supervised semantic segmentation task. Following the configuration in \cite{hung2019adversarial}, we apply our method to Cityscapes for semi-supervised training with different proportions of data as labeled data. Specifically, IAST and HIAST are deployed based on the ``Labeled Only'', and HPLA is performed only on unlabelled data for HIAST. As shown in Table \ref{table:semi}, we have better performances than recent SSL methods\cite{hung2019adversarial ,ClassMix_olsson2021classmix, ReCo_liu2021bootstrapping, GuideMix_tu2021guidedmix}.

\noindent\textbf{Limitations.} Although HIAST has significantly improved the performance of hard classes and achieved great success on semantic segmentation for not only UDA but also SSL, the designed copy-and-paste strategy is suboptimal due to the performing of direct copying, which may ruin some original confident regions in pseudo-labels. Considering that there exist many ignored regions in pseudo-labels, future works can focus on how to implement copy-and-paste adaptively based on ignored regions to avoid the sacrifice of confident regions. Besides, the introduction of noise is inevitable during pseudo-label generation, and generic solutions to filter out noisy areas will decrease available training data, hence there need effective ways to optimize the noise and preserve more available pseudo-labels.

\begin{table}[htb]
\centering
\caption{{Results of applying HIAST to different UDA methods (GTA5 $\rightarrow$ Cityscapes). Only one round of self-training is performed.}}
\vspace{-7.5pt}
\begin{tabular}{lccc}
\toprule
Method & Baseline & +HIAST & Improvement \\ \midrule
AdaptSeg\cite{tsai2018learning} & 42.4 & 53.6 & +11.2 \\
FADA\cite{Haoran_2020_ECCV}  & 46.9 & 55.4 & +8.5 \\
BCDM\cite{BCDM_Li2021BiClassifierDM}  & 46.6 & 56.5 & +9.9 \\
ProDA\cite{ProDA_zhang2021prototypical} & 57.5 & 60.1 & +2.6 \\
CPSL\cite{li2022class} & 60.8 & 63.0 & +2.2 \\
SePiCo\cite{xie2022sepico} & 61.0 & 63.7 & +2.7 \\
\bottomrule
\end{tabular}
\label{table:warmup_gta5}
\end{table}

\begin{table}[htb]
\centering
\caption{Results of applying our method to semi-supervised semantic segmentation on the Cityscapes validation set. 1/8, 1/4, and 1/2 mean the proportions of labeled images, and the numbers of labeled images are shown inside the parentheses.}
\vspace{-7.5pt}
\begin{tabular}{lccc}
\toprule
\multirow{2}{*}{Method} & \multicolumn{3}{c}{Labeled Proportion} \\ \cmidrule{2-4} 
                        & \multicolumn{1}{c}{1/8 (371)} & \multicolumn{1}{c}{1/4 (743)} & \multicolumn{1}{c}{1/2 (1487)} \\ \midrule
AdvSemi\cite{hung2019adversarial}        & 58.8                    & 62.3                    & 65.7 \\
ClassMix\cite{ClassMix_olsson2021classmix} & 61.4 & 63.6 & 66.3 \\
ReCo\cite{ReCo_liu2021bootstrapping} & 64.9 & 67.5 & 68.7 \\
GuidedMix\cite{GuideMix_tu2021guidedmix} & 65.8 & 67.5 & 69.8 \\ \midrule
Labeled Only           & 57.3                    & 59.0                    & 61.2 \\
IAST (ours)                    & 64.6                    & 66.7                    & 69.8 \\
HIAST (ours) & \textbf{68.1} & \textbf{70.1} & \textbf{70.3} \\ \bottomrule
\end{tabular}
\label{table:semi}
\end{table}

\section{Conclusion}
\label{sec:conclusion}
In this paper, we propose a hard-aware instance adaptive self-training framework for UDA semantic segmentation. Compared with other popular UDA methods, HIAST still has a significant improvement in performance. Moreover, HIAST is a method with no model or special structure dependency, which means that it can be easily applied to other UDA methods with almost no additional cost to improve performance. In addition, HIAST can also be applied to the semi-supervised semantic segmentation task, which also achieves state-of-the-art performance. We hope this work will prompt people to rethink the potential of self-training on UDA or SSL tasks.

\section{Acknowledgement}
This work was supported in part by the Natural Science Foundation of Beijing Municipality under Grant 4182044, and in part by the National Natural Science Foundation of China (61602011).

\ifCLASSOPTIONcaptionsoff
  \newpage
\fi



%



\bibliographystyle{IEEEtran}
\bibliography{IEEEabrv,IEEEexample}

\begin{thebibliography}{10}
\providecommand{\url}[1]{#1}
\csname url@samestyle\endcsname
\providecommand{\newblock}{\relax}
\providecommand{\bibinfo}[2]{#2}
\providecommand{\BIBentrySTDinterwordspacing}{\spaceskip=0pt\relax}
\providecommand{\BIBentryALTinterwordstretchfactor}{4}
\providecommand{\BIBentryALTinterwordspacing}{\spaceskip=\fontdimen2\font plus
\BIBentryALTinterwordstretchfactor\fontdimen3\font minus
  \fontdimen4\font\relax}
\providecommand{\BIBforeignlanguage}[2]{{%
\expandafter\ifx\csname l@#1\endcsname\relax
\typeout{** WARNING: IEEEtran.bst: No hyphenation pattern has been}%
\typeout{** loaded for the language `#1'. Using the pattern for}%
\typeout{** the default language instead.}%
\else
\language=\csname l@#1\endcsname
\fi
#2}}
\providecommand{\BIBdecl}{\relax}
\BIBdecl

\bibitem{chen2017deeplab}
L.-C. Chen, G.~Papandreou, I.~Kokkinos, K.~Murphy, and A.~L. Yuille, ``Deeplab:
  Semantic image segmentation with deep convolutional nets, atrous convolution,
  and fully connected crfs,'' \emph{IEEE transactions on pattern analysis and
  machine intelligence}, vol.~40, no.~4, pp. 834--848, 2017,
  \mbox{doi}:\url{10.1109/TPAMI.2017.2699184}.

\bibitem{DeepLabv3_chen2017rethinking}
L.-C. Chen, G.~Papandreou, F.~Schroff, and H.~Adam, ``Rethinking atrous
  convolution for semantic image segmentation,'' \emph{arXiv preprint
  arXiv:1706.05587}, 2017.

\bibitem{chen2018encoder}
L.-C. Chen, Y.~Zhu, G.~Papandreou, F.~Schroff, and H.~Adam, ``Encoder-decoder
  with atrous separable convolution for semantic image segmentation,'' in
  \emph{ECCV}, 2018, pp. 801--818.

\bibitem{chen2018domain}
Y.~Chen, W.~Li, C.~Sakaridis, D.~Dai, and L.~Van~Gool, ``Domain adaptive faster
  r-cnn for object detection in the wild,'' in \emph{CVPR}, 2018, pp.
  3339--3348.

\bibitem{tsai2018learning}
Y.-H. Tsai, W.-C. Hung, S.~Schulter, K.~Sohn, M.-H. Yang, and M.~Chandraker,
  ``Learning to adapt structured output space for semantic segmentation,'' in
  \emph{CVPR}, 2018, pp. 7472--7481.

\bibitem{luo2019taking}
Y.~Luo, L.~Zheng, T.~Guan, J.~Yu, and Y.~Yang, ``Taking a closer look at domain
  shift: Category-level adversaries for semantics consistent domain
  adaptation,'' in \emph{CVPR}, 2019, pp. 2507--2516.

\bibitem{luosignificance}
Y.~Luo, P.~Liu, T.~Guan, J.~Yu, and Y.~Yang, ``Significance-aware information
  bottleneck for domain adaptive semantic segmentation,'' in \emph{ICCV}, 2019,
  pp. 6778--6787.

\bibitem{du2019ssf}
L.~Du, J.~Tan, H.~Yang, J.~Feng, X.~Xue, Q.~Zheng, X.~Ye, and X.~Zhang,
  ``Ssf-dan: Separated semantic feature based domain adaptation network for
  semantic segmentation,'' in \emph{ICCV}, 2019, pp. 982--991.

\bibitem{vu2019advent}
T.-H. Vu, H.~Jain, M.~Bucher, M.~Cord, and P.~P{\'e}rez, ``Advent: Adversarial
  entropy minimization for domain adaptation in semantic segmentation,'' in
  \emph{CVPR}, 2019, pp. 2517--2526.

\bibitem{Tsai_adaptseg_ICCV19}
Y.-H. Tsai, K.~Sohn, S.~Schulter, and M.~Chandraker, ``Domain adaptation for
  structured output via discriminative patch representations,'' in \emph{ICCV},
  2019, pp. 1456--1465.

\bibitem{yang2019adversarial}
J.~Yang, R.~Xu, R.~Li, X.~Qi, X.~Shen, G.~Li, and L.~Lin, ``An adversarial
  perturbation oriented domain adaptation approach for semantic segmentation,''
  in \emph{AAAI}, vol.~34, no.~07, 2020, pp. 12\,613--12\,620.

\bibitem{huang2020contextual}
J.~Huang, S.~Lu, D.~Guan, and X.~Zhang, ``Contextual-relation consistent domain
  adaptation for semantic segmentation,'' in \emph{ECCV}, 2020, pp. 705--722.

\bibitem{Haoran_2020_ECCV}
H.~Wang, T.~Shen, W.~Zhang, L.-Y. Duan, and T.~Mei, ``Classes matter: A
  fine-grained adversarial approach to cross-domain semantic segmentation,'' in
  \emph{ECCV}, 2020, pp. 642--659.

\bibitem{BCDM_Li2021BiClassifierDM}
S.~Li, F.~Lv, B.~Xie, C.~Liu, J.~Liang, and C.~Qin, ``Bi-classifier determinacy
  maximization for unsupervised domain adaptation,'' in \emph{AAAI}, 2021.

\bibitem{CDGA_Kim_Joung_Kim_Park_Kim_Sohn_2021}
M.~Kim, S.~Joung, S.~Kim, J.~Park, I.-J. Kim, and K.~Sohn, ``Cross-domain
  grouping and alignment for domain adaptive semantic segmentation,''
  \emph{AAAI}, vol.~35, pp. 1799--1807, 2021.

\bibitem{zou2018unsupervised}
Y.~Zou, Z.~Yu, B.~Kumar, and J.~Wang, ``Unsupervised domain adaptation for
  semantic segmentation via class-balanced self-training,'' in \emph{ECCV},
  2018, pp. 289--305.

\bibitem{zou2019confidence}
Y.~Zou, Z.~Yu, X.~Liu, B.~Kumar, and J.~Wang, ``Confidence regularized
  self-training,'' in \emph{ICCV}, 2019, pp. 5982--5991.

\bibitem{lian2019constructing}
Q.~Lian, F.~Lv, L.~Duan, and B.~Gong, ``Constructing self-motivated pyramid
  curriculums for cross-domain semantic segmentation: A non-adversarial
  approach,'' in \emph{ICCV}, 2019, pp. 6758--6767.

\bibitem{MLSL_iqbal2020mlsl}
J.~Iqbal and M.~Ali, ``Mlsl: Multi-level self-supervised learning for domain
  adaptation with spatially independent and semantically consistent labeling,''
  in \emph{WACV}, 2020, pp. 1864--1873.

\bibitem{CCM_li2020content}
G.~Li, G.~Kang, W.~Liu, Y.~Wei, and Y.~Yang, ``Content-consistent matching for
  domain adaptive semantic segmentation,'' in \emph{ECCV}, 2020, pp. 440--456.

\bibitem{PIT_lv2020cross}
F.~Lv, T.~Liang, X.~Chen, and G.~Lin, ``Cross-domain semantic segmentation via
  domain-invariant interactive relation transfer,'' in \emph{CVPR}, 2020, pp.
  4334--4343.

\bibitem{PixMatch_melas2021pixmatch}
L.~Melas-Kyriazi and A.~K. Manrai, ``Pixmatch: Unsupervised domain adaptation
  via pixelwise consistency training,'' in \emph{CVPR}, 2021, pp.
  12\,435--12\,445.

\bibitem{CLST_Marsden2021ContrastiveLA}
R.~A. Marsden, A.~Bartler, M.~D{\"o}bler, and B.~Yang, ``Contrastive learning
  and self-training for unsupervised domain adaptation in semantic
  segmentation,'' in \emph{IJCNN}.\hskip 1em plus 0.5em minus 0.4em\relax IEEE,
  2022, pp. 1--8.

\bibitem{SPCL_xie2021spcl}
B.~Xie, K.~Yin, S.~Li, and X.~Chen, ``Spcl: A new framework for domain adaptive
  semantic segmentation via semantic prototype-based contrastive learning,''
  \emph{arXiv preprint arXiv:2111.12358}, 2021.

\bibitem{MetaCorrection_guo2021metacorrection}
X.~Guo, C.~Yang, B.~Li, and Y.~Yuan, ``Metacorrection: Domain-aware meta loss
  correction for unsupervised domain adaptation in semantic segmentation,'' in
  \emph{CVPR}, 2021, pp. 3927--3936.

\bibitem{DACS_Tranheden_2021_WACV}
W.~Tranheden, V.~Olsson, J.~Pinto, and L.~Svensson, ``Dacs: Domain adaptation
  via cross-domain mixed sampling,'' in \emph{WACV}, 2021, pp. 1379--1389.

\bibitem{RCCR_zhou2021domain}
Q.~Zhou, C.~Zhuang, R.~Yi, X.~Lu, and L.~Ma, ``Domain adaptive semantic
  segmentation via regional contrastive consistency regularization,'' in
  \emph{ICME}.\hskip 1em plus 0.5em minus 0.4em\relax IEEE, 2022, pp. 01--06.

\bibitem{SAC_araslanov2021self}
N.~Araslanov and S.~Roth, ``Self-supervised augmentation consistency for
  adapting semantic segmentation,'' in \emph{CVPR}, 2021, pp. 15\,384--15\,394.

\bibitem{DSP_gao2021dsp}
L.~Gao, J.~Zhang, L.~Zhang, and D.~Tao, ``Dsp: Dual soft-paste for unsupervised
  domain adaptive semantic segmentation,'' \emph{arXiv preprint
  arXiv:2107.09600}, 2021.

\bibitem{CAMix_zhou2021context}
Q.~Zhou, Z.~Feng, Q.~Gu, J.~Pang, G.~Cheng, X.~Lu, J.~Shi, and L.~Ma,
  ``Context-aware mixup for domain adaptive semantic segmentation,'' \emph{IEEE
  Transactions on Circuits and Systems for Video Technology}, vol.~33, no.~2,
  pp. 804--817, 2022, \mbox{doi}:\url{10.1109/TCSVT.2022.3206476}.

\bibitem{xie2022sepico}
B.~Xie, S.~Li, M.~Li, C.~H. Liu, G.~Huang, and G.~Wang, ``Sepico:
  Semantic-guided pixel contrast for domain adaptive semantic segmentation,''
  \emph{IEEE Transactions on Pattern Analysis and Machine Intelligence},
  vol.~45, no.~7, pp. 9004--9021, 2023,
  \mbox{doi}:\url{10.1109/TPAMI.2023.3237740}.

\bibitem{chen2022deliberated}
L.~Chen, Z.~Wei, X.~Jin, H.~Chen, M.~Zheng, K.~Chen, and Y.~Jin, ``Deliberated
  domain bridging for domain adaptive semantic segmentation,'' \emph{NeurIPS},
  vol.~35, pp. 15\,105--15\,118, 2022.

\bibitem{zhang2019category}
Q.~Zhang, J.~Zhang, W.~Liu, and D.~Tao, ``Category anchor-guided unsupervised
  domain adaptation for semantic segmentation,'' in \emph{NeurIPS}, 2019, pp.
  433--443.

\bibitem{IntraDA_pan2020unsupervised}
F.~Pan, I.~Shin, F.~Rameau, S.~Lee, and I.~S. Kweon, ``Unsupervised
  intra-domain adaptation for semantic segmentation through self-supervision,''
  in \emph{CVPR}, 2020, pp. 3764--3773.

\bibitem{zheng2019unsupervised}
Z.~Zheng and Y.~Yang, ``Unsupervised scene adaptation with memory
  regularization in vivo,'' in \emph{IJCAI}, 2020.

\bibitem{DTST_wang2020differential}
Z.~Wang, M.~Yu, Y.~Wei, R.~Feris, J.~Xiong, W.-m. Hwu, T.~S. Huang, and H.~Shi,
  ``Differential treatment for stuff and things: A simple unsupervised domain
  adaptation method for semantic segmentation,'' in \emph{CVPR}, 2020, pp.
  12\,635--12\,644.

\bibitem{DAST_yu2021dast}
F.~Yu, M.~Zhang, H.~Dong, S.~Hu, B.~Dong, and L.~Zhang, ``Dast: Unsupervised
  domain adaptation in semantic segmentation based on discriminator attention
  and self-training,'' in \emph{AAAI}, vol.~35, no.~12, 2021, pp.
  10\,754--10\,762.

\bibitem{Zheng2021RectifyingPL}
Z.~Zheng and Y.~Yang, ``Rectifying pseudo label learning via uncertainty
  estimation for domain adaptive semantic segmentation,'' \emph{IJCV}, vol.
  129, no.~4, pp. 1106--1120, 2021.

\bibitem{UPLR_wang2021uncertainty}
Y.~Wang, J.~Peng, and Z.~Zhang, ``Uncertainty-aware pseudo label refinery for
  domain adaptive semantic segmentation,'' in \emph{ICCV}, 2021, pp.
  9092--9101.

\bibitem{ProDA_CRA_wang2021cross}
Z.~Wang, X.~Liu, M.~Suganuma, and T.~Okatani, ``Cross-region domain adaptation
  for class-level alignment,'' \emph{arXiv preprint arXiv:2109.06422}, 2021.

\bibitem{li2022class}
R.~Li, S.~Li, C.~He, Y.~Zhang, X.~Jia, and L.~Zhang, ``Class-balanced
  pixel-level self-labeling for domain adaptive semantic segmentation,'' in
  \emph{CVPR}, 2022, pp. 11\,593--11\,603.

\bibitem{hoffman2018cycada}
J.~Hoffman, E.~Tzeng, T.~Park, J.-Y. Zhu, P.~Isola, K.~Saenko, A.~Efros, and
  T.~Darrell, ``Cycada: Cycle-consistent adversarial domain adaptation,'' in
  \emph{ICML}, 2018, pp. 1989--1998.

\bibitem{LDR_yang2020label}
J.~Yang, W.~An, S.~Wang, X.~Zhu, C.~Yan, and J.~Huang, ``Label-driven
  reconstruction for domain adaptation in semantic segmentation,'' in
  \emph{ECCV}, 2020, pp. 480--498.

\bibitem{CDAM_yang2021context}
J.~Yang, W.~An, C.~Yan, P.~Zhao, and J.~Huang, ``Context-aware domain
  adaptation in semantic segmentation,'' in \emph{WACV}, 2021, pp. 514--524.

\bibitem{Yang_2020_CVPR}
Y.~Yang and S.~Soatto, ``Fda: Fourier domain adaptation for semantic
  segmentation,'' in \emph{CVPR}, 2020, pp. 4085--4095.

\bibitem{MCS_chung2022maximizing}
I.~Chung, D.~Kim, and N.~Kwak, ``Maximizing cosine similarity between spatial
  features for unsupervised domain adaptation in semantic segmentation,'' in
  \emph{WACV}, 2022, pp. 1351--1360.

\bibitem{SFG_DA_cardace2022shallow}
A.~Cardace, P.~Z. Ramirez, S.~Salti, and L.~Di~Stefano, ``Shallow features
  guide unsupervised domain adaptation for semantic segmentation at class
  boundaries,'' in \emph{WACV}, 2022, pp. 1160--1170.

\bibitem{li2019bidirectional}
Y.~Li, L.~Yuan, and N.~Vasconcelos, ``Bidirectional learning for domain
  adaptation of semantic segmentation,'' in \emph{CVPR}, 2019, pp. 6936--6945.

\bibitem{LTIR_kim2020learning}
M.~Kim and H.~Byun, ``Learning texture invariant representation for domain
  adaptation of semantic segmentation,'' in \emph{CVPR}, 2020, pp.
  12\,975--12\,984.

\bibitem{SAI2I_Luigi2020}
L.~Musto and A.~Zinelli, ``Semantically adaptive image-to-image translation for
  domain adaptation of semantic segmentation,'' in \emph{BMVC}, 2020.

\bibitem{ContenTransfer_Lee2021UnsupervisedDA}
S.~Lee, J.~Hyun, H.~Seong, and E.~Kim, ``Unsupervised domain adaptation for
  semantic segmentation by content transfer,'' in \emph{AAAI}, 2021.

\bibitem{ITEN_piva2021exploiting}
F.~J. Piva and G.~Dubbelman, ``Exploiting image translations via ensemble
  self-supervised learning for unsupervised domain adaptation,'' \emph{arXiv
  preprint arXiv:2107.06235}, 2021.

\bibitem{TridentAdapt_shen2021tridentadapt}
F.~Shen, A.~Gurram, A.~F. Tuna, O.~Urfalioglu, and A.~Knoll, ``Tridentadapt:
  Learning domain-invariance via source-target confrontation and self-induced
  cross-domain augmentation,'' in \emph{BMVC}, 2021.

\bibitem{DPL_cheng2021dual}
Y.~Cheng, F.~Wei, J.~Bao, D.~Chen, F.~Wen, and W.~Zhang, ``Dual path learning
  for domain adaptation of semantic segmentation,'' in \emph{CVPR}, 2021, pp.
  9082--9091.

\bibitem{CoarseToFine_ma2021coarse}
H.~Ma, X.~Lin, Z.~Wu, and Y.~Yu, ``Coarse-to-fine domain adaptive semantic
  segmentation with photometric alignment and category-center regularization,''
  in \emph{CVPR}, 2021, pp. 4051--4060.

\bibitem{mei2020instance}
K.~Mei, C.~Zhu, J.~Zou, and S.~Zhang, ``Instance adaptive self-training for
  unsupervised domain adaptation,'' in \emph{ECCV}, 2020, pp. 415--430.

\bibitem{li2005setred}
M.~Li and Z.-H. Zhou, ``Setred: Self-training with editing,'' in \emph{PAKDD},
  2005, pp. 611--621.

\bibitem{triguero2015self}
I.~Triguero, S.~Garc{\'\i}a, and F.~Herrera, ``Self-labeled techniques for
  semi-supervised learning: taxonomy, software and empirical study,''
  \emph{KAIS}, vol.~42, no.~2, pp. 245--284, 2015.

\bibitem{gong2023continuous}
R.~Gong, Q.~Wang, M.~Danelljan, D.~Dai, and L.~Van~Gool, ``Continuous
  pseudo-label rectified domain adaptive semantic segmentation with implicit
  neural representations,'' in \emph{Proceedings of the IEEE/CVF Conference on
  Computer Vision and Pattern Recognition}, 2023, pp. 7225--7235.

\bibitem{9852150}
Z.~Zheng and Y.~Yang, ``Adaptive boosting for domain adaptation: Toward robust
  predictions in scene segmentation,'' \emph{IEEE Transactions on Image
  Processing}, vol.~31, pp. 5371--5382, 2022,
  \mbox{doi}:\url{10.1109/TIP.2022.3195642}.

\bibitem{MixUp_zhang2017mixup}
H.~Zhang, M.~Cisse, Y.~N. Dauphin, and D.~Lopez-Paz, ``mixup: Beyond empirical
  risk minimization,'' \emph{arXiv preprint arXiv:1710.09412}, 2017.

\bibitem{CutMix_yun2019cutmix}
S.~Yun, D.~Han, S.~J. Oh, S.~Chun, J.~Choe, and Y.~Yoo, ``Cutmix:
  Regularization strategy to train strong classifiers with localizable
  features,'' in \emph{ICCV}, 2019, pp. 6023--6032.

\bibitem{CopyPaste_ghiasi2021simple}
G.~Ghiasi, Y.~Cui, A.~Srinivas, R.~Qian, T.-Y. Lin, E.~D. Cubuk, Q.~V. Le, and
  B.~Zoph, ``Simple copy-paste is a strong data augmentation method for
  instance segmentation,'' in \emph{CVPR}, 2021, pp. 2918--2928.

\bibitem{dwibedi2017cut}
D.~Dwibedi, I.~Misra, and M.~Hebert, ``Cut, paste and learn: Surprisingly easy
  synthesis for instance detection,'' in \emph{ICCV}, 2017, pp. 1301--1310.

\bibitem{remez2018learning}
T.~Remez, J.~Huang, and M.~Brown, ``Learning to segment via cut-and-paste,'' in
  \emph{ECCV}, 2018, pp. 37--52.

\bibitem{ClassMix_olsson2021classmix}
V.~Olsson, W.~Tranheden, J.~Pinto, and L.~Svensson, ``Classmix:
  Segmentation-based data augmentation for semi-supervised learning,'' in
  \emph{WACV}, 2021, pp. 1369--1378.

\bibitem{goodfellow2016deep}
I.~Goodfellow, Y.~Bengio, and A.~Courville, \emph{Deep learning}.\hskip 1em
  plus 0.5em minus 0.4em\relax MIT press, 2016.

\bibitem{kukavcka2017regularization}
J.~Kuka{\v{c}}ka, V.~Golkov, and D.~Cremers, ``Regularization for deep
  learning: A taxonomy,'' \emph{arXiv preprint arXiv:1710.10686}, 2017.

\bibitem{krizhevsky2012imagenet}
A.~Krizhevsky, I.~Sutskever, and G.~E. Hinton, ``Imagenet classification with
  deep convolutional neural networks,'' \emph{NeurIPS}, vol.~25, pp.
  1097--1105, 2012.

\bibitem{szegedy2016rethinking}
C.~Szegedy, V.~Vanhoucke, S.~Ioffe, J.~Shlens, and Z.~Wojna, ``Rethinking the
  inception architecture for computer vision,'' in \emph{CVPR}, 2016, pp.
  2818--2826.

\bibitem{MixMatch_berthelot2019mixmatch}
D.~Berthelot, N.~Carlini, I.~Goodfellow, N.~Papernot, A.~Oliver, and C.~Raffel,
  ``Mixmatch: A holistic approach to semi-supervised learning,'' \emph{arXiv
  preprint arXiv:1905.02249}, 2019.

\bibitem{ReMixMatch_berthelot2019remixmatch}
D.~Berthelot, N.~Carlini, E.~D. Cubuk, A.~Kurakin, K.~Sohn, H.~Zhang, and
  C.~Raffel, ``Remixmatch: Semi-supervised learning with distribution alignment
  and augmentation anchoring,'' \emph{arXiv preprint arXiv:1911.09785}, 2019.

\bibitem{FixMatch_sohn2020fixmatch}
K.~Sohn, D.~Berthelot, C.-L. Li, Z.~Zhang, N.~Carlini, E.~D. Cubuk, A.~Kurakin,
  H.~Zhang, and C.~Raffel, ``Fixmatch: Simplifying semi-supervised learning
  with consistency and confidence,'' \emph{arXiv preprint arXiv:2001.07685},
  2020.

\bibitem{UDA_NEURIPS2020_44feb009}
Q.~Xie, Z.~Dai, E.~Hovy, T.~Luong, and Q.~Le, ``Unsupervised data augmentation
  for consistency training,'' in \emph{NeurIPS}, vol.~33, 2020, pp. 6256--6268.

\bibitem{CrDoCo_chen2019crdoco}
Y.-C. Chen, Y.-Y. Lin, M.-H. Yang, and J.-B. Huang, ``Crdoco: Pixel-level
  domain transfer with cross-domain consistency,'' in \emph{CVPR}, 2019, pp.
  1791--1800.

\bibitem{zhou2020uncertainty}
Q.~Zhou, Z.~Feng, Q.~Gu, G.~Cheng, X.~Lu, J.~Shi, and L.~Ma,
  ``Uncertainty-aware consistency regularization for cross-domain semantic
  segmentation,'' \emph{arXiv preprint arXiv:2004.08878}, 2020.

\bibitem{MeanTeacher_Tarvainen2017MeanTA}
A.~Tarvainen and H.~Valpola, ``Mean teachers are better role models:
  Weight-averaged consistency targets improve semi-supervised deep learning
  results,'' in \emph{NeurIPS}, 2017.

\bibitem{ProDA_zhang2021prototypical}
P.~Zhang, B.~Zhang, T.~Zhang, D.~Chen, Y.~Wang, and F.~Wen, ``Prototypical
  pseudo label denoising and target structure learning for domain adaptive
  semantic segmentation,'' in \emph{CVPR}, 2021, pp. 12\,414--12\,424.

\bibitem{richter2016playing}
S.~R. Richter, V.~Vineet, S.~Roth, and V.~Koltun, ``Playing for data: Ground
  truth from computer games,'' in \emph{ECCV}, 2016, pp. 102--118.

\bibitem{ros2016synthia}
G.~Ros, L.~Sellart, J.~Materzynska, D.~Vazquez, and A.~M. Lopez, ``The synthia
  dataset: A large collection of synthetic images for semantic segmentation of
  urban scenes,'' in \emph{CVPR}, 2016, pp. 3234--3243.

\bibitem{cordts2016cityscapes}
M.~Cordts, M.~Omran, S.~Ramos, T.~Rehfeld, M.~Enzweiler, R.~Benenson,
  U.~Franke, S.~Roth, and B.~Schiele, ``The cityscapes dataset for semantic
  urban scene understanding,'' in \emph{CVPR}, 2016, pp. 3213--3223.

\bibitem{RobotCarDatasetIJRR}
W.~Maddern, G.~Pascoe, C.~Linegar, and P.~Newman, ``1 year, 1000 km: The oxford
  robotcar dataset,'' \emph{IJRR}, vol.~36, no.~1, pp. 3--15, 2017.

\bibitem{truong2023fredom}
T.-D. Truong, N.~Le, B.~Raj, J.~Cothren, and K.~Luu, ``Fredom: Fairness domain
  adaptation approach to semantic scene understanding,'' in \emph{Proceedings
  of the IEEE/CVF Conference on Computer Vision and Pattern Recognition}, 2023,
  pp. 19\,988--19\,997.

\bibitem{zhao2023learning}
D.~Zhao, S.~Wang, Q.~Zang, D.~Quan, X.~Ye, R.~Yang, and L.~Jiao, ``Learning
  pseudo-relations for cross-domain semantic segmentation,'' in
  \emph{Proceedings of the IEEE/CVF International Conference on Computer
  Vision}, 2023, pp. 19\,191--19\,203.

\bibitem{10766055}
S.~Jeong, J.~Kim, S.~Kim, and D.~Min, ``Revisiting domain-adaptive semantic
  segmentation via knowledge distillation,'' \emph{IEEE Transactions on Image
  Processing}, 2024, \mbox{doi}:\url{10.1109/TIP.2024.3501076}.

\bibitem{he2016deep}
K.~He, X.~Zhang, S.~Ren, and J.~Sun, ``Deep residual learning for image
  recognition,'' in \emph{CVPR}, 2016, pp. 770--778.

\bibitem{deng2009imagenet}
J.~Deng, W.~Dong, R.~Socher, L.-J. Li, K.~Li, and L.~Fei-Fei, ``Imagenet: A
  large-scale hierarchical image database,'' in \emph{CVPR}, 2009, pp.
  248--255.

\bibitem{cubuk2020randaugment}
E.~D. Cubuk, B.~Zoph, J.~Shlens, and Q.~V. Le, ``Randaugment: Practical
  automated data augmentation with a reduced search space,'' in \emph{CVPR
  Workshops}, 2020, pp. 702--703.

\bibitem{hoyer2022daformer}
L.~Hoyer, D.~Dai, and L.~Van~Gool, ``Daformer: Improving network architectures
  and training strategies for domain-adaptive semantic segmentation,'' in
  \emph{Proceedings of the IEEE/CVF conference on computer vision and pattern
  recognition}, 2022, pp. 9924--9935.

\bibitem{hoyer2022hrda}
------, ``Hrda: Context-aware high-resolution domain-adaptive semantic
  segmentation,'' in \emph{European conference on computer vision}.\hskip 1em
  plus 0.5em minus 0.4em\relax Springer, 2022, pp. 372--391.

\bibitem{hung2019adversarial}
W.~C. Hung, Y.~H. Tsai, Y.~T. Liou, Y.~Y. Lin, and M.~H. Yang, ``Adversarial
  learning for semi-supervised semantic segmentation,'' in \emph{BMVC}, 2019.

\bibitem{ReCo_liu2021bootstrapping}
S.~Liu, S.~Zhi, E.~Johns, and A.~J. Davison, ``Bootstrapping semantic
  segmentation with regional contrast,'' \emph{arXiv preprint
  arXiv:2104.04465}, 2021.

\bibitem{GuideMix_tu2021guidedmix}
P.~Tu, Y.~Huang, R.~Ji, F.~Zheng, and L.~Shao, ``Guidedmix-net: Learning to
  improve pseudo masks using labeled images as reference,'' \emph{arXiv
  preprint arXiv:2106.15064}, 2021.

\end{thebibliography}

%

\newpage
\begin{IEEEbiography}[{\includegraphics[width=1in,height=1.25in,clip,keepaspectratio]{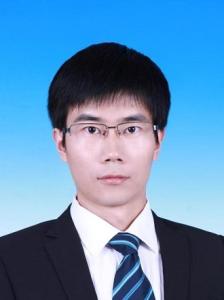}}]{Chuang Zhu}
is currently an associate professor with Beijing University of Posts and Telecommunications (BUPT), Beijing, China, where he leads the image processing group. He is also a member of the Center for Data Science, BUPT. Before that, he was a Post-Doctoral Research Fellow with the Department of the School of Electronics Engineering and Computer Science, Peking University, Beijing, China, from 2015 to 2017. He received a Ph.D. degree in Microelectronics from Peking University, Beijing, China. His research interests are in the areas of deep learning, image processing, multimedia content analysis, and machine learning algorithm optimization. He has published more than 50 publications in international magazines and conferences in these areas.
\end{IEEEbiography}

\begin{IEEEbiography}[{\includegraphics[width=1in,height=1.25in,clip,keepaspectratio]{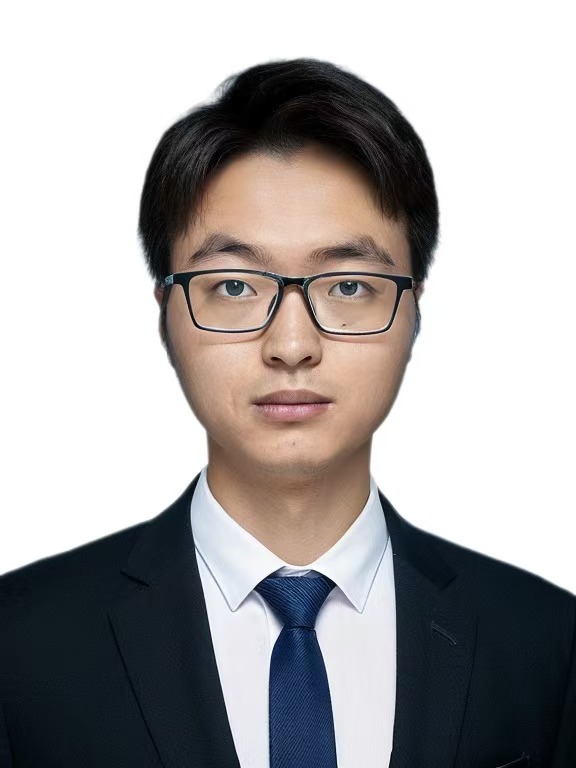}}]{Kebin Liu}
received the B.E. degree from Beijing University of Posts and Telecommunication, Beijing, China, in 2022. He is currently pursuing an M.E. degree in Artificial Intelligence at Beijing University of Posts and Telecommunication. His research interests include semantic segmentation and domain adaptation.
\end{IEEEbiography}

\begin{IEEEbiography}[{\includegraphics[width=1in,height=1.25in,clip,keepaspectratio]{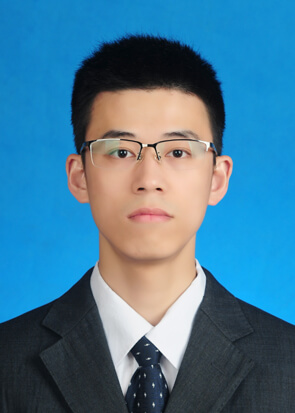}}]{Wenqi Tang}
received the B.E. degree from Chongqing University of Posts and Telecommunication, Chongqing, China, in 2020. He is currently pursuing an M.E. degree in Information and Communication at Beijing University of Posts and Telecommunication. His research interests include semantic segmentation and transfer learning.
\end{IEEEbiography}

\begin{IEEEbiography}[{\includegraphics[width=1in,height=1.25in,clip,keepaspectratio]{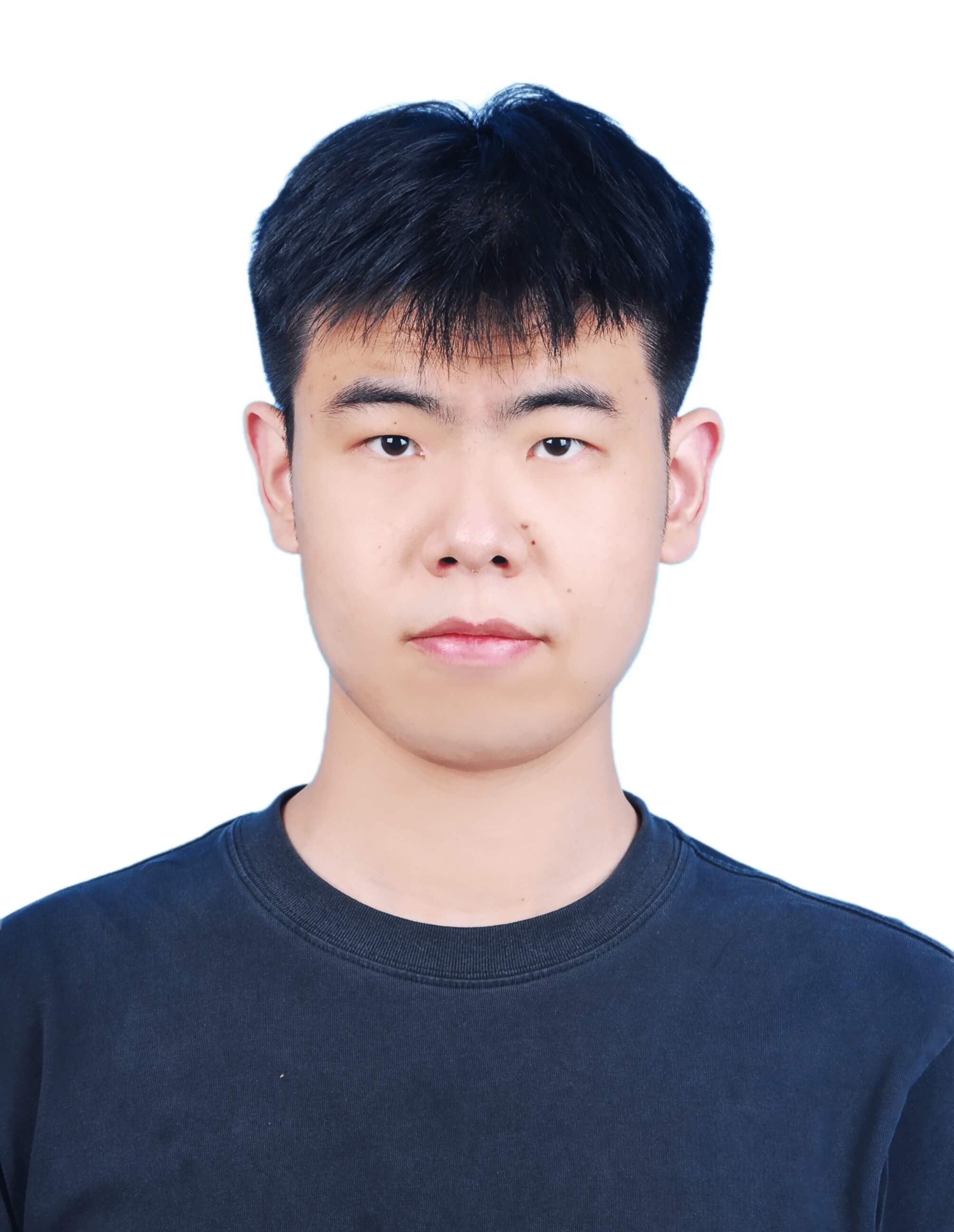}}]{Ke Mei}
received the B.E. and M.E. degrees from Beijing University of Posts and Telecommunications, Beijing, China, in 2018 and 2021 respectively. He is currently a computer vision researcher at Tencent Wechat AI, Beijing, China. His research interests include deep learning and computer vision.
\end{IEEEbiography}

\begin{IEEEbiography}[{\includegraphics[width=1in,height=1.25in,clip,keepaspectratio]{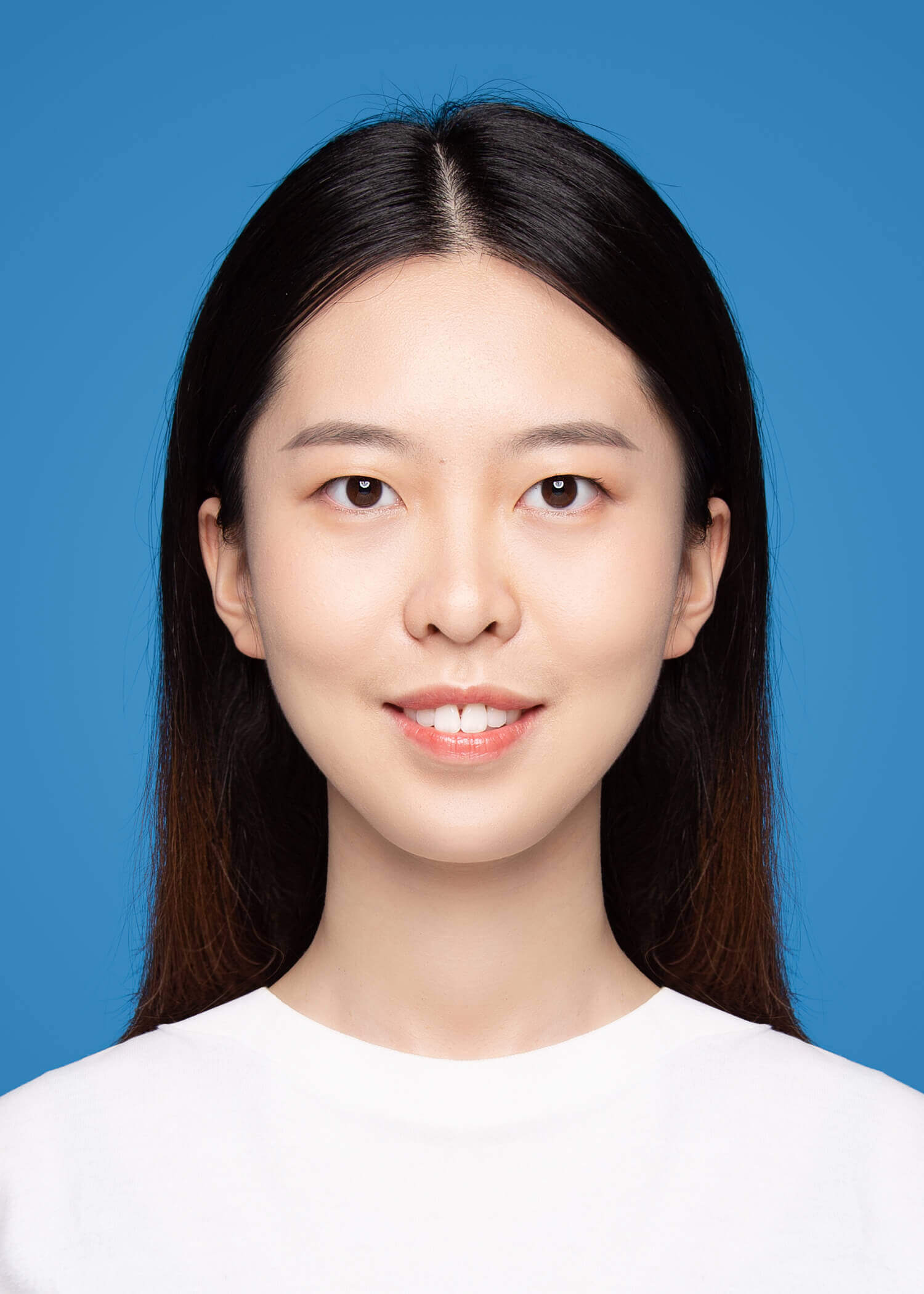}}]{Jiaqi Zou}
received the B.E. degree from Beijing University of Posts and Telecommunications, Beijing, China, in 2020. She is currently pursuing a Ph.D. degree at Beijing University of Posts and Telecommunications. Her research interests include computer vision and signal processing.
\end{IEEEbiography}

\begin{IEEEbiography}[{\includegraphics[width=1in,height=1.25in,clip,keepaspectratio]{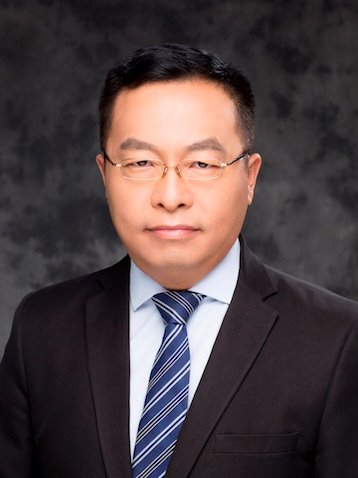}}]{Tiejun Huang}
(M'01-SM'12) is currently a Professor with the School of Electronic Engineering and Computer Science, Peking University, Beijing, China, where he is also the Director of the Institute for Digital Media Technology. He received the Ph.D. degree in pattern recognition and intelligent system from Huazhong (Central China) University of Science and Technology, Wuhan, China, in 1998, and the masters and bachelor’s degree in computer science from the Wuhan University of Technology, Wuhan, in 1995 and 1992, respectively. His research area includes video coding, image understanding, digital right management, and digital library. He has authored or co-authored over 100 peer-reviewed papers and three books. He is a member of the Board of Director for Digital Media Project, the Advisory Board of the IEEE Computing Society, and the Board of the Chinese Institute of Electronics.
\end{IEEEbiography}







\end{document}